\RequirePackage{fix-cm}
\documentclass[10pt,twocolumn,letterpaper]{article}

\usepackage[pagenumbers]{cvpr}

\usepackage[T1]{fontenc}

\usepackage{amsmath,amsfonts,bm}

\def\eqref#1{equation~\ref{#1}}

\def\1{\bm{1}}

\newcommand{\valid}{\mathcal{D_{\mathrm{valid}}}}

\DeclareMathAlphabet{\mathsfit}{\encodingdefault}{\sfdefault}{m}{sl}
\SetMathAlphabet{\mathsfit}{bold}{\encodingdefault}{\sfdefault}{bx}{n}

\usepackage{url}
\usepackage{xcolor}
\usepackage{xspace}
\usepackage[normalem]{ulem}
\usepackage{booktabs}
\usepackage{enumitem}
\usepackage{colortbl}
\usepackage{multirow}
\usepackage{tabularx}
\definecolor{EgoTableBaseline}{RGB}{236,181,176}
\definecolor{EgoTableQuestion}{RGB}{217,223,226}
\definecolor{EgoTableFrames}{RGB}{150,173,184}
\definecolor{EgoTableVideo}{RGB}{235,202,121}
\definecolor{EgoTableNeutral}{RGB}{217,223,226}
\definecolor{EgoTablePurple}{RGB}{180,169,193}
\newcommand{\EgoTableSetup}[2]{%
  \fontsize{#1}{\dimexpr #1pt*118/100\relax}\selectfont
  \setlength{\tabcolsep}{#2pt}%
  \renewcommand{\arraystretch}{1.10}%
  \setlength{\extrarowheight}{0pt}%
  \setlength{\aboverulesep}{2pt}%
  \setlength{\belowrulesep}{2pt}%
}

\usepackage{graphicx}
\usepackage{tikz}\usetikzlibrary{patterns}\usepackage{wrapfig}\usepackage{needspace}\usepackage{placeins}\usepackage{flafter}
\usepackage{amssymb}
\definecolor{EgoPlotTeal}{HTML}{227F78}
\definecolor{EgoPlotGold}{HTML}{D1942E}
\definecolor{EgoPlotGray}{HTML}{D1DBE1}
\definecolor{ResultTotalBG}{HTML}{E8F1FA}
\definecolor{ResultTotalText}{HTML}{1F4E79}

\colorlet{RevisionBlue}{black}
\colorlet{RevisionGreen}{black}
\DeclareRobustCommand{\rev}[1]{\textcolor{RevisionBlue}{#1}}
\DeclareRobustCommand{\revcite}[1]{\textcolor{RevisionGreen}{\citep{#1}}}
\definecolor{EgoPassBlue}{HTML}{1D4ED8}
\definecolor{EgoPassTeal}{HTML}{0F766E}
\definecolor{EgoPassPurple}{HTML}{7C3AED}
\definecolor{EgoPassOchre}{HTML}{B45309}
\definecolor{EgoPassMagenta}{HTML}{BE185D}

\newif\ifcomments
\commentstrue

\commentsfalse

\newcommand{\ours}{\textsc{EgoPass-Bench}\xspace}

\definecolor{cvprblue}{rgb}{0.21,0.49,0.74}
\usepackage[pagebackref,breaklinks,colorlinks,allcolors=cvprblue]{hyperref}

\title{\ours: Identifying Objects That May Affect the Wearer's Passage in \mbox{Egocentric} Video}

\author{%
Hongyu Wang$^{1}$ \quad Chuhan Li$^{1}$ \quad Congjin Xu$^{2}$ \quad
Xuandong Zhao$^{3}$ \quad Hua Shen$^{4}$\\[0.55em]
{\normalsize $^{1}$University of California, Santa Barbara \quad $^{2}$The University of Hong Kong}\\
{\normalsize $^{3}$University of California, Berkeley \quad $^{4}$New York University}%
}

\makeatletter
\patchcmd{\@maketitle}{\vspace*{12pt}}{\vspace*{20pt}}{}{%
    \PackageError{egopass}{Unable to adjust the title block spacing}{}%
}
\makeatother

\hypersetup{%
    pdftitle={EgoPass-Bench: Identifying Objects That May Affect the Wearer's Passage in Egocentric Video},
    pdfauthor={Hongyu Wang, Chuhan Li, Congjin Xu, Xuandong Zhao, Hua Shen}
}

\usepackage{longtable}
\usepackage{float}
\begin{document}
\maketitle
\begin{abstract}
Understanding an egocentric environment requires reasoning about how surrounding entities affect the observer's actions. Existing evaluations of multimodal foundation models (MFMs) examine observer localization, motion trajectories, and action feasibility, but leave the discovery and interpretation of evolving observer--entity interactions insufficiently assessed. We introduce \ours, a benchmark for observer-relative interaction understanding in real-world egocentric video. \ours comprises \rev{284} manually reviewed \rev{evaluation clips} recorded using Ray-Ban Meta (Gen 2) smart glasses and \rev{2,201} distinct questions across five tasks assessing motion perception, interaction judgments about specified entities, and relevant-entity discovery without target cues. Our evaluation of \rev{eight frame-based and six native-video configurations} reveals a pronounced discrepancy between motion perception and interaction understanding. \rev{The strongest evaluated model, GPT-6 Astra, correctly recognizes object motion in 91.37\% of 394 comparisons involving the same targets, but correctly identifies changes in passage constraints in only 66.50\%. Across all 574 entity-discovery questions, it trails the two-person human baseline by $23.65\pm7.45$ percentage points (paired 95\% CI), while exceeding the always-empty baseline (36.76\%) by at least 18.9 percentage points.} Further analysis reveals frequent inclusion of entities that neither pose a plausible physical conflict nor constrain passage, showing that accurate motion recognition does not ensure reliable relevant-entity discovery. 
\end{abstract}

\begin{figure*}[t]
\centering
\includegraphics[width=0.92\linewidth]{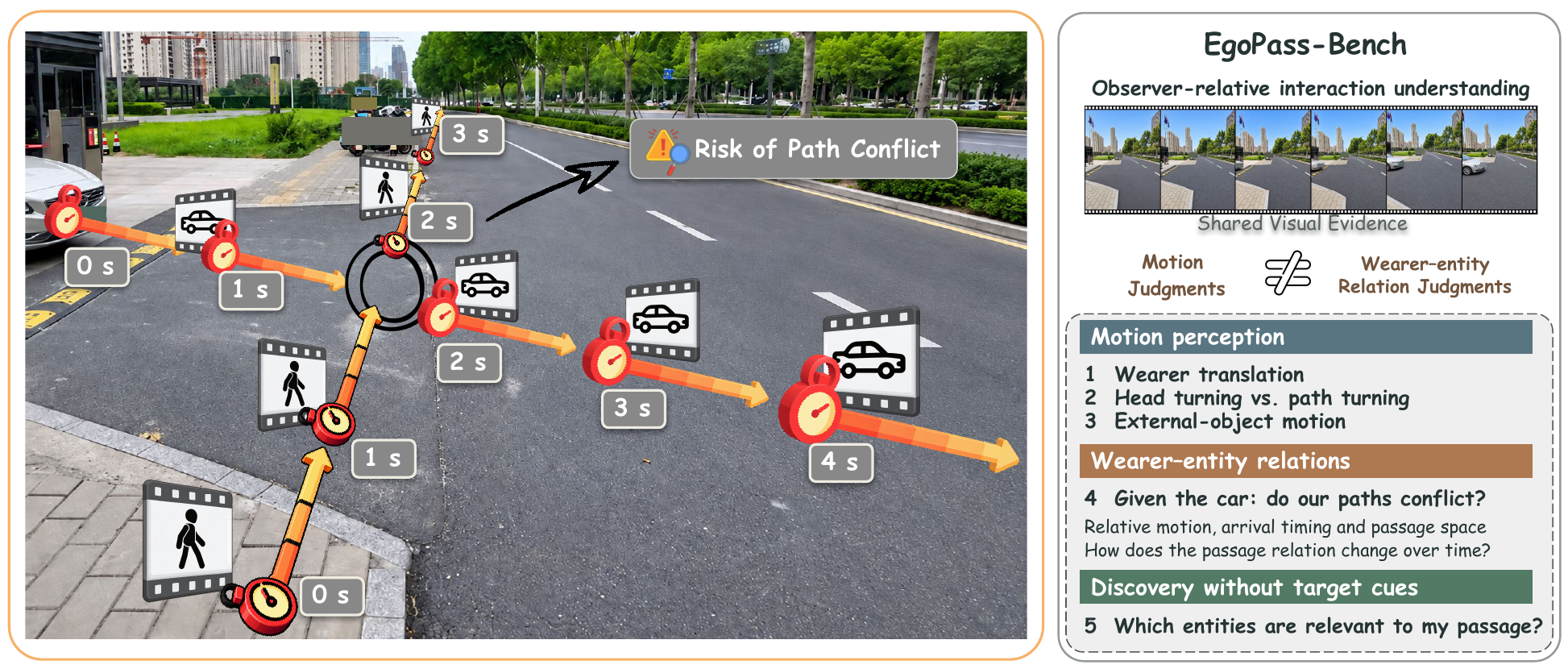}
\caption{\textbf{From motion perception to passage understanding.} \ours evaluates how wearer motion and external entities shape passage constraints. Models must separate observed states from their implications for the wearer and identify relevant entities without target cues.}
\label{fig:overview}
\end{figure*}

\section{Introduction}
\label{sec:introduction}

Visual assistants for smart glasses need to understand how the surrounding environment bears on wearers' ongoing actions. 
Recent work on continuous egocentric video understanding and proactive assistance has begun to move multimodal models beyond answering explicit questions toward deciding whether to intervene based on the user's context~\citep{gong2026vinci2,ran2026egoprobenchbenchmarkingpersonalizedproactive,xu2026pro2assistcontinuousstepawareproactive}.
Safe passage provides a concrete setting for this capability: an assistant must use continuous visual evidence to identify which external entities may require the wearer to adjust their speed or direction. 
This setting requires more than recognizing objects or describing scene changes; it requires understanding the consequences of those changes for the wearer's actions.

Such implications depend on the relationship between the wearer and external entities, rather than the motion state of either alone. 
Motion planning research analyzes potential collisions using the positions, velocities, and occupied space of both parties~\citep{Fiorini_1998}. 
Research on pedestrian interactions likewise shows that anticipated time to collision can explain avoidance behavior that physical distance alone does not adequately capture~\citep{PhysRevLett.113.238701}. 
Accordingly, an object's relevance cannot be determined solely by whether it is visible, approaching, or moving.
Instead, relevance depends on the motion trends of both parties, their relative arrival times, and the available passage space. 
An object on a separate path may not affect the wearer, whereas a stationary structure may block passage.
The challenge is to identify visually supported potential physical conflicts and passage constraints, rather than treating all salient objects as requiring caution.

This capability extends beyond wearable assistance. In robotics, systems must plan collision-avoiding velocities and paths based on their motion relative to surrounding obstacles~\citep{Fiorini_1998}. Similarly, in autonomous driving, detecting vehicles or pedestrians is only a starting point: systems must also determine which road users pose a risk and how that risk affects driving decisions~\citep{Malla_2023_WACV,10610270}. These settings share a common requirement: relating environmental states to an agent's actions.

Despite the importance of this capability across these settings, existing benchmarks assess only parts of it. MotionBench \citep{Hong_2025_CVPR} evaluates fine-grained motion understanding; SAW-Bench \citep{li2026sawbench} assesses observer-centric spatial reasoning and action feasibility; and BOSTVG \citep{ICLR2026_18374cc2} grounds entities specified by referring expressions.
Together, they leave untested whether a model can identify passage-relevant entities without target cues and determine whether their changing relation to the wearer calls for an adjustment in speed or direction.

To address this gap, we introduce \ours, a benchmark for evaluating observer-relative interaction understanding in real-world egocentric video. The benchmark grounds relevance in potential physical conflicts and safe-passage constraints. \rev{It contains 1,627 multiple-choice questions (MCQs) and 574 free-response questions about the recorded walks. In free-form generation (FFG), the respondent names the relevant entities and describes how the attention each requires changes.} Five tasks assess wearer translation, viewpoint and path turning, external-object motion, passage relations for specified objects, and relevant-entity discovery without target cues. The benchmark uses \rev{284} manually reviewed \rev{evaluation clips}, with each task answered independently so that motion recognition, relational reasoning, and entity discovery can be evaluated separately. We also evaluate models using the current clip alone or with the immediately preceding clip as context. This comparison tests whether recent visual history helps models understand the current situation.

Our evaluation of \rev{14 visual configurations} reveals a gap between motion perception and relevant-entity discovery. 
\rev{On the same clips, motion recognition and passage judgment can diverge even when the object is specified. Of 313 comparisons in which Astra correctly identifies both wearer translation and object motion, it misjudges the object's effect on passage in 96. When it correctly judges a named object's effect, it can still omit that object from its uncued response.}

\section{Related Work}
\label{sec:related_work}

%
\begin{table*}[t]
\color{RevisionBlue}
\centering
\caption{\looseness=-1\relax\textbf{Evaluation design of representative benchmarks.} Dimensions: \textsuperscript{1}wearable video, \textsuperscript{2}viewpoint--path distinction, \textsuperscript{3}cued reasoning and uncued entity discovery, \textsuperscript{4}changes in passage constraints, \textsuperscript{5}insufficient evidence versus irrelevance, \textsuperscript{6}separate entity-selection and label scoring, \textsuperscript{7}motion--relation--discovery evaluation on shared clips, and \textsuperscript{8}paired evaluation with and without visual history. $\checkmark$, $\triangle$, and $\times$ denote explicit, partial or related, and no explicit coverage. Group colors indicate general video (blue), egocentric video (purple), and safety and proactive assistance (yellow).}
\label{tab:benchmark_comparison}
\begingroup
\scriptsize
\setlength{\tabcolsep}{2.5pt}
\newlength{\bcBodyWidth}
\setlength{\bcBodyWidth}{\dimexpr\linewidth-40pt\relax}
\renewcommand{\arraystretch}{1.13}
\newcommand{\bcYes}{\textcolor[HTML]{007A60}{\ensuremath{\checkmark}}}
\newcommand{\bcPart}{\textcolor[HTML]{8A6500}{\ensuremath{\triangle}}}
\newcommand{\bcNo}{\textcolor[HTML]{AA5050}{\ensuremath{\times}}}
\begingroup
\EgoTableSetup{7.5}{1.8}
\begin{tabular}{@{\hspace{\tabcolsep}}>{\raggedright\arraybackslash}p{\dimexpr0.215001273\linewidth-3.655021645\tabcolsep+2pt\relax}>{\centering\arraybackslash}p{\dimexpr0.092772853\linewidth-1.577138495\tabcolsep-2pt\relax}>{\centering\arraybackslash}p{\dimexpr0.106655712\linewidth-1.813147105\tabcolsep\relax}>{\centering\arraybackslash}p{\dimexpr0.127542536\linewidth-2.168223107\tabcolsep\relax}>{\centering\arraybackslash}p{\dimexpr0.061505152\linewidth-1.045587579\tabcolsep\relax}>{\centering\arraybackslash}p{\dimexpr0.112450659\linewidth-1.911661207\tabcolsep\relax}>{\centering\arraybackslash}p{\dimexpr0.098567800\linewidth-1.675652597\tabcolsep\relax}>{\centering\arraybackslash}p{\dimexpr0.083475923\linewidth-1.419090697\tabcolsep\relax}>{\centering\arraybackslash}p{\dimexpr0.102028092\linewidth-1.734477568\tabcolsep\relax}@{}}
\arrayrulecolor{black!70}
\toprule
{\fontsize{7}{8.26}\selectfont\bfseries \textbf{Benchmark}} & {\fontsize{7}{8.26}\selectfont\bfseries \textbf{Wearable}} & {\fontsize{7}{8.26}\selectfont\bfseries \textbf{View--Path}} & {\fontsize{7}{8.26}\selectfont\bfseries \mbox{\textbf{Cued/Uncued}}} & {\fontsize{7}{8.26}\selectfont\bfseries \mbox{\textbf{Rel. $\Delta$}}} & {\fontsize{7}{8.26}\selectfont\bfseries \mbox{\textbf{Uncertainty}}} & {\fontsize{7}{8.26}\selectfont\bfseries \textbf{Set--Label}} & {\fontsize{7}{8.26}\selectfont\bfseries \mbox{\textbf{M--R--D}}} & {\fontsize{7}{8.26}\selectfont\bfseries \mbox{\textbf{Paired ctx.}}} \\
\midrule
\rowcolor{EgoTableFrames}\multicolumn{9}{l}{\textbf{General video}} \\
MVBench & \bcPart & \bcNo & \bcNo & \bcNo & \bcPart & \bcNo & \bcNo & \bcNo \\
Video-MME & \bcNo & \bcNo & \bcNo & \bcNo & \bcNo & \bcNo & \bcNo & \bcNo \\
TempCompass & \bcNo & \bcPart & \bcNo & \bcNo & \bcNo & \bcNo & \bcNo & \bcNo \\
TOMATO & \bcNo & \bcPart & \bcNo & \bcNo & \bcNo & \bcPart & \bcNo & \bcNo \\
MotionBench & \bcNo & \bcPart & \bcNo & \bcNo & \bcNo & \bcNo & \bcNo & \bcNo \\
VLM4D & \bcPart & \bcPart & \bcNo & \bcNo & \bcNo & \bcNo & \bcNo & \bcNo \\
\midrule
\rowcolor{EgoTablePurple}\multicolumn{9}{l}{\textbf{Egocentric video}} \\
EgoSchema & \bcYes & \bcNo & \bcNo & \bcNo & \bcNo & \bcNo & \bcNo & \bcNo \\
Minerva-Ego & \bcYes & \bcNo & \bcPart & \bcNo & \bcNo & \bcPart & \bcPart & \bcNo \\
SAW-Bench & \bcYes & \bcYes & \bcPart & \bcNo & \bcNo & \bcNo & \bcPart & \bcPart \\
\midrule
\rowcolor{EgoTableVideo}\multicolumn{9}{l}{\textbf{Safety and proactive assistance}} \\
DRAMA & \bcNo & \bcNo & \bcPart & \bcNo & \bcNo & \bcPart & \bcPart & \bcNo \\
RiskBench & \bcNo & \bcNo & \bcPart & \bcPart & \bcNo & \bcNo & \bcPart & \bcNo \\
PaSBench-Video & \bcNo & \bcNo & \bcPart & \bcPart & \bcNo & \bcPart & \bcPart & \bcPart \\
EgoPro-Bench & \bcPart & \bcNo & \bcPart & \bcNo & \bcNo & \bcPart & \bcPart & \bcPart \\
\midrule
\textbf{\mbox{\ours~(ours)}} & \bcYes & \bcYes & \bcYes & \bcYes & \bcYes & \bcYes & \bcYes & \bcYes \\
\bottomrule
\end{tabular}
\endgroup
\endgroup
\end{table*}

\paragraph{General video understanding.}
General video understanding evaluates whether models can recognize events and reason about temporal structure from continuous visual observations ~\citep{Li_2024_CVPR,Fu_2025_CVPR}. Table~\ref{tab:benchmark_comparison} summarizes representative benchmarks in general video understanding, egocentric video understanding, and safety understanding and proactive assistance. Beyond comprehensive evaluation across content domains and task types, specialized benchmarks examine event order and fine-grained motion~\citep{liu-etal-2024-tempcompass,Hong_2025_CVPR}, as well as motion, viewpoint changes, and the integration of spatial information across observations~\citep{Zhou_2025_VLM4D,zhang2026gstbenchvlmsdevelopglobal}. Recent work examines consistency across related questions and the temporal evidence supporting answers~\citep{ICLR2025_16ba99f2, fu2026videommev2stagebenchmarkscomprehensive,huang2026egvqabenchmarkingverifiablevideo}. However, recognizing objects and their motion does not establish that a model understands what these changes mean for the observer. We examine whether models distinguish environmental changes that affect the observer's actions from visible dynamics that do not constrain passage.

\paragraph{Egocentric video understanding.}
Egocentric video understanding studies environments from the wearer's perspective, where image changes reflect both the wearer's own motion and external activity. Egocentric datasets and benchmarks cover daily activities, episodic memory, and anticipation~\citep{Grauman_2022_CVPR,NEURIPS2023_90ce332a}. Recent research examines the wearer's perspective and human attention~\citep{jung2026egogapbenchbenchmarkingegocentricaction,tang2026egoeverythingbenchmarkhumanbehavior}, alongside visually grounded reasoning about interactions and object-state changes~\citep{dai2026egocotbenchbenchmarkinggroundedverifiable}. Another line of research concerns retaining and updating continuous experience, covering streaming episodic memory, reasoning about past and ongoing interactions, and the integration of experiences across days~\citep{forte2026egostreamdiagnosticbenchmarkstreaming,lei2026egosatcomprehensivebenchmarkegocentric,chen2026egomonthmonthlevelegocentricvideo}. Among the closest benchmarks, Minerva-Ego~\citep{Nagrani_2026_CVPR} emphasizes the visual grounding of reasoning, using spatiotemporal localization and reasoning traces to assess whether models identify and correctly use relevant evidence; SAW-Bench~\citep{li2026sawbench} focuses on the observer's spatial situation, including localization, trajectory reasoning, spatial memory, and action feasibility. However, these evaluations provide limited coverage of discovering passage-relevant entities without target cues and judging how their passage constraints change over time. We examine how an individual entity's passage constraint strengthens or eases as the wearer and entity move. Approaching alone does not imply increasing relevance, and leaving the field of view does not establish that a constraint has eased; unresolved cases retain an insufficient-evidence label.

\paragraph{Safety understanding and proactive assistance.}
Safety understanding and proactive assistance study how visual observations inform judgments about risks and user needs. Risk localization and explanation ground relevant entities and explain their implications for action~\citep{Malla_2023_WACV,Zhou_2025_WWW,gao2026unidriveunifiedvisionlanguagegrounding}. Other benchmarks assess hazard anticipation, timely warnings, and avoidance of unnecessary alarms~\citep{10610270,zhao2026pasbenchvideostreamingvideobenchmark}. As continuous video perception develops, research increasingly considers evidence accumulation and response timing~\citep{liu2026thinkingstreamingvideo,gong2026vinci2}, while incorporating individual needs, task progress, and ongoing interactions to determine whether, when, and how to provide assistance~\citep{ran2026egoprobenchbenchmarkingpersonalizedproactive,xu2026pro2assistcontinuousstepawareproactive,zhang2026streamarenacontinuousinteractivelonghorizon}. \rev{UNav provides route guidance for blind and low-vision people~\revcite{yang2025unav}; wearable warnings can reduce obstacle contacts alongside mobility aids~\revcite{pundlik2021homeuse}. \ours evaluates uncued entity discovery and changes in passage constraints, helping diagnose visual errors that could cause missed or unnecessary warnings.}

\begin{figure*}[!t]
\centering
\includegraphics[width=0.92\linewidth]{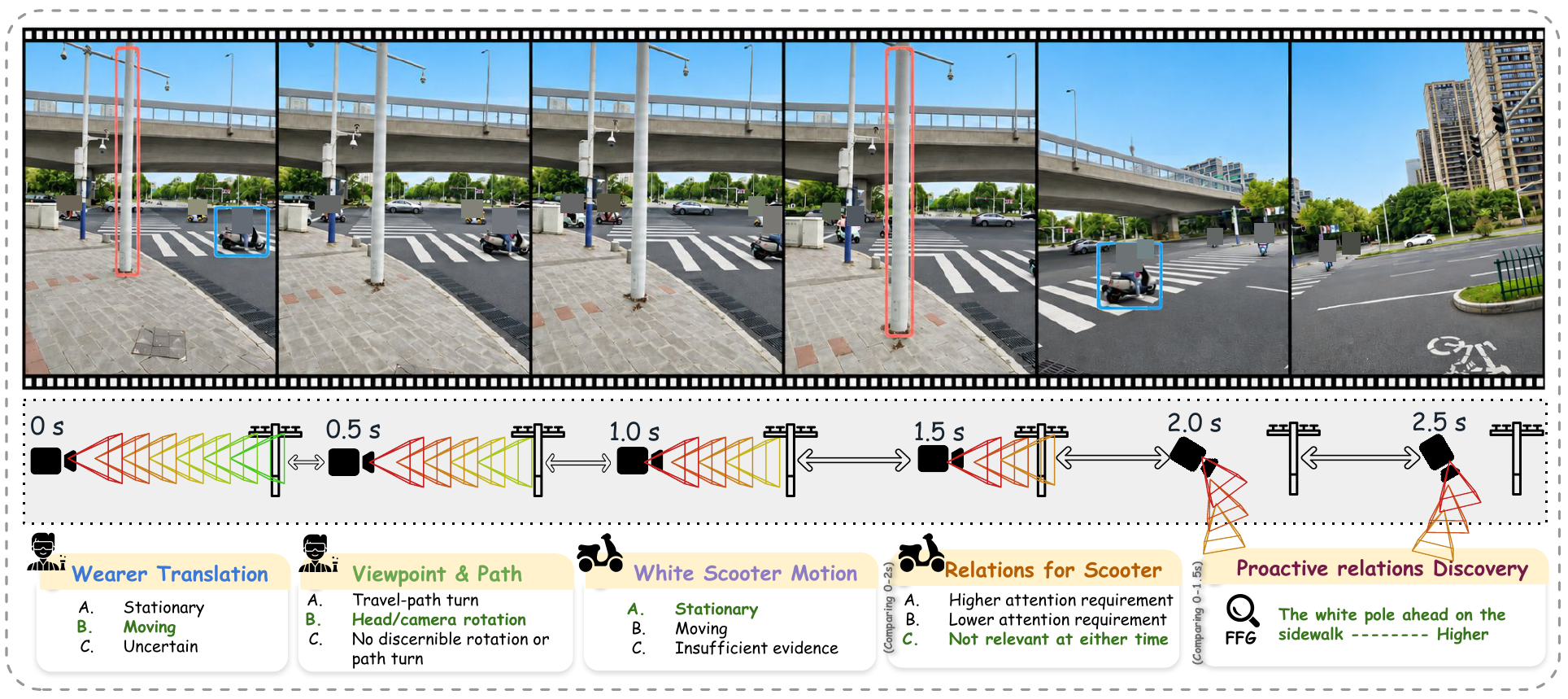}
\caption{\textbf{Five tasks share the same visual evidence.} Timestamped frames support judgments about wearer translation, turning, external object motion, passage relations for a specified entity, and autonomous discovery of all relevant entities.}
\label{fig:task_example}
\end{figure*}

\section{Task Definitions and Benchmark Construction}
\label{sec:benchmark}

\subsection{Task Definition}
\label{sec:task_definition}

\ours evaluates passage understanding in dynamic egocentric environments. \rev{For the frame-based models, each} input consists of a timestamped frame sequence $V=\{(I_j,t_j)\}_{j=1}^{N}$ and a question $q$, where $I_j$ is the $j$-th frame and $t_j$ is its timestamp. \rev{Native-video models receive the corresponding clip.} The question specifies an interval to examine or two time points to compare. Additional visual history is denoted by $H$: $H=\varnothing$ for No History; History provides the frame sequence from the immediately preceding clip.

Four MCQ tasks assess wearer translation, viewpoint and path turning, external-object motion, and passage relations for specified objects. For each question $q$, the model selects an answer from the corresponding candidate-answer set $\mathcal A_q$:
\begin{equation}
\hat a=f(V,H,q),\qquad \hat a\in\mathcal A_q,
\label{eq:task_mcq_output}
\end{equation}
where $f$ denotes the evaluated model and $\hat a$ is its selected answer.

Autonomous relevant-entity discovery uses FFG and provides no candidate-object cues. An entity should be included if it is passage-relevant at either of the two comparison time points. The passage-relevance criteria are defined in Section~\ref{sec:passage_relevance}.

The model must identify these entities and return a description and a change label for each:
\begin{equation}
\hat{\mathcal S}=f(V,H,q)=\{(\hat d_i,\hat\ell_i)\}_{i=1}^{m},
\label{eq:task_ffg_output}
\end{equation}
where $m$ is the number of reported entities, $\hat d_i$ is a description that enables the entity to be located in the scene, and $\hat\ell_i$ indicates the change in its required attention from $t_1$ to $t_2$. The label is \texttt{HIGHER}, \texttt{LOWER}, or \texttt{INSUFFICIENT\_EVIDENCE}, indicating increased attention, decreased attention, or insufficient evidence for a reliable comparison, respectively. When no entity satisfies the passage-relevance criteria, the correct output is the empty set.

All five tasks are answered independently to assess motion perception, passage-relation reasoning, and relevant-entity discovery separately.

\subsection{Defining Passage Relevance}
\label{sec:passage_relevance}

Reported driver perception--brake response times to unexpected traffic events are approximately 1.25--1.5 seconds~\citep{Green_2000}. \rev{For annotation, we use a two-second lookahead: annotators judge whether an object may constrain the wearer's path within that time, or whether an adjustment must begin within it.}

At either queried time, an object is passage-relevant if visual evidence and the wearer's and object's current motion indicate plausible collision risk or insufficient passage space within two seconds, or if avoiding a later conflict requires adjustment to begin within that window. Required attention denotes monitoring, preparation, or movement adjustment warranted by these passage constraints, rather than visual salience. Comparing this attention requirement between the two queried times captures how the object's passage constraint evolves. Even when the requirement decreases (\texttt{LOWER}), the object may still warrant monitoring or avoidance and remains a discovery target if it qualifies at either time. Full annotation criteria are provided in Appendix~\ref{app:evaluation_prompts}.


\subsection{Diagnostic Tasks}
\label{sec:diagnostic_tasks}

Situation awareness research distinguishes the perception of environmental elements, comprehension of their meaning, and projection of their subsequent states~\citep{Endsley_1995}. Motion planning and pedestrian interaction research further indicate that passage judgments depend on relative motion and potential conflicts between an agent and surrounding entities, rather than the state of a single entity~\citep{Fiorini_1998,PhysRevLett.113.238701}. These perspectives motivate five tasks assessing motion perception, passage-relation understanding, and relevant-entity discovery:

\begin{itemize}[leftmargin=*]
    \item[\textcolor{EgoPassBlue}{$\bullet$}] \textcolor{EgoPassBlue}{\textbf{Wearer Translation.}} Determine whether the wearer changes position during the specified interval, distinguishing bodily movement from image changes caused by head or camera rotation.
    \item[\textcolor{EgoPassTeal}{$\bullet$}] \textcolor{EgoPassTeal}{\textbf{Viewpoint and Path Turning.}} Identify the source, left--right direction, and temporal order of turning events, distinguishing head or camera rotation from changes in the actual travel path.
    \item[\textcolor{EgoPassPurple}{$\bullet$}] \textcolor{EgoPassPurple}{\textbf{External-Object Motion.}} Determine the motion state of a specified entity in the physical world, distinguishing stationary and moving states from confirmed starts or stops. The task requires avoiding the misinterpretation of image changes caused by wearer motion as object motion.
    \item[\textcolor{EgoPassOchre}{$\bullet$}] \textcolor{EgoPassOchre}{\textbf{Passage Relations for Specified Objects.}} Given a target object and two comparison time points, determine whether the object meets the passage-relevance criteria and whether the attention it requires increases or decreases. Objects that are irrelevant at both time points and cases with insufficient evidence are handled separately.
    \item[\textcolor{EgoPassMagenta}{$\bullet$}] \textcolor{EgoPassMagenta}{\textbf{Autonomous Relevant-Entity Discovery.}} Without candidate-object cues, identify all entities that meet the inclusion criteria at either comparison time point. For each entity, provide a description that enables it to be located in the scene and indicate the change in required attention. Return an empty set when no entity qualifies.
\end{itemize}

\subsection{\rev{Data Collection}}
\label{sec:video_collection}

We use second-generation Ray-Ban Meta smart glasses to record everyday passage scenes from the wearer's perspective, covering traffic intersections, sidewalks, residential areas, commercial districts, pedestrian streets, and parks. Recordings preserve natural movement, pauses, and leftward and rightward head turns. We manually select suitable segments from these recordings to create the \rev{evaluation clips}, which provide shared visual evidence for the five tasks. \rev{Across the 284 \rev{evaluation clips}, the evaluated targets span 11 object types. Encounters include pedestrians crossing the wearer's path, cyclists passing the wearer, and narrow passages bounded by parked vehicles or vegetation (Appendix~\ref{app:expanded_coverage}). The six settings describe collection locations; the object and interaction categories describe the constraints actually tested. These categories can overlap within a clip.}

\subsection{\rev{Evaluation Clips and Visual Evidence}}
\label{sec:clip_evidence}

\rev{Evaluation clips} localize passage judgments in time, avoiding a single judgment over changing interaction states while preserving sequential evidence of wearer motion, external-object motion, and viewpoint changes. \ours comprises \rev{284} manually reviewed \rev{evaluation clips} and \rev{2,201} questions across five tasks. \rev{The frame-based models receive frames} sampled at a nominal 2 fps and presented chronologically with \rev{their actual timestamps}. \rev{The median observation interval is 2.5\,s (range 2--5\,s).} Each question specifies an interval within the current clip.

\subsection{Question Construction and Quality Control}
\label{sec:question_construction}

\noindent\textbf{Question construction.} \rev{The camera wearers proposed reference answers based on the walks and the recorded footage. They also nominated the targets for the object-motion and passage-relation questions. The annotation team reviewed the references against the footage available to the respondent and made corrections where needed.} Annotations cover observer and object localization, motion trajectories, and judgments of observer--object relations to minimize ambiguity in the reference answers.

\noindent\textbf{Quality control.} Each recording is manually reviewed for video quality. Recordings with overly uniform scenes, poor visibility of key objects, or other issues that may hinder spatial reasoning are discarded and recorded again under the same filming guidelines. 

\rev{\noindent\textbf{Human baseline.} Two participants who took no part in filming or setting the reference answers practiced on 17 separate clips (144 questions), then independently answered the 2,201 benchmark questions. For FFG, participants and models received the same question and time window. Both had to identify the relevant objects themselves and supply a change label for each, or return an empty set. Neither saw candidate objects or reference answers. Practice included discussion of the reference criteria; the formal test provided no answer feedback, and the participants did not discuss their responses. Agreement was 82.69\% for turning ($\kappa=0.745$) and 82.07\% for passage relations ($\kappa=0.750$). Appendix~\ref{app:expanded_humans} gives agreement by task.}

\begin{table*}[!t]
\centering\color{RevisionBlue}
\caption{\rev{\textbf{Performance on \ours without visual history (\%).} For visual models, bold and underline mark the best and second-best results. FFG denominators, group-resampled 95\% confidence intervals and unresolved-identity bounds are in Appendix~\ref{app:expanded_results}.}}
\label{tab:nohistory}
\begingroup\small
\definecolor{TableBaseline}{RGB}{236,181,176}
\definecolor{TableQuestion}{RGB}{217,223,226}
\definecolor{TableFrames}{RGB}{150,173,184}
\definecolor{TableVideo}{RGB}{235,202,121}
\EgoTableSetup{8.5}{2.4}
\fontsize{8.5}{9.5}\selectfont
\renewcommand{\arraystretch}{1.04}
\setlength{\aboverulesep}{1.2pt}\setlength{\belowrulesep}{1.2pt}
\begin{tabular}{@{\hspace{\tabcolsep}}>{\raggedright\arraybackslash}p{85pt}>{\columncolor{black!5}\centering\arraybackslash}p{\dimexpr(\linewidth-85pt-17\tabcolsep)/8-1.5pt\relax}>{\centering\arraybackslash}p{\dimexpr(\linewidth-85pt-17\tabcolsep)/8+4pt\relax}>{\centering\arraybackslash}p{\dimexpr(\linewidth-85pt-17\tabcolsep)/8-1pt\relax}>{\centering\arraybackslash}p{\dimexpr(\linewidth-85pt-17\tabcolsep)/8-1pt\relax}>{\centering\arraybackslash}p{\dimexpr(\linewidth-85pt-17\tabcolsep)/8-0.5pt\relax}>{\centering\arraybackslash}p{\dimexpr(\linewidth-85pt-17\tabcolsep)/8-1pt\relax}>{\centering\arraybackslash}p{\dimexpr(\linewidth-85pt-17\tabcolsep)/8+3pt\relax}>{\centering\arraybackslash}p{\dimexpr(\linewidth-85pt-17\tabcolsep)/8-2pt\relax}@{}}
\arrayrulecolor{black!70}
\toprule
 & \multicolumn{5}{c}{Multiple choice} & \multicolumn{3}{c}{Free response} \\
\cmidrule(lr){2-6}\cmidrule(l){7-9}
{\fontsize{8}{9}\selectfont\textbf{Model}} & {\fontsize{8}{9}\selectfont\textbf{Overall}} & {\fontsize{8}{9}\selectfont\textbf{\shortstack{Wearer\\translation}}} & {\fontsize{8}{9}\selectfont\textbf{Turning}} & {\fontsize{8}{9}\selectfont\textbf{\shortstack{Object\\motion}}} & {\fontsize{8}{9}\selectfont\textbf{\shortstack{Passage\\relation}}} & {\fontsize{8}{9}\selectfont\textbf{Exact}} & {\fontsize{8}{9}\selectfont\textbf{Discovery}} & {\fontsize{8}{9}\selectfont\textbf{Labels}} \\
\midrule
\rowcolor{TableBaseline}\multicolumn{9}{l}{\textbf{Baselines}} \\
\textbf{Human (mean of two)} & \textbf{85.80} & \textbf{89.89} & \textbf{83.27} & \textbf{94.12} & \textbf{79.93} & \textbf{77.79} & \textbf{79.46} & \textbf{98.49} \\
Always empty & --- & --- & --- & --- & --- & 36.76 & 36.76 & --- \\
\midrule
\rowcolor{TableQuestion}\multicolumn{9}{l}{\textbf{Question-only model}} \\
GPT-5.6 Sol & 45.54 & 49.81 & \phantom{0}8.46 & 78.35 & 37.48 & --- & --- & --- \\
\midrule
\rowcolor{TableFrames}\multicolumn{9}{l}{\textbf{Multimodal models with chronological frames}} \\
GPT-6 Astra & \textbf{65.52} & \textbf{86.14} & 43.85 & \textbf{91.29} & \textbf{49.48} & \textbf{48.26} & \textbf{55.73} & \textbf{82.97} \\
Gemini 3.8 Flash & \underline{60.97} & 66.67 & \underline{45.77} & 87.29 & \underline{48.00} & \underline{43.80} & \underline{52.11} & \underline{76.09} \\
Qwen3.8 Max & 58.39 & 74.91 & 45.00 & \underline{87.53} & 38.67 & 33.45 & 43.88 & 71.48 \\
GPT-5.6 Sol & 57.47 & 61.05 & 33.85 & 86.59 & 46.81 & 35.54 & 44.56 & 74.46 \\
Gemini 3.1 Pro & 57.47 & 69.66 & 28.46 & 84.24 & 46.96 & 34.21 & 45.40 & 67.27 \\
Qwen3.8 Flash & 54.46 & 64.04 & 30.77 & 85.41 & 40.30 & 29.49 & 40.17 & 68.97 \\
Llama 4 Maverick & 47.20 & 39.33 & 38.08 & 80.00 & 33.19 & 14.44 & 29.46 & 38.46 \\
Llama 4 Scout & 44.38 & 36.33 & 35.00 & 73.88 & 32.59 & 28.70 & 39.91 & 49.59 \\
\midrule
\rowcolor{TableVideo}\multicolumn{9}{l}{\textbf{Multimodal models with native video}} \\
Qwen3.8 Max & 58.57 & \underline{79.03} & 40.38 & 84.00 & 41.48 & 28.97 & 40.08 & 59.16 \\
Qwen3.8 Flash & 56.36 & 77.53 & 36.92 & 83.29 & 38.52 & 29.07 & 34.82 & 52.63 \\
Qwen3.5-397B & 55.87 & 77.15 & 44.62 & 79.06 & 37.19 & 24.30 & 29.20 & 47.54 \\
GLM-5.3 Flash & 55.13 & 74.91 & 28.08 & 81.65 & 41.04 & 29.82 & 39.65 & 48.55 \\
Gemma 4 31B & 54.76 & 57.30 & \textbf{55.00} & 81.18 & 37.04 & 28.80 & 37.00 & 54.74 \\
MiniMax M3 & 52.49 & 66.29 & 34.23 & 81.41 & 35.85 & 26.18 & 31.68 & 50.00 \\
\bottomrule
\end{tabular}
\endgroup
\end{table*}
\subsection{Context Conditions}
\label{sec:context_conditions}

To examine the role of prior observations, we evaluate a subset of 54 \rev{evaluation clips} under two input conditions: \textbf{No History} and \textbf{History}. No History provides only the current clip. History additionally provides images from the immediately preceding clip of the same recording, without earlier footage, previous questions, model answers, or textual summaries. The first clip has identical inputs under both conditions. In both settings, models answer questions about the current interval. The added visual history thus serves as evidence for inferring the current state, rather than as a basis for carrying forward conclusions about earlier moments.

Both conditions use the same relevance criteria and scoring rules. \rev{The camera wearers provided references for each condition using the footage available in that condition. The preceding clip can reveal information that changes the reference answer.} Within each condition, all models use the same references.

\section{Evaluation Setup and Results}
\label{sec:results}

\subsection{Models and Metrics}
\label{sec:metrics}
\label{sec:human_reference}
\paragraph{Evaluation setting.}
\rev{The frame-based evaluation includes} eight models from four families: GPT~\citep{singh2026openaigpt5card}, Gemini~\citep{comanici2025gemini25pushingfrontier}, Llama~\citep{grattafiori2024llama3herdmodels}, and Qwen~\citep{qwen38}. The evaluated models are GPT-6 Astra, GPT-5.6 Sol, Gemini 3.8 Flash, Gemini 3.1 Pro, Llama 4 Scout, Llama 4 Maverick, Qwen3.8 Max, and Qwen3.8 Flash. Appendix~\ref{app:model_configurations} reports their inference configurations. \rev{Table~\ref{tab:nohistory} also includes six native-video configurations, the mean of two independent participants, and an always-empty baseline for FFG. All visual configurations are evaluated on the same 2,201 questions without history. In the question-only condition, GPT-5.6 Sol receives the question and answer options for each of the 1,627 MCQs.}

\paragraph{Evaluation metric.}
MCQ accuracy is the fraction of \rev{the 1,627 MCQs} answered correctly. Overall accuracy pools the questions across the four tasks rather than averaging task accuracies. The same 574 FFG questions are evaluated for entity discovery, conditional label accuracy, and exact set-and-label correctness. We match predicted entities to reference entities one to one. True positives (TP) count correct matches, false positives (FP) count additional reports, and false negatives (FN) count missed reference entities. Each reference entity can be matched once, so duplicate reports do not increase TP. For reference set $G$, the object score for one question is
\begin{equation}
s(G,P)=\begin{cases}
\max\{0,(\mathrm{TP}-\mathrm{FP})/|G|\},& |G|>0,\\
\mathbf{1}[P=\varnothing],& |G|=0,
\end{cases}
\label{eq:object_score}
\end{equation}
where $P$ is the predicted set. When the reference set is empty, an empty prediction receives a score of one and any nonempty prediction receives zero.

\rev{Discovery averages the per-question object scores. Exact counts responses with the complete entity set and all change labels correct. Labels is the proportion of matched entities with a correct change label, pooled across questions. A missing response earns zero on question-level metrics, even when the reference is empty. In Table~\ref{tab:nohistory}, each FFG point estimate uses the cases with a determinate score for that metric; the corresponding 95\% confidence intervals appear in Appendix~\ref{app:expanded_results}. We use 10,000 shared, stratified resamples of the 284 evaluation clips, keeping clips from the same recording and their questions together. The first scorer matches descriptions with model names hidden. A second scorer joins ambiguous cases; both revisit the permitted frames and record a judgment. Appendix~\ref{app:expanded_results} gives the denominators and the bounds obtained by including unresolved matches. Human means weight the two participants equally.}

\subsection{Performance without History}
\label{sec:nohistory_results}
Table~\ref{tab:nohistory} reports model performance across the five task types. \rev{The two participants average 85.80\% on MCQs and 79.46\% on Discovery. GPT-6 Astra has the highest point estimate among the 14 visual configurations on both: $65.5\%\pm2.7$ percentage points for MCQs and $55.7\%\pm6.7$ percentage points for Discovery (95\% CIs). Gemma 4 31B with native video scores highest on turning (55.0\%).} A model's overall rank therefore does not describe its strengths across all tasks. Performance is strongest on external-object motion and weaker on passage relations. \rev{The same ordering holds in all 14 configurations on 394 comparisons of the same targets. Astra answers 360 motion questions and 262 passage questions correctly, a paired difference of $24.9\pm5.2$ percentage points (95\% CI).}

\subsection{Differences within Model Families}
\label{sec:model_families}
Within-family rankings can change by task. Gemini 3.8 Flash leads Gemini 3.1 Pro on MCQs and discovery, \rev{and also scores higher on conditional labels}. \rev{Llama 4 Maverick outscores Scout on multiple choice, whereas Scout scores higher on Discovery.} \rev{Adding frames to the same questions improves GPT-5.6 Sol from 45.54\% to 57.47\% overall (paired gain of $11.9\pm3.1$ percentage points, 95\% CI). Its largest gain is on turning, from 8.46\% to 33.85\%. Native-video configurations range from 52.49\% to 58.57\% overall. Their input format and inference settings both differ from the frame-based configurations.}

\section{Analysis and Discussion}
\label{sec:analysis}
Appendix~\ref{app:matched_questions} reports a comparison with the question-only baseline on matched items and traces passage judgments and discovery of the same targets.
We analyze entity selection, change labels and insufficient evidence to locate the gap between motion recognition and passage understanding.

\subsection{Finding More Entities Does Not Ensure Better Understanding}
\label{sec:selection}
\begin{figure*}[!t]
\centering
\includegraphics[width=0.92\linewidth]{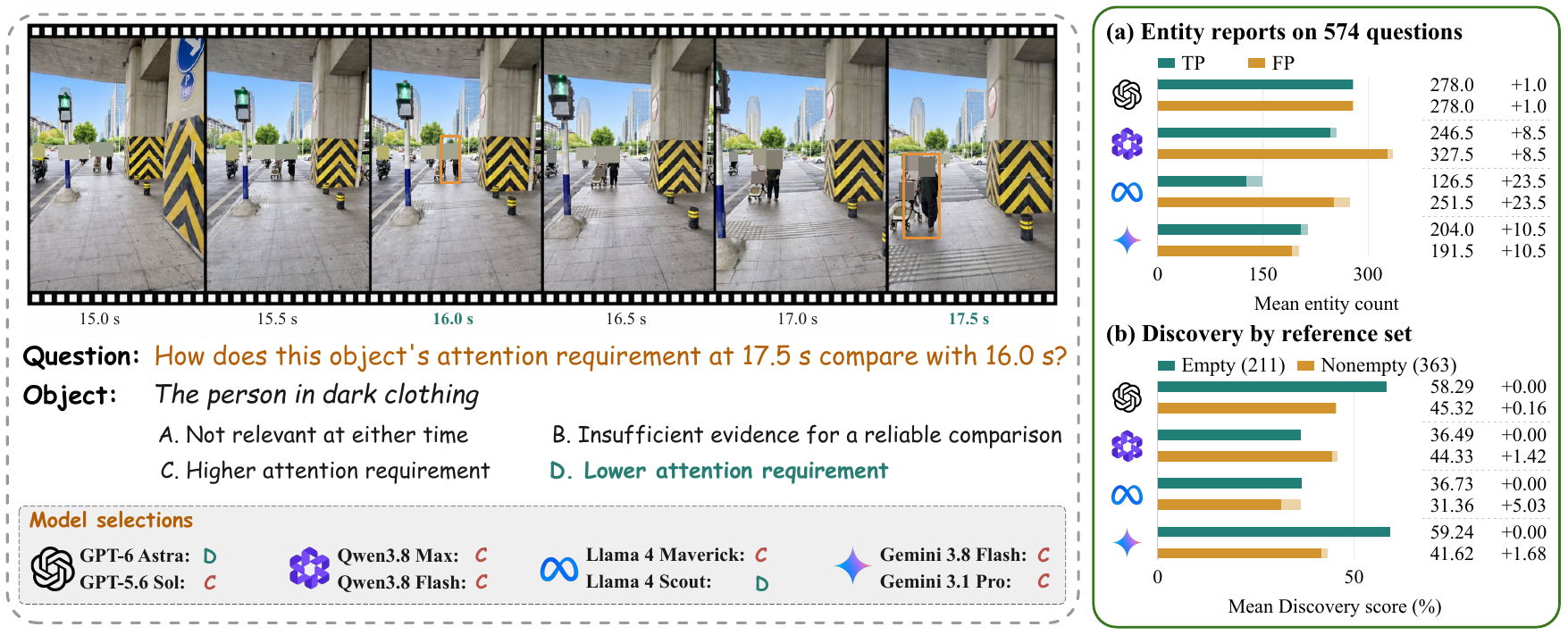}
\caption{\rev{\textbf{Left.} The approaching pedestrian yields, but six of eight frame-based models predict greater required attention. \textbf{Right.} Each family averages two frame-based models over 574 FFG questions. (a) Matched (TP) and extra (FP) entities. (b) Discovery on 211 empty and 363 nonempty references. Values are lower bounds; the rightmost ``+'' values give maximum additions from unresolved identities (entities in a, percentage points in b).}}
\label{fig:selection}
\label{fig:approaching}
\end{figure*}

Discovery errors are not limited to omissions. \rev{Figure~\ref{fig:selection} averages each family's two frame-based models on the same 574 FFG questions. Qwen averages 327.5--336 additional reports and 246.5--255 correct matches per model. GPT averages 278--279 correct matches, the highest of the four families even allowing for unresolved identities. Appendix~\ref{app:expanded_results} reports entity precision and recall for all 14 visual configurations.}

\rev{An empty response earns full Discovery credit on the 211 questions with no relevant entity, but none on the other 363. Always returning an empty set therefore scores 36.8\%. Astra correctly returns an empty set in 69.7\% of the 211 empty cases. On the 362 nonempty questions with a determinate score, its Discovery score falls to 47.6\%; the two participants average 77.0\% on all 363. Even using Astra's lowest score compatible with the unresolved match, its paired gain over always returning an empty set is $18.9\pm8.5$ percentage points (95\% CI; Table~\ref{tab:emptycontrols}).} \rev{Astra scores 60.48\% on crossing encounters ($n=62$ nonempty questions), versus 29.83--30.31\% on fixture- or vegetation-constrained passage ($n=138$); all 14 configurations share this ordering (Appendix~\ref{app:expanded_coverage}). Astra also exceeds the empty-set baseline after halving or doubling the false-positive penalty (Appendix~\ref{app:penalty_sensitivity}).}

\subsection{Approaching Does Not Necessarily Increase Passage Conflict}
\label{sec:approaching}

Approach describes relative motion, whereas conflict also depends on movement direction, the timing of entry into shared space, and available width. Being closer does not imply that the paths are harder to separate. Pedestrian interaction research likewise finds that anticipated time to collision explains observed interactions better than physical distance alone~\citep{PhysRevLett.113.238701}. This supports interpreting an object's implications through its dynamic relation to the observer.

\rev{From 16.0 to 17.5\,s in Figure~\ref{fig:approaching} (left), the pedestrian moves closer but steps left, opening a path for the wearer. DRAMA localizes risk objects and explains their risk from the driver's perspective~\revcite{Malla_2023_WACV}. \ours asks whether the same pedestrian now constrains passage more or less. The correct answer is less, yet six of eight frame-based models answer more even when the question names the pedestrian. In the free-response task, the model must also find relevant entities without a cue. In 67 comparisons with stationary targets and increasing constraints, Astra correctly recognizes stationarity in all 67 but misjudges 21 passage changes. It also predicts increased constraint in 112 of 181 decreasing-constraint questions.}

\subsection{Correct Local Judgments Can Still Yield Incomplete Answers}
\label{sec:outcomes}
A model may judge discovered entities accurately without describing the relevant environment completely. \rev{Figure~\ref{fig:outcomes} in Appendix~\ref{app:selection_full} covers all 574 FFG questions and their 431 reference entities for all 14 visual configurations. Astra correctly discovers and labels 231 entities, while GPT-5.6 Sol does so for 207--208, although both match 278--279 entities. Conditional label accuracy alone therefore does not establish complete coverage.} \rev{In 152 comparisons with identical time windows, the annotations link the named target to the sole FFG reference entity. Astra answered 56 of those named-target questions correctly. Asked independently to find all relevant entities, it omitted that target in eight of the 56, named it but made another entity or label error in 14, and returned the complete set with correct labels in 34.}

\rev{The entity-level breakdown separates answer quality from answer coverage. Astra's conditional label accuracy is 82.80--83.09\%, yet it omits 152--153 of the 431 reference entities. For the six native-video configurations, model-specific lower bounds range from 29 to 113 correctly discovered and labeled entities and from 230 to 365 omissions. The two-person human mean is 319 correctly discovered and labeled entities, five wrong labels, and 107 omissions on the same reference set. Correctly recognizing motion or interpreting a named target does not ensure that a model will discover that target without a cue.}

Situation awareness theory distinguishes perceiving environmental elements, understanding their meaning, and anticipating their subsequent states~\citep{Endsley_1995}. Our results echo this distinction: correct judgments about some entities can coexist with incorrect entity selection. Better understanding requires both accurate judgments and an appropriate, sufficiently complete entity set.

\subsection{Recognizing Insufficient Evidence Remains Difficult}
\label{sec:uncertainty}
Understanding the environment also requires recognizing what the observations cannot establish. \rev{Across all 72 questions with an insufficient-evidence reference, nine of 14 configurations give no correct answer and the highest score is eight (Table~\ref{tab:uncertainty} in Appendix~\ref{app:uncertainty}).} Occlusion, viewpoint changes, and an entity leaving view can limit what egocentric observations establish. Reliable understanding must distinguish supported changes from unresolved states. Together, these results expose gaps between entity selection, change interpretation, and evidence assessment.

\subsection{Discovery Remains Challenging with Visual History}
\label{sec:history}
\enlargethispage{-2\baselineskip}
History adds the immediately preceding clip while models continue to answer about the current clip. Figure~\ref{fig:history} in Appendix~\ref{app:history_full} compares both conditions on the same 257 questions, including 39 FFG questions. MCQ Overall improves for seven of eight models, but Discovery improves for only five. Gemini 3.8 Flash gains 12.82 percentage points in Discovery, whereas Llama 4 Scout loses 16.67 despite gaining 4.13 points in MCQ Overall. Gains also vary within a model: GPT-6 Astra improves in turning and object motion while declining in wearer translation. Thus, broader visual context can benefit some judgments without consistently improving which entities are identified as relevant to passage.

This uneven response is consistent with prior evidence that additional frames yield limited gains in fine-grained temporal understanding~\citep{cai2024temporalbenchbenchmarkingfinegrainedtemporal}, and that temporal perception varies across temporal aspects and task formats~\citep{liu-etal-2024-tempcompass}. \rev{The benefit of visual history therefore depends on the task and the model.} \rev{The preceding clip adds evidence and can change the reference answer. Each condition uses the same judgment criteria and a reference shared by all models; earlier answers are never supplied. Appendix~\ref{app:history_full} gives absolute scores, reference sources and coverage for the full 293-question History evaluation.}

\FloatBarrier
\section{Conclusion}
\label{sec:conclusion}
\rev{We introduce \ours to evaluate how models identify passage-relevant entities and interpret changing passage constraints in egocentric video. Across 14 visual configurations, accurate motion recognition and correct judgments about named targets do not ensure complete or reliable passage understanding. Models omit relevant entities, report irrelevant ones, and give definite answers when evidence is insufficient, while additional visual history offers uneven benefits. By evaluating motion, passage relations, and uncued entity discovery on shared visual evidence, \ours helps diagnose these gaps and assess how environmental changes affect the observer's actions.}
\label{sec:conclusion_end}
\clearpage
\section*{Ethics Statement}
\label{sec:ethics}
\rev{All camera wearers consented to the recording. Bystanders could enter the field of view incidentally. We have obscured visible faces in the manuscript figures. After acceptance, we will publicly release the benchmark with face-blurred evaluation clips, questions, reference annotations, and scoring materials. Before release, we will obscure all faces, readable license plates, and other identifying details, review each clip, and verify that the evidence for each question remains interpretable. The public release documentation will specify the applicable data license and identify any questions affected by redaction.}

\section*{LLM Usage Statement}
\rev{We used GPT to assist with language polishing. We also used generative AI to create some Scalable Vector Graphics (SVG) icons in the main paper, including icons representing the GPT, Qwen, Gemini, and Llama model families. The authors take full responsibility for the final content of this work, including text and graphical materials produced with AI assistance.}

\FloatBarrier
{
    \small
    \bibliographystyle{ieeenat_fullname}
    \bibliography{references}

\begin{thebibliography}{41}
\providecommand{\natexlab}[1]{#1}
\providecommand{\url}[1]{\texttt{#1}}
\expandafter\ifx\csname urlstyle\endcsname\relax
  \providecommand{\doi}[1]{doi: #1}\else
  \providecommand{\doi}{doi: \begingroup \urlstyle{rm}\Url}\fi

\bibitem[Cai et~al.(2024)Cai, Tan, Zhang, Zou, Zhang, Yao, Zhu, Gu, Zhong,
  Shang, Dou, Park, Gao, Lee, and
  Yang]{cai2024temporalbenchbenchmarkingfinegrainedtemporal}
Mu Cai, Reuben Tan, Jianrui Zhang, Bocheng Zou, Kai Zhang, Feng Yao, Fangrui
  Zhu, Jing Gu, Yiwu Zhong, Yuzhang Shang, Yao Dou, Jaden Park, Jianfeng Gao,
  Yong~Jae Lee, and Jianwei Yang.
\newblock Temporalbench: Benchmarking fine-grained temporal understanding for
  multimodal video models, 2024.

\bibitem[Chen et~al.(2026)Chen, Jiaxin, Tianyidan, Li, Qian, Xu, Tang, Wang,
  Yu, Li, Shi, Wang, Xu, Qiu, Li, Yang, Wang, and
  Yi]{chen2026egomonthmonthlevelegocentricvideo}
Weitao Chen, Hu Jiaxin, Xie Tianyidan, Yang Li, Yuyi Qian, Banghao Xu, Ziheng
  Tang, Shenyi Wang, Mingyue Yu, Duo Li, Jiacheng Shi, Gao Wang, Zhan Xu,
  Zhicheng Qiu, Xuanfu Li, Jian Yang, Lanjun Wang, and Zili Yi.
\newblock Egomonth: A month-level egocentric video benchmark for long-term
  spatiotemporal memory, 2026.

\bibitem[Comanici et~al.(2025)Comanici, Bieber, Schaekermann, Pasupat,
  Sachdeva, Dhillon, Blistein, Ram, Zhang, Rosen, Marris, Petulla, Gaffney,
  Aharoni, Lintz, Pais, Jacobsson, Szpektor, Jiang, Haridasan, Omran, Saunshi,
  Bahri, Mishra, Chu, Boyd, Hekman, Parisi, Zhang, Kawintiranon, Bedrax-Weiss,
  Wang, Xu, Purkiss, Mendlovic, Deutel, Nguyen, Langley, Korn, Rossazza, Ramé,
  Waghmare, Miller, Byrd, Sheshan, Hadsell, Bhardwaj, Janus, Rissa, Horgan,
  Abdagic, Belenki, Allingham, Singh, Guidroz, Srinivasan, Schmit, Chiafullo,
  Elisseeff, Jha, Kolhar, Berrada, Ding, Si, Mallick, Och, Erell, Ni, Latkar,
  Yang, Sirkovic, Feng, Leland, Hornung, Wu, Blundell, Alvari, Huang, Yip,
  Deur, Liu, Surita, Duque, Damen, Jia, Guez, Mircea, Sinha, Magni, Stradomski,
  Marian, Galić, Chen, Husain, Singhal, Grewe, Aubet, Song, Blanco, Rechis,
  Ho, Munoz, Zheng, Hamrick, Mather, Taitelbaum, Rutherford, Lei, Chen, Shukla,
  Moreira, Doi, Isik, Shabat, Rogozińska, Kolipaka, Chang, Vušak,
  Venkatachary, Noghabi, Bharti, Jun, Zaks, Green, Challagundla, Wong,
  Mohammad, Hirsch, Cheng, Naim, Proleev, Vincent, Singh, Krikun, Krishnan,
  Ghahramani, Atias, Aggarwal, Kirov, Vytiniotis, Koh, Chronopoulou, Dogra,
  Ion, Tyen, Lee, Weissenberger, Strohman, Balakrishna, Rae, Velic,
  de~Liedekerke, Elyada, Yuan, Liu, Shani, Kishchenko, Alessio, Li, Song, Kwei,
  Jankowski, Pappu, Namiki, Ma, Tripuraneni, Cherry, Ikonomidis, Ling, Ji,
  Westberg, Wright, Yu, Parkinson, Ramaswamy, Connor, Yeganeh, Grover,
  Kenwright, Litchev, Apps, Tomala, Halim, Castro-Ros, Li, Boral, Sho, Yarom,
  Malmi, Klinghoffer, Lin, Ansell, S, Zhao, Zuo, Santoro, Cheng, Demmessie,
  Liu, Brichtova, Culp, Braun, Graur, Ng, Mehta, Phillips, Sundberg, Godbole,
  Liu, Katariya, Rim, Seyedhosseini, Ammirati, Valfridsson, Malihi, Knight,
  Toor, Lampe, Ittycheriah, Chiang, Yeung, Fréchette, Rao, Wang, Srivastava,
  Zhang, Rhodes, Brand, Weesner, Figotin, Gimeno, Fellinger, Marcenac, Leal,
  Marcus, Cotruta, Cabrera, Luo, Garrette, Axelrod, Baltateanu, Barker, Chen,
  Toma, Ingram, Riesa, Kulkarni, Zhang, Liu, Wang, Polacek, Wu, Hui, Reyes, Su,
  Barnes, Malhi, Siddiqui, Feng, Damaschin, Pighin, Steiner, Yang, Boppana,
  Ivanov, Kandoor, Shah, Mujika, Huang, Choquette-Choo, Patel, Yu, Creswell,
  Jerry, Liu, Barros, Razeghi, Roy, Culliton, Xiong, Pan, Strohmann, Powell,
  Seal, DeCarlo, Shyam, Katircioglu, Wang, Hardin, Odisho, Broder, Chang, Nair,
  Shtefan, O'Brien, Agarwal, Potluri, Goyal, Jhindal, Thakur, Stuken, Lyon,
  Toutanova, Feng, Wu, Horn, Wang, Cullum, Taubman, Shrivastava, Shi,
  Tomlinson, Patel, Tu, Oflazer, Pongetti, Yang, Taïga, Perot, Pierse, Han,
  Drori, Iturrate, Chakrabarti, Yeung, Dopson, ting Chen, Kulshreshtha, Guo,
  Pham, Schuster, Chen, Polozov, Xing, Zhou, Kacham, Kukliansky, Miech,
  Yaroshenko, Chi, Douglas, Fei, Blondel, Myla, Madmoni, Wu, Keysers, Kjems,
  Albuquerque, Yu, D'sa, Plantan, Ionescu, Elias, Gupta, Vuyyuru, Alcober,
  Zhou, Ji, Hartmann, Puttagunta, Song, Amid, Stefanoiu, Lee, Pucciarelli,
  Wang, Raul, Petrov, Tian, Anklin, Nti, Gomes, Schumacher, Vesom,
  Panagopoulos, Bousmalis, Andor, Jacob, Zhang, Rosgen, Kecman, Tung, Belias,
  Goodman, Covington, Wieder, Saxena, Davoodi, Huang, Maddineni, Roulet,
  Campbell-Ajala, Sessa, Xintian, Wu, Lai, Collins, Haig, Sakenas, Xu,
  Giustina, Shafey, Charoenpanit, Garg, Ainslie, Severson, Arenas, Pathak,
  Rajayogam, Feng, Bakker, Li, Wichers, Rogers, Geng, Li, Jagerman, Jia,
  Olmert, Sharon, Mauger, Mariserla, Ma, Mohabey, Kim, Andreev, Pollom, Love,
  Jain, Agrawal, Schroecker, Fortin, Warmuth, Liu, Leach, Blok, Girirajan,
  Aharoni, Uria, Sozanschi, Goldberg, Ionita, Ribeiro, Zlocha, Birodkar,
  Lachgar, Yuan, Choudhury, Ginsberg, Zheng, Dibb, Graves, Lokhande, Rasskin,
  Muraru, Quick, Tata, Sermanet, Chawla, Karo, Wang, Zhang, Keller, Dragan, Su,
  Chou, Liu, Tao, Prabhakara, Wilson, Liu, Wang, Evans, Du, Castaño, Prasad,
  Mahdy, Gerlach, Reid, Kahn, Zait, Pillai, Ulrich, Wang, Wassenberg, Farkash,
  Yalasangi, Wang, Bauza, Bucher, Liu, Yan, Leung, Sindhwani, Barnes, Singh,
  Jurin, Chang, Bhumihar, Eiger, Citovsky, Withbroe, Li, Xue, Santo, Stoyanov,
  Raimond, Zheng, Gao, Listík, Kwasiborski, Saputro, Ozturel, Mallya,
  Majmundar, West, Caron, Wei, Castrejon, Vikram, Ramachandran, Dhawan, Park,
  Smoot, van~den Driessche, Blau, Malik, Liang, Hirsch, dos Santos, Weinstein,
  van~den Oord, Lall, FitzGerald, Jiang, Yang, Webster, Elqursh, Pope, Rotival,
  Raposo, Zhu, Dean, Alabed, Tran, Gupta, Gleicher, Austin, Rosseel, Umekar,
  Das, Sun, Chen, Misiunas, Zhou, Di, Loo, Newlan, Li, Ramasesh, Xu, Chen,
  Gandhe, Soricut, Gupta, Hu, El-Sayed, Garcia, Brusilovsky, Chen, Bolt, Huang,
  Gurney, Zhang, Pritzel, Wilkiewicz, Seybold, Shamanna, Fischer, Dean, Gill,
  Mcilroy, Bhowmick, Selier, Yang, Cheng, Magay, Tan, Varma, Walder, Kocisky,
  Nakashima, Natsev, Kwong, Gog, Zhang, Dieleman, Jimma, Ryabtsev, Brahma,
  Steiner, Du, Žužul, Žanić, Raghavachari, Gierke, Zheng, Petrova, Dauphin,
  Liu, Kessler, Hand, Duvarney, Kim, Lee, Hussenot, Hui, Smith, Jain, Xia,
  Tomar, Amiri, Phan, Fuchs, Weyand, Tomasev, Cordell, Liu, Mallinson, Joshi,
  Crawford, Suggala, Chien, Fernando, Sanchez-Vargas, Williams, Crone, Luo,
  Karpov, Shan, Thurk, Strudel, Voigtlaender, Patil, Dozat, Khodaei, Singla,
  Ambroszczyk, Wu, Chang, Roark, Hegde, Ding, Filos, Wu, Pinto, Liu, Khanna,
  Pandey, Mcloughlin, Li, Haves, Zhou, Buchatskaya, Leal, de~Boursac, Akazawa,
  Anderson, Chen, Somandepalli, Liang, Goenka, Winkler, Grushetsky, Ding,
  Smith, Ye, Pont-Tuset, Li, Li, Golany, Wegner, Jiang, Barak, Shangguan,
  Vértes, Wong, Bornschein, Tudor, Bevilacqua, Schaul, Rawat, Zhao, Axiotis,
  Meng, McLean, Lai, Beattie, Kushman, Liu, Kutzman, Lang, Ye, Netrapalli,
  Mishra, Khan, Goel, Willoughby, Tian, Zhuang, Chen, Tsai, Kementsietsidis,
  Khare, Keeling, Xu, Waters, Altché, Popat, Mittal, Saxton, Badawy, Mathieu,
  Zheng, Zhou, Ranka, Shin, Duan, Salimans, Mihailescu, Shaham, Chang, Assael,
  Dikkala, Izzard, Cohen-Addad, Graves, Feinberg, Chung, Strouse, Karmon,
  Sharifzadeh, Ashwood, Pham, Blanton, Vasiloff, Barber, Geller, Zhou, Zubach,
  Huang, Zhang, Gupta, Young, Proskurnia, Votel, Gabeur, Barcik, Tripathi, Yu,
  Yan, Changpinyo, Pavetić, Coyle, Fujii, Mendez, Zhou, Rajamani, Hechtman,
  Cao, Juan, Tan, Dalibard, Du, Clay, Yao, Jia, Vijaykumar, Zhou, Bai, Hung,
  Pecht, Todorov, Khadke, Gupta, Lahoti, Autef, Duddu, Lee-Thorp, Bykovsky,
  Misiunas, Flennerhag, Thangaraj, McGiffin, Nado, Kunesch, Noever, Hertz,
  Liang, Stone, Palmer, Daruki, Pramanik, Põder, Kyker, Khan, Sluzhaev,
  Ritter, Ruderman, Zhou, Nagpal, Vodrahalli, Necula, Barham, Pavlick,
  Hartford, Shafran, Zhao, Mikuła, Eccles, Shimokawa, Garg, Vilnis, Chen,
  Shumailov, Lee, Abdelhamed, Xie, Cohen, Hlavnova, Malkin, Sitawarin, Lottes,
  Coquinot, Yu, Kumar, Zhang, Mahendru, Ahmed, Martens, Chen, Boag, Peng,
  Devin, Klimovskiy, Phuong, Vainstein, Xie, Ramabhadran, Howard, Yu, Goswami,
  Cui, Shleifer, Pinto, Yeh, Yang, Javanmardi, Ethier, Lee, Orbay, Kotecha,
  Bromberg, Shaw, Thornton, Rosenthal, Gu, Thomas, Gemp, Ayyar, Ushio, Selvan,
  Wee, Liu, Majzoubi, Yu, Abernethy, Liechty, Pan, Nguyen, Qiong, Hu, Perrin,
  Arora, Pitler, Wang, Shivakumar, Prost, Limonchik, Wang, Gao, Cour, Buch,
  Gui, Ivanova, Neubeck, Chan, Kim, Chen, Goyal, Chung, Liu, Su, Petrushkina,
  Shen, Joulin, Xu, Lin, Kulizhskaya, Chelba, Vasudevan, Collins, Bashlovkina,
  Lu, Fritz, Park, Zhou, Su, Tanburn, Sushkov, Rasquinha, Li, Prendki, Li, LV,
  Sharma, Fitoussi, Huang, Dai, Dao, Burrows, Prior, Qin, Pundak, Sjoesund,
  Khurshudov, Zhu, Webson, Kemp, Tan, Agrawal, Sargsyan, Cheng, Stephan,
  Kwiatkowski, Reid, Byravan, Michaely, Heess, Zhou, Goenka, Carpenter,
  Levskaya, Wang, Roberts, Leblond, Chikkerur, Ginzburg, Chang, Riachi,
  Chuqiao, Xu, Borsos, Pliskin, Pawar, Lustman, Kirkwood, Anand, Chaudhary,
  Kalb, Milan, Augenstein, Goldie, Prince, Raman, Sun, Xia, Cohen, Huo, Camp,
  Ellis, Zilka, Torres, Patel, Arora, Chan, Adler, Ayoub, Liang, Jamil, Jiang,
  Baumgartner, Sun, Karov, Akulov, Zheng, Cai, Fantacci, Rubin, Acha, Wang,
  D'Souza, Sathyanarayana, Dai, Rowe, Simanovsky, Goldman, Kuang, Pan,
  Rosenberg, Rojas-Esponda, Dutta, Zeng, Jurenka, Farquhar, Bansal, Iqbal,
  Roelofs, Joung, Beak, Ryu, Poplin, Wu, Alayrac, Buthpitiya, Ronneberger,
  Habtegebriel, Li, Cavallaro, Wei, Bensky, Denk, Ganapathy, Stanway, Joshi,
  Bertolini, Lo, Ma, Charles, Sampemane, Sahni, Chen, Askham, Gaddy, Young,
  Tan, Eyal, Bražinskas, Zhong, Wu, Epstein, Bailey, Hard, Lee, Goldshtein,
  Ruiz, Badawi, Lochbrunner, Kearns, Brown, Pardo, Weber, Yang, Jiang, Akin,
  Fu, Wainwright, Zou, Gaba, Manzagol, Kan, Song, Zainullina, Lin, Ko,
  Deshmukh, Jindal, Svensson, Tyam, Zhao, Kaeser-Chen, Baird, Moradi, Hall,
  Guo, Tsang, Liang, Pereira, Ganesh, Korotkov, Adamek, Thiagarajan, Tran,
  Chen, Tar, Jain, Dasgupta, Bilal, Reitter, Zhao, Vezzani, Gehman, Mehta,
  Beltrone, Dotiwalla, Guadarrama, Abbas, Karp, Georgiev, Ferng, Brockschmidt,
  Peng, Hirnschall, Verma, Bi, Xiao, Dabush, Xu, Wallis, Parker, Wang, Xu,
  Safarli, Tewari, Zhang, Kim, Gesmundo, Thomas, Levi, Chowdhury, Rao, Garst,
  Conway-Rahman, Ran, McKinney, Xiao, Yu, Agrawal, Stjerngren, Ionescu, Chen,
  Sharma, Chiu, Liu, Franko, Sanford, Cai, Michel, Ganapathy, Labanowski,
  Garrett, Vargas, Sun, Gale, Buschmann, Desjardins, Ghelani, Jain, Verma,
  Asawaroengchai, Eisenschlos, Harlalka, Kazawa, Metzler, Howland, Jian, Ades,
  Shah, Gangwani, Lee, Ring, Hernandez, Reich, Sinha, Sathe, Kovac, Gill,
  Kannan, D'olimpio, Sevenich, Whang, Kim, Sim, Chen, Zhang, Lall, Matias, Jia,
  Friesen, Nasso, Thapliyal, Perozzi, Yu, Shekhawat, Huda, Grabowski, Wang,
  Sreevatsa, Dib, Hassen, Schuh, Milutinovic, Welty, Quinn, Shah, Wang,
  Barth-Maron, Frye, Axelsson, Zhu, Ma, Giannoumis, Sedghi, Ye, Luan, Aydin,
  Chandra, Sampathkumar, Huang, Lavrenko, Eleryan, Hong, Hansen, Carthy,
  Samanta, Ćevid, Wang, Li, Voznesensky, Hoffman, Terzis, Sehwag, Fidel, He,
  Cai, He, Feng, Nikoltchev, Phatale, Chase, Lawton, Zhang, Ouyang, Tragut,
  Manshadi, Narayanan, Shen, Gao, Bolukbasi, Roy, Li, Golovin, Panait, Qin,
  Han, Anthony, Kudugunta, Patraucean, Ray, Chen, Yang, Bhatia, Talluri,
  Morris, Ražnatović, Brownfield, An, Peng, Kane, Zheng, Duduta, Kessinger,
  Noraky, Liu, Rong, Veličković, Rush, Goldin, Wei, Garlapati, Pantofaru,
  Kwon, Ni, Noland, Trapani, Beaufays, Roy, Chow, Turker, Cideron, Mei, Clark,
  Dou, Bošnjak, Leith, Du, Yazdanbakhsh, Nasr, Kwak, Sheth, Kaskasoli, Anand,
  Lakshminarayanan, Jerome, Bieber, Chu, Senges, Shen, Sridhar, Ndebele,
  Beyret, Mohamed, Chen, Freitag, Guo, Liu, Roit, Chen, Yan, Stone, Co-Reyes,
  Cole, Scellato, Azizi, Hashemi, Jin, Iyer, Valentine, György, Ahuja, Diaz,
  Lee, Clement, Kong, Garmon, Watts, Bhatia, Gupta, Miecnikowski, Vallet, Taly,
  Loper, Joshi, Atwood, Chick, Collier, Iliopoulos, Trostle, Gunel,
  Leal-Cavazos, Hrafnkelsson, Guzman, Ju, Forbes, Emond, Chauhan, Caine, Xiao,
  Zeng, Moufarek, Murphy, Meng, Gupta, Riedel, Das, Lawal, Narayan, Sosea,
  Swirhun, Friso, Neyshabur, Lu, Girgin, Wunder, Yvinec, Pyne, Carbune,
  Rijhwani, Guo, Doshi, Briukhov, Bain, Hitron, Wang, Gupta, Chen, Du, Zhang,
  Shah, Akula, Dylla, Kachra, Kuo, Zou, Wang, Xu, Zhu, Snyder, Menon, Firat,
  Mordatch, Yuan, Ponomareva, Blevins, Moore, Wang, Chen, Scholz, Dwornik, Lin,
  Li, Antognini, I, Song, Miller, Kalra, Raveret, Akerlund, Wu, Nystrom,
  Godbole, Liu, DeBalsi, Zhao, Liu, Caciularu, Lax, Khandelwal, Langston,
  Bailey, Lattanzi, Wang, Kovelamudi, Mondal, Guruganesh, Hua, Roval,
  Wesołowski, Ingale, Halcrow, Sohn, Angermueller, Raad, Stickgold, Lu, Kosik,
  Xie, Lillicrap, Huang, Zhang, Paulus, Farabet, Wertheim, Wang, Joshi, ling
  Ko, Wu, Agrawal, Lin, Sheng, Sung, Breland-King, Butterfield, Gawde, Singh,
  Zhang, Apte, Shetty, Hutter, Li, Salesky, Lebron, Kanerva, Paganini, Nguyen,
  Vallu, Peter, Velury, Kao, Hoover, Bortsova, Bishop, Jakobovits, Agostini,
  Agarwal, Liu, Kwong, Tavakkol, Bica, Greve, GP, Marcus, Hou, Duerig,
  Moroshko, Lacey, Davis, Amelot, Wang, Kim, Strinopoulos, Wan, Lan, Krishnan,
  Tang, Humphreys, Bai, Shtacher, Machado, Pang, Burke, Liu, Aravamudhan, Song,
  Hirst, Singh, Jou, Bai, Piccinno, Fu, Alazard, Meiri, Winter, Chen, Zhang,
  Heitkaemper, Lambert, Lee, Frömmgen, Rogulenko, Nair, Niemczyk, Bulyenov,
  Xu, Shemtov, Zadimoghaddam, Toropov, Wirth, Dai, Gollapudi, Zheng, Kurakin,
  Lee, Bullard, Serrano, Balazevic, Li, Schalkwyk, Murphy, Zhang, Sequeira,
  Datta, Agrawal, Sutton, Attaluri, Chiang, Farhan, Thornton, Lin, Choma,
  Nguyen, Dasgupta, Robinson, Comşa, Riley, Pillai, Mustafa, Golan, Zandieh,
  Lespiau, Porter, Ross, Rajayogam, Agarwal, Venugopalan, Shahriari, Yan, Xu,
  Tobin, Dubov, Shi, Recasens, Kovsharov, Borgeaud, Dery, Vasanth, Gribovskaya,
  Qiu, Mahdieh, Skut, Nielsen, Zheng, Yu, Bostock, Gupta, Archer, Rawles,
  Davies, Svyatkovskiy, Tsai, Halpern, Reisswig, Wydrowski, Chang, Puigcerver,
  Taege, Li, Schnider, Li, Dena, Xu, Telang, Shi, Zen, Kastner, Ko,
  Subramaniam, Kumar, Blois, Dai, Wieting, Lu, Zeldes, Xie, Hauth, Ţifrea, Li,
  El-Husseini, Abolafia, Zhou, Ding, Ghalebikesabi, Guía, Maksai, Ágoston
  Weisz, Arik, Sukhanov, Świetlik, Jia, Yu, Wang, Brand, Bloxwich, Kirmani,
  Chen, Go, Sprechmann, Kannen, Carin, Sandhu, Edkins, Nooteboom, Gupta,
  Maggiore, Azizi, Pritch, Yin, Gupta, Tarlow, Smith, Ivanov, Babaeizadeh,
  Goel, Kambala, Chu, Kastelic, Liu, Soltau, Stone, Agrawal, Kim, Soparkar,
  Tadepalli, Bunyan, Soh, Kannan, Kim, Chen, Halumi, Roy, Wang, Sercinoglu,
  Gibson, Bhatnagar, Sano, von Dincklage, Ren, Mitrevski, Olšák, She,
  Doersch, Jilei, Wang, Liu, Tan, Yakar, Warkentin, Ramirez, Lebsack, Dillon,
  Mathews, Cobley, Wu, Chen, Simon, Nath, Sainath, Bendebury, Julian,
  Mankalale, Ćurko, Zacchello, Brown, Sodhia, Howard, Caelles, Gupta, Evans,
  Bulanova, Katzen, Goldenberg, Tsitsulin, Stanton, Schillings, Kovalev, Fry,
  Shah, Lin, Upadhyay, Li, Radpour, Maggioni, Xiong, Haas, Brennan, Kamath,
  Savinov, Nagrani, Yacovone, Kappedal, Andriopoulos, Lao, Li, Rozhdestvenskiy,
  Hashimoto, Audibert, Austin, Rodriguez, Ruoss, Honke, Karkhanis, Xiong, Wei,
  Huang, Leng, Premachandran, Bileschi, Evangelopoulos, Mensink, Pavagadhi,
  Teplyashin, Chang, Xue, Tanzer, Goldman, Patel, Li, Wiesner, Zheng,
  Stewart-Binks, Han, Li, Luo, Lenc, Lučić, Xue, Mullins, Guseynov, Chang,
  Galatzer-Levy, Zhang, Bingham, Hu, Hartman, Ma, Griffith, Irpan, Radebaugh,
  Yue, Fan, Ungureanu, Sorokin, Teufel, Li, Anil, Paparas, Wang, Lin, Peng,
  Shum, Petrovic, Brady, Nguyen, Macherey, Li, Singh, Yenugula, Iinuma, Chen,
  Kopparapu, Stern, Dave, Thekkath, Perot, Kumar, Li, Xiao, Bilotti, Bateni,
  Noble, Lee, Vázquez-Reina, Salazar, Yang, Wang, Gruzewska, Rao, Raghuram,
  Xu, Ben-David, Mei, Dalmia, Zhang, Liu, Bansal, Pankov, Schwarcz, Burns,
  Chan, Sanghai, Liang, Liang, He, Stuart, Narayanan, Zhu, Frank, Fatemi,
  Sabne, Lang, Bhattacharya, Settle, Wang, McMahan, Tacchetti, Soares, Hadian,
  Cabi, Chung, Putikhin, Li, Chen, Tarango, Michalewski, Kazemi, Masoom,
  Sheftel, Shivanna, Vadali, Comanescu, Reid, Moore, Neelakantan, Sander,
  Herzig, Rosenberg, Dehghani, Choi, Fink, Hayes, Ge, Weng, Ho, Karro, Krishna,
  Thiet, Skerry-Ryan, Eppens, Andreetto, Sarma, Bonacina, Ayan, Nawhal, Shan,
  Dusenberry, Thakoor, Gubbi, Nguyen, Tsarfaty, Albanie, Mitrović, Gandhi,
  Chen, Epasto, Stephanov, Jin, Gehman, Amini, Weber, Behbahani, Xu, Allamanis,
  Chen, Ott, Sha, Jastrzebski, Qi, Greene, Wu, Toki, Vlasic, Shapiro,
  Kotikalapudi, Shen, Saeki, Xie, Cassirer, Bharadwaj, Kiyono, Bhojanapalli,
  Rosenfeld, Ritter, Mao, Oliveira, Egyed, Bandemer, Parisotto, Kinoshita,
  Pluto, Maniatis, Li, Guo, Ghiasi, Tarbouriech, Chatterjee, Jin, Katrina, Xu,
  Palomaki, Arnold, Sewak, Piccinini, Sharma, Albrecht, Purser-haskell,
  Vaswani, Chen, Wisniewski, Cao, Aslanides, Phu, Sieb, Agubuzu, Zheng, Sohn,
  Selvi, Andreassen, Subudhi, Eruvbetine, Woodman, Mery, Krause, Ren, Ma, Luo,
  Chen, Fan, Griffiths, Schuler, Li, Zhang, Sarr, Luo, Patana, Watson,
  Naboulsi, Collins, Sidhwani, Hoogeboom, Silver, Caveness, Zhao, Rodriguez,
  Deines, Bai, Griffin, Tagliasacchi, Xue, Babbula, Pang, Ding, Shen, Peake,
  Crocker, Raghvendra, Swisher, Han, Singh, Wu, Pchelin, Munkhdalai, Alon,
  Bacon, Robles, Bulian, Johnson, Powell, Ferreira, Li, Benzing, Velimirović,
  Soyer, Kong, Tony, Nguyên, Yang, Liu, van Amersfoort, Gillick, Sun,
  Rauschmayr, Zhang, Zhan, Zhou, Frolov, Yang, Vnukov, Rouillard, Li, Mandhane,
  Fallen, Venkataraman, Hu, Brennan, Lee, Chang, Sundermeyer, Pan, Ke, Tong,
  Fabrikant, Bono, Gu, Foley, Mao, Delakis, Bhaswar, Frostig, Li, Zipori, Hope,
  Kozlova, Mishra, Djolonga, Schiff, Merey, Briakou, Morgan, Wan, Hassidim,
  Skerry-Ryan, Sengupta, Jasarevic, Kallakuri, Kunkle, Brennan, Lieber,
  Mansoor, Walker, Zhang, Xie, Žužić, Chukwuka, Druinsky, Cho, Yao, Naeem,
  Butt, Kim, Jia, Jordan, Lelkes, Kurzeja, Wang, Zhao, Over, Chakladar,
  Prasetya, Jha, Ganapathy, Cong, Shroff, Saroufim, Miryoosefi, Hammad, Nasir,
  Xi, Gao, Maeng, Hora, Cheng, Haghani, Lewenberg, Lu, Matysiak, Raisinghani,
  Wang, Baugher, Sukthankar, Giang, Schultz, Fiedel, Chen, Lee, Dey, Zheng,
  Paul, Smith, Ly, Wang, Bansal, Perz, Ricco, Blank, Keshava, Sharma, Chow,
  Lad, Jalan, Osindero, Swanson, Scott, Ilić, Li, Jonnalagadda, Soudagar,
  Xiong, Batsaikhan, Jarrett, Kumar, Shah, Lawlor, Waters, Graham, May, Ramos,
  Lefdal, Cankara, Cano, O'Donoghue, Borovik, Liu, Grimstad, Alnahlawi,
  Tsihlas, Hudson, Grigorev, Jia, Huang, Igwe, Lebedev, Tang, Krivokon, Garcia,
  Tan, Jia, Stys, Vashishth, Liang, Venkatraman, Gu, Kementsietsidis, Zhu,
  Jung, Bai, Hosseini, Ahmed, Gupta, Yuan, Ashraf, Nigam, Vasudevan, Awasthi,
  Gilady, Mariet, Eskander, Li, Hu, Garrido, Schlattner, Zhang, Saxena, Dević,
  Muralidharan, Murthy, Zhou, Choi, Wongpanich, Wang, Shah, Xu, Huang, Spencer,
  Chen, Cohan, Wang, Tompson, Wu, Haroun, Li, Huergo, Yang, Yin, Wendt,
  Bendersky, Chaabouni, Snaider, Ferret, Jindal, Thompson, Xue, Bishop, Phal,
  Sharma, Sung, Radhakrishnan, Shomrat, Ingle, Vij, Gilmer, Istin, Sobell, Lu,
  Nottage, Sadigh, Willcock, Zhang, Xu, Brown, Lee, Wang, Zhu, Tay, Kim,
  Gutierrez, Sharma, Xian, Seo, Cui, Pochernina, Baetu, Jastrzębski, Ly,
  Elhawaty, Suh, Sezener, Wang, Yuen, Tucker, Cai, Yang, Wang, Muzio, Qian,
  Yoo, Lockhart, McKee, Guo, Mehrotra, Mendonça, Mehta, Ben, Tekur, Mu, Zhu,
  Krakovna, Lee, Maschinot, Cevey, Choe, Bai, Srinivasan, Gasaway, Young,
  Siegler, Holtmann-Rice, Piratla, Baumli, Yogev, Hofer, van Hasselt, Grant,
  Chervonyi, Silver, Hogue, Agarwal, Wang, Singh, Flynn, Lipschultz, David,
  Bellot, Yang, Le, Graziano, Olszewska, Hui, Maurya, Parotsidis, Chen,
  Oguntebi, Kelley, Baddepudi, Mauerer, Shaw, Siegman, Yang, Shetty, Roy, Song,
  Stokowiec, Burnell, Savant, Busa-Fekete, Miao, Ghosh, MacDermed, Lippe,
  Dektiarev, Behrman, Mentzer, Nguyen, Wei, Verma, Knutsen, Dasari, Yan,
  Mitrichev, Wang, Shejwalkar, Austin, Sunkara, Potti, Virin, Wright, Liu,
  Riva, Pot, Kochanski, Le, Balasubramaniam, Dhar, Liao, Bloniarz, Shukla,
  Cole, Lee, Zhang, Kafle, Vashishtha, Mahmoudieh, Chen, Hoffmann, Srinivasan,
  Lago, Shalom, Wang, Elabd, Sharma, Oh, Kothawade, Le, Monteiro, Yang,
  Alarakyia, Geirhos, Mincu, Garnes, Kobayashi, Mariooryad, Krasowiak, Zhixin,
  Lai, Mourad, Wang, Bu, Aharoni, Chen, Goyal, Zubov, Bapna, Dabir, Kothari,
  Lamerigts, Cao, Shar, Yew, Kulkarni, Mahaarachchi, Joshi, Zhu, Lichtarge,
  Zhou, Muckenhirn, Selo, Vinyals, Chen, Brohan, Mehta, Cogan, Wang, Geri, Ko,
  Chen, Viola, Shivam, Wang, Elish, Popa, Pereira, Liu, Koster, Kim, Zhang,
  Ebrahimi, Talukdar, Zheng, Poklukar, Mikhalap, Johnson, Vijayakumar,
  Omernick, Dibb, Dubey, Hu, Suman, Aggarwal, Kornakov, Xia, Lowe, Kolganov,
  Xiao, Nikolaev, Hemingray, Li, Iljazi, Rybiński, Sandhu, Lu, Luong,
  Jenatton, Govindaraj, Hui, Li, Dulac-Arnold, Park, Wang, Modi, Pouget-Abadie,
  Greller, Gupta, Berry, Ramachandran, Xie, McCafferty, Wang, Gupta, Lim,
  Bratanič, Brock, Akolzin, Sproch, Karliner, Kim, Goedeckemeyer, Shazeer,
  Schmid, Calandriello, Bhatia, Choromanski, Montgomery, Dua, Ramalho, King,
  Gao, Nguyen, Lindner, Pitta, Johnson, Salama, Ardila, Han, Farnese, Odoom,
  Wang, Ding, Rink, Smith, Lehri, Cohen, Vats, He, Gopavarapu, Paszke, Patel,
  Gansbeke, Loher, Castro, Voitovich, von Glehn, George, Niklaus, Eaton-Rosen,
  Rakićević, Jue, Perel, Zhang, Bahat, Pouget, Xing, Huot, Shenoy, Bos,
  Coriou, Richter, Noy, Wang, Ontanon, Qin, Makarchuk, Hassabis, Li, Sharma,
  Venkatesan, Kemaev, Daniel, Huang, Shah, Ponce, Warren, Chen, Faruqui, Wu,
  Andačić, Payrits, McDuff, Hume, Cao, Tessler, Wang, Wang, Rendulic,
  Agustsson, Johnson, Lando, Howard, Padmanabhan, Daswani, Banino, Kilgore,
  Heek, Ji, Caceres, Li, Kassner, Vlaskin, Liu, Grills, Hou, Sukkerd, Cheon,
  Shetty, Markeeva, Stanczyk, Iyer, Gong, Gao, Gopalakrishnan, Blyth, Reynolds,
  Bhoopchand, Bilenko, Gharibian, Zayats, Faust, Singh, Ma, Jiao,
  Vijayanarasimhan, Aroyo, Yadav, Chakera, Kakarla, Meshram, Gregor, Botea,
  Senter, Jia, Kovacs, Sharma, Baur, Kang, He, Zhuo, Kostelac, Laish, Peng,
  O'Bryan, Kasenberg, Rao, Leurent, Zhang, Stevens, Salazar, Zhang, Lobov,
  Walker, Porter, Redshaw, Ke, Rao, Lee, Lam, Moffitt, Kim, Qiao, Koo, Dadashi,
  Song, Sundararajan, Xu, Kawamoto, Zhong, Barbu, Reddy, Verzetti, Li,
  Papamakarios, Klimczak-Plucińska, Cassin, Kavukcuoglu, Swavely, Vaucher,
  Zhao, Hemsley, Tschannen, Ge, Menghani, Yu, Ha, He, Wu, Song, Sterneck,
  Zinke, Calian, Marsden, Ruiz, Hessel, Gueta, Lee, Farris, Gupta, Li, Saleh,
  Misra, Xiao, Mendolicchio, Buttimore, Krayvanova, Nayakanti, Wiethoff, Pande,
  Mirhoseini, Lao, Liu, Hua, Chen, Malkov, Kalashnikov, Gupta, Audhkhasi, Zhai,
  Kopalle, Jain, Ofek, Meyer, Baatarsukh, Strejček, Qian, Freedman, Figueira,
  Sokolik, Bachem, Lin, Kharrat, Hidey, Xu, Duan, Li, Ersoy, Everett, Cen,
  Santamaria-Fernandez, Taubenfeld, Mackinnon, Deng, Zablotskaia, Viswanadha,
  Goel, Yates, Deng, Choy, Chen, Sinha, Mossin, Wang, Szlam, Hao, Rubenstein,
  Toksoz-Exley, Aperghis, Zhong, Ahn, Isard, Lacombe, Luisier, Anastasiou,
  Kalley, Prabhu, Dunleavy, Bijwadia, Mao-Jones, Chen, Pasumarthi, Wood,
  Dostmohamed, Hurley, Simsa, Parrish, Pajarskas, Harvey, Skopek, Kochinski,
  Rey, Rieser, Zhou, Lee, Acharya, Li, Jiang, Zhang, Gipson, Mahintorabi,
  Gelmi, Khajehnouri, Yeh, Lee, Matthey, Baker, Pham, Fu, Pak, Gupta,
  Vasconcelos, Sadovsky, Walker, Hsiao, Zochbauer, Marzoca, Velan, Zeng,
  Baechler, Driess, Jain, Huang, Tao, Maggs, Levine, Schneider, Gemzer, Petit,
  Han, Fisher, Zelle, Biles, Ie, Fadeeva, Liu, Franco, Collister, Zhang, Wang,
  Zhao, Kieliger, Shuster, Zhu, Gong, Chan, Sun, Basu, Zimmermann, Hayes,
  Bapna, Snoek, Yang, Datta, Abdallah, Kilgour, Li, Mah, Jun, Rivière,
  Karmarkar, Spalink, Huang, Gonzalez, Tran, Nowak, Palowitch, Chadwick,
  Talius, Mehta, Sellam, Fränken, Nicosia, He, Kini, Amos, Basu, Jobe, Shaw,
  Xu, Evans, Ikeda, Yan, Jin, Wang, Yadav, Labzovsky, Sampath, Ma, Schumann,
  Siddhant, Shah, Youssef, Agarwal, Dabney, Tonioni, Ambar, Li, Guyon, Li,
  Soergel, Fang, Karadzhov, Udrescu, Trinh, Raunak, Noury, Guo, Gupta,
  Finkelstein, Petek, Liang, Billock, Sun, Wood, Song, Yu, Matejovicova, Cohen,
  Andra, D'Ambrosio, Deng, Nallatamby, Songhori, Dangovski, Lampinen, Botadra,
  Hillier, Cao, Baddi, Kuncoro, Yoshino, Bhagatwala, Ranzato, Schaeffer, Liu,
  Ye, Sarvana, Nham, Kuang, Gao, Baek, Mittal, Wahid, Gergely, Ni, Feldman,
  Muir, Lamblin, Macherey, Dyer, Kilpatrick, Campos, Bhutani, Fort, Ahmad,
  Severyn, Chatziprimou, Ferludin, Dimarco, Kusupati, Heyward, Bahir, Villela,
  Millican, Marcus, Bahargam, Unlu, Roth, Wei, Gopal, Ghoshal, Lee, Lin, Lees,
  Lee, Hosseini, Fan, Neel, Wu, Altun, Cai, Piqueras, Woodward, Bissacco,
  Haykal, Bordbar, Sundaram, Hodkinson, Toyama, Polovets, Myers, Sinha,
  Levinboim, Krishnakumar, Chhaparia, Sholokhova, Gundavarapu, Jawahar,
  Qureshi, Hu, Momchev, Rahtz, Wu, S, Dhamdhere, Guo, Gupta, Eslami, Schain,
  Blokzijl, Welling, Orr, Bolelli, Perez-Nieves, Sirotenko, Prasad, Kar, Pigem,
  Terzi, Weisz, Ghosh, Mavalankar, Madeka, Daugaard, Adam, Shah, Berman, Tran,
  Baker, Andrejczuk, Chole, Raboshchuk, Mirzazadeh, Kagohara, Wu, Schallhart,
  Orlando, Wang, Rrustemi, Xiong, Liu, Vezer, Ramsden, yiin Chang, Mudgal, Li,
  Vieillard, Hoshen, Ahmad, Slone, Hua, Potikha, Rossini, Stritar, Prakash,
  Wang, Dong, Nazari, Nehoran, Tekelioglu, Li, Badola, Funkhouser, Li, Yerram,
  Ganeshan, Formoso, Langner, Shi, Li, Yamamori, Panda, Saade, Scarpati,
  Breaux, Carey, Zhou, Hsieh, Bridgers, Butryna, Gupta, Tulsyan, Woo, Eltyshev,
  Grathwohl, Parks, Benjamin, Panigrahy, Dodhia, Freitas, Sauer, Song, Alet,
  Tolins, Paduraru, Zhou, Albert, Zhang, Shu, Bansal, Nguyen, Globerson, Xiao,
  Manyika, Hennigan, Rong, Matak, Bakalov, Sharma, Sinopalnikov, Pierson,
  Roller, Brown, Gao, Fukuzawa, Ghafouri, Vassigh, Barr, Wang, Korsun, Jayaram,
  Ren, Zaman, Khan, Lunts, Deutsch, Uthus, Katz, Samsikova, Khalifa, Sethi,
  Sun, Tang, Alon, Luo, Yu, Nayyar, Petrini, Truong, Hellendoorn, Chinaev,
  Alberti, Wang, Hu, Mirrokni, Balashankar, Aharon, Mehta, Iscen, Kready,
  Manning, Mohananey, Chen, Tripathi, Wu, Petrovski, Hwang, Baeuml,
  Chandrakaladharan, Liu, Coaguila, Chen, Ma, Tafti, Tatineni, Spitz, Ye,
  Vicol, Rosca, Puigdomènech, Yahav, Ghemawat, Lin, Kirk, Nabulsi, Brin,
  Bohnet, Caluwaerts, Veerubhotla, Zheng, Dai, Petrov, Xu, Mehran, Xu,
  Zintgraf, Choi, Hombaiah, Thoppilan, Reddi, Lew, Li, Webster, Sawhney,
  Lamprou, Shakeri, Lunayach, Chen, Bagri, Salcianu, Chen, Donchev, Magister,
  Nørly, Rodrigues, Izo, Noga, Zou, Köppe, Zhou, Lee, Long, Eisenbud, Chen,
  Schenck, To, Zhong, Taropa, Truong, Levy, Martins, Zhang, Semturs, Zhang,
  Yakubovich, Moreno, McConnaughey, Lu, Redmond, Weerts, Bitton, Refice,
  Lacasse, Conmy, Tallec, Odell, Forbes-Pollard, Socala, Hoech, Kohli, Walton,
  Wang, Sazanovich, Zhu, Kapishnikov, Galt, Denton, Murdoch, Sikora, Mohamed,
  Wei, First, McConnell, Cobo, Qin, Avrahami, Balle, Watanabe, Louis, Kraft,
  Ariafar, Gu, Rives, Yoon, Rusu, Cobon-Kerr, Hahn, Luo, Yuvein, Zhu, Ahuja,
  Benenson, Kaufman, Yu, Hightower, Zhang, Ni, Hendricks, Wang, Yona, Jain,
  Barrio, Bhupatiraju, Velusamy, Dafoe, Riedel, Thomas, Yuan, Bellaiche,
  Panthaplackel, Kloboves, Jauhari, Akbulut, Davchev, Gladchenko, Madras,
  Chuklin, Hill, Yuan, Madhavan, Leonhard, Scandinaro, Chen, Niu, Douillard,
  Damoc, Onoe, Pedregosa, Bertsch, Leichner, Pagadora, Malmaud, Ponda, Twigg,
  Duzhyi, Shen, Wang, Garg, Chen, Evci, Lee, Liu, Kojima, Yamaguchi, Rajendran,
  Piergiovanni, Rajendran, Fornoni, Ibagon, Ragan, Khan, Blitzer, Bunner, Sun,
  Kosakai, Lundberg, Elue, Guu, Park, Park, Narayanaswamy, Wu, Mudigonda, Cohn,
  Mu, Kumar, Graesser, Zhang, Killam, Zhuang, Giménez, Jishi, Ley-Wild, Zhai,
  Osawa, Cedillo, Liu, Upadhyay, Sieniek, Sharma, Paine, Angelova, Addepalli,
  Parada, Majumder, Lamp, Kumar, Deng, Myaskovsky, Sabolić, Dudek, York,
  de~Chaumont~Quitry, Nie, Cattle, Gunjan, Piot, Khawaja, Bang, Wang,
  Khodadadeh, R, Rawlani, Powell, Lee, Griesser, Oh, Magalhaes, Li, Tokumine,
  Vogel, Hsu, BC, Jindal, Cohen, Yang, Yuan, de~Cesare, Bruguier, Xu, Roy,
  Jacovi, Belov, Arya, Meadowlark, Cohen-Ganor, Ye, Morris-Suzuki, Banzal,
  Song, Ponnuramu, Zhang, Scrivener, Zaiem, Rochman, Han, Ghazi, Lee, Drath,
  Suo, Girgis, Shenoy, Nguyen, Eck, Gupta, Yan, Carreira, Gulati, Sang,
  Mirylenka, Cooney, Chou, Ling, Fan, Coleman, Tubone, Kumar, Baldridge,
  Hernandez-Campos, Lazaridou, Besley, Yona, Bulut, Wellens, Pierigiovanni,
  George, Green, Han, Tao, Clark, You, Abdolmaleki, Fu, Chen, Chaugule,
  Chandorkar, Rahman, Thompson, Koanantakool, Bernico, Ren, Vlasov,
  Vassilvitskii, Kula, Liang, Kim, Huang, Ye, Lepikhin, and
  Helmholz]{comanici2025gemini25pushingfrontier}
Gheorghe Comanici, Eric Bieber, Mike Schaekermann, Ice Pasupat, Noveen
  Sachdeva, Inderjit Dhillon, Marcel Blistein, Ori Ram, Dan Zhang, Evan Rosen,
  Luke Marris, Sam Petulla, Colin Gaffney, Asaf Aharoni, Nathan Lintz,
  Tiago~Cardal Pais, Henrik Jacobsson, Idan Szpektor, Nan-Jiang Jiang, Krishna
  Haridasan, Ahmed Omran, Nikunj Saunshi, Dara Bahri, Gaurav Mishra, Eric Chu,
  Toby Boyd, Brad Hekman, Aaron Parisi, Chaoyi Zhang, Kornraphop Kawintiranon,
  Tania Bedrax-Weiss, Oliver Wang, Ya Xu, Ollie Purkiss, Uri Mendlovic, Ilaï
  Deutel, Nam Nguyen, Adam Langley, Flip Korn, Lucia Rossazza, Alexandre Ramé,
  Sagar Waghmare, Helen Miller, Nathan Byrd, Ashrith Sheshan, Raia Hadsell,
  Sangnie Bhardwaj, Pawel Janus, Tero Rissa, Dan Horgan, Alvin Abdagic, Lior
  Belenki, James Allingham, Anima Singh, Theo Guidroz, Srivatsan Srinivasan,
  Herman Schmit, Kristen Chiafullo, Andre Elisseeff, Nilpa Jha, Prateek Kolhar,
  Leonard Berrada, Frank Ding, Xiance Si, Shrestha~Basu Mallick, Franz Och,
  Sofia Erell, Eric Ni, Tejasi Latkar, Sherry Yang, Petar Sirkovic, Ziqiang
  Feng, Robert Leland, Rachel Hornung, Gang Wu, Charles Blundell, Hamidreza
  Alvari, Po-Sen Huang, Cathy Yip, Sanja Deur, Li Liu, Gabriela Surita, Pablo
  Duque, Dima Damen, Johnson Jia, Arthur Guez, Markus Mircea, Animesh Sinha,
  Alberto Magni, Paweł Stradomski, Tal Marian, Vlado Galić, Wenhu Chen,
  Hisham Husain, Achintya Singhal, Dominik Grewe, François-Xavier Aubet,
  Shuang Song, Lorenzo Blanco, Leland Rechis, Lewis Ho, Rich Munoz, Kelvin
  Zheng, Jessica Hamrick, Kevin Mather, Hagai Taitelbaum, Eliza Rutherford, Yun
  Lei, Kuangyuan Chen, Anand Shukla, Erica Moreira, Eric Doi, Berivan Isik, Nir
  Shabat, Dominika Rogozińska, Kashyap Kolipaka, Jason Chang, Eugen Vušak,
  Srinivasan Venkatachary, Shadi Noghabi, Tarun Bharti, Younghoon Jun,
  Aleksandr Zaks, Simon Green, Jeshwanth Challagundla, William Wong, Muqthar
  Mohammad, Dean Hirsch, Yong Cheng, Iftekhar Naim, Lev Proleev, Damien
  Vincent, Aayush Singh, Maxim Krikun, Dilip Krishnan, Zoubin Ghahramani, Aviel
  Atias, Rajeev Aggarwal, Christo Kirov, Dimitrios Vytiniotis, Christy Koh,
  Alexandra Chronopoulou, Pawan Dogra, Vlad-Doru Ion, Gladys Tyen, Jason Lee,
  Felix Weissenberger, Trevor Strohman, Ashwin Balakrishna, Jack Rae, Marko
  Velic, Raoul de Liedekerke, Oded Elyada, Wentao Yuan, Canoee Liu, Lior Shani,
  Sergey Kishchenko, Bea Alessio, Yandong Li, Richard Song, Sam Kwei, Orion
  Jankowski, Aneesh Pappu, Youhei Namiki, Yenai Ma, Nilesh Tripuraneni, Colin
  Cherry, Marissa Ikonomidis, Yu-Cheng Ling, Colin Ji, Beka Westberg, Auriel
  Wright, Da Yu, David Parkinson, Swaroop Ramaswamy, Jerome Connor,
  Soheil~Hassas Yeganeh, Snchit Grover, George Kenwright, Lubo Litchev, Chris
  Apps, Alex Tomala, Felix Halim, Alex Castro-Ros, Zefei Li, Anudhyan Boral,
  Pauline Sho, Michal Yarom, Eric Malmi, David Klinghoffer, Rebecca Lin, Alan
  Ansell, Pradeep~Kumar S, Shubin Zhao, Siqi Zuo, Adam Santoro, Heng-Tze Cheng,
  Solomon Demmessie, Yuchi Liu, Nicole Brichtova, Allie Culp, Nathaniel Braun,
  Dan Graur, Will Ng, Nikhil Mehta, Aaron Phillips, Patrik Sundberg, Varun
  Godbole, Fangyu Liu, Yash Katariya, David Rim, Mojtaba Seyedhosseini, Sean
  Ammirati, Jonas Valfridsson, Mahan Malihi, Timothy Knight, Andeep Toor,
  Thomas Lampe, Abe Ittycheriah, Lewis Chiang, Chak Yeung, Alexandre
  Fréchette, Jinmeng Rao, Huisheng Wang, Himanshu Srivastava, Richard Zhang,
  Rocky Rhodes, Ariel Brand, Dean Weesner, Ilya Figotin, Felix Gimeno, Rachana
  Fellinger, Pierre Marcenac, José Leal, Eyal Marcus, Victor Cotruta, Rodrigo
  Cabrera, Sheryl Luo, Dan Garrette, Vera Axelrod, Sorin Baltateanu, David
  Barker, Dongkai Chen, Horia Toma, Ben Ingram, Jason Riesa, Chinmay Kulkarni,
  Yujing Zhang, Hongbin Liu, Chao Wang, Martin Polacek, Will Wu, Kai Hui,
  Adrian~N Reyes, Yi Su, Megan Barnes, Ishaan Malhi, Anfal Siddiqui, Qixuan
  Feng, Mihai Damaschin, Daniele Pighin, Andreas Steiner, Samuel Yang,
  Ramya~Sree Boppana, Simeon Ivanov, Arun Kandoor, Aditya Shah, Asier Mujika,
  Da Huang, Christopher~A. Choquette-Choo, Mohak Patel, Tianhe Yu, Toni
  Creswell, Jerry, Liu, Catarina Barros, Yasaman Razeghi, Aurko Roy, Phil
  Culliton, Binbin Xiong, Jiaqi Pan, Thomas Strohmann, Tolly Powell, Babi Seal,
  Doug DeCarlo, Pranav Shyam, Kaan Katircioglu, Xuezhi Wang, Cassidy Hardin,
  Immanuel Odisho, Josef Broder, Oscar Chang, Arun Nair, Artem Shtefan, Maura
  O'Brien, Manu Agarwal, Sahitya Potluri, Siddharth Goyal, Amit Jhindal,
  Saksham Thakur, Yury Stuken, James Lyon, Kristina Toutanova, Fangxiaoyu Feng,
  Austin Wu, Ben Horn, Alek Wang, Alex Cullum, Gabe Taubman, Disha Shrivastava,
  Chongyang Shi, Hamish Tomlinson, Roma Patel, Tao Tu, Ada~Maksutaj Oflazer,
  Francesco Pongetti, Mingyao Yang, Adrien~Ali Taïga, Vincent Perot, Nuo~Wang
  Pierse, Feng Han, Yoel Drori, Iñaki Iturrate, Ayan Chakrabarti, Legg Yeung,
  Dave Dopson, Yi ting Chen, Apoorv Kulshreshtha, Tongfei Guo, Philip Pham, Tal
  Schuster, Junquan Chen, Alex Polozov, Jinwei Xing, Huanjie Zhou, Praneeth
  Kacham, Doron Kukliansky, Antoine Miech, Sergey Yaroshenko, Ed Chi, Sholto
  Douglas, Hongliang Fei, Mathieu Blondel, Preethi Myla, Lior Madmoni, Xing Wu,
  Daniel Keysers, Kristian Kjems, Isabela Albuquerque, Lijun Yu, Joel D'sa,
  Michelle Plantan, Vlad Ionescu, Jaume~Sanchez Elias, Abhirut Gupta,
  Manish~Reddy Vuyyuru, Fred Alcober, Tong Zhou, Kaiyang Ji, Florian Hartmann,
  Subha Puttagunta, Hugo Song, Ehsan Amid, Anca Stefanoiu, Andrew Lee, Paul
  Pucciarelli, Emma Wang, Amit Raul, Slav Petrov, Isaac Tian, Valentin Anklin,
  Nana Nti, Victor Gomes, Max Schumacher, Grace Vesom, Alex Panagopoulos,
  Konstantinos Bousmalis, Daniel Andor, Josh Jacob, Yuan Zhang, Bill Rosgen,
  Matija Kecman, Matthew Tung, Alexandra Belias, Noah Goodman, Paul Covington,
  Brian Wieder, Nikita Saxena, Elnaz Davoodi, Muhuan Huang, Sharath Maddineni,
  Vincent Roulet, Folawiyo Campbell-Ajala, Pier~Giuseppe Sessa, Xintian, Wu,
  Guangda Lai, Paul Collins, Alex Haig, Vytenis Sakenas, Xiaowei Xu, Marissa
  Giustina, Laurent~El Shafey, Pichi Charoenpanit, Shefali Garg, Joshua
  Ainslie, Boone Severson, Montse~Gonzalez Arenas, Shreya Pathak, Sujee
  Rajayogam, Jie Feng, Michiel Bakker, Sheng Li, Nevan Wichers, Jamie Rogers,
  Xinyang Geng, Yeqing Li, Rolf Jagerman, Chao Jia, Nadav Olmert, David Sharon,
  Matthew Mauger, Sandeep Mariserla, Hongxu Ma, Megha Mohabey, Kyuyeun Kim,
  Alek Andreev, Scott Pollom, Juliette Love, Vihan Jain, Priyanka Agrawal,
  Yannick Schroecker, Alisa Fortin, Manfred Warmuth, Ji Liu, Andrew Leach,
  Irina Blok, Ganesh~Poomal Girirajan, Roee Aharoni, Benigno Uria, Andrei
  Sozanschi, Dan Goldberg, Lucian Ionita, Marco~Tulio Ribeiro, Martin Zlocha,
  Vighnesh Birodkar, Sami Lachgar, Liangzhe Yuan, Himadri Choudhury, Matt
  Ginsberg, Fei Zheng, Gregory Dibb, Emily Graves, Swachhand Lokhande, Gabriel
  Rasskin, George-Cristian Muraru, Corbin Quick, Sandeep Tata, Pierre Sermanet,
  Aditya Chawla, Itay Karo, Yan Wang, Susan Zhang, Orgad Keller, Anca Dragan,
  Guolong Su, Ian Chou, Xi Liu, Yiqing Tao, Shruthi Prabhakara, Marc Wilson,
  Ruibo Liu, Shibo Wang, Georgie Evans, David Du, Alfonso Castaño, Gautam
  Prasad, Mona~El Mahdy, Sebastian Gerlach, Machel Reid, Jarrod Kahn, Amir
  Zait, Thanumalayan~Sankaranarayana Pillai, Thatcher Ulrich, Guanyu Wang, Jan
  Wassenberg, Efrat Farkash, Kiran Yalasangi, Congchao Wang, Maria Bauza, Simon
  Bucher, Ting Liu, Jun Yan, Gary Leung, Vikas Sindhwani, Parker Barnes, Avi
  Singh, Ivan Jurin, Jichuan Chang, Niket~Kumar Bhumihar, Sivan Eiger, Gui
  Citovsky, Ben Withbroe, Zhang Li, Siyang Xue, Niccolò~Dal Santo, Georgi
  Stoyanov, Yves Raimond, Steven Zheng, Yilin Gao, Vít Listík, Sławek
  Kwasiborski, Rachel Saputro, Adnan Ozturel, Ganesh Mallya, Kushal Majmundar,
  Ross West, Paul Caron, Jinliang Wei, Lluis Castrejon, Sharad Vikram, Deepak
  Ramachandran, Nikhil Dhawan, Jiho Park, Sara Smoot, George van~den Driessche,
  Yochai Blau, Chase Malik, Wei Liang, Roy Hirsch, Cicero~Nogueira dos Santos,
  Eugene Weinstein, Aäron van~den Oord, Sid Lall, Nicholas FitzGerald, Zixuan
  Jiang, Xuan Yang, Dale Webster, Ali Elqursh, Aedan Pope, Georges Rotival,
  David Raposo, Wanzheng Zhu, Jeff Dean, Sami Alabed, Dustin Tran, Arushi
  Gupta, Zach Gleicher, Jessica Austin, Edouard Rosseel, Megh Umekar, Dipanjan
  Das, Yinghao Sun, Kai Chen, Karolis Misiunas, Xiang Zhou, Yixian Di, Alyssa
  Loo, Josh Newlan, Bo Li, Vinay Ramasesh, Ying Xu, Alex Chen, Sudeep Gandhe,
  Radu Soricut, Nikita Gupta, Shuguang Hu, Seliem El-Sayed, Xavier Garcia, Idan
  Brusilovsky, Pu-Chin Chen, Andrew Bolt, Lu Huang, Alex Gurney, Zhiying Zhang,
  Alexander Pritzel, Jarek Wilkiewicz, Bryan Seybold, Bhargav~Kanagal Shamanna,
  Felix Fischer, Josef Dean, Karan Gill, Ross Mcilroy, Abhishek Bhowmick,
  Jeremy Selier, Antoine Yang, Derek Cheng, Vladimir Magay, Jie Tan, Dhriti
  Varma, Christian Walder, Tomas Kocisky, Ryo Nakashima, Paul Natsev, Mike
  Kwong, Ionel Gog, Chiyuan Zhang, Sander Dieleman, Thomas Jimma, Andrey
  Ryabtsev, Siddhartha Brahma, David Steiner, Dayou Du, Ante Žužul, Mislav
  Žanić, Mukund Raghavachari, Willi Gierke, Zeyu Zheng, Dessie Petrova, Yann
  Dauphin, Yuchuan Liu, Ido Kessler, Steven Hand, Chris Duvarney, Seokhwan Kim,
  Hyo Lee, Léonard Hussenot, Jeffrey Hui, Josh Smith, Deepali Jain, Jiawei
  Xia, Gaurav~Singh Tomar, Keyvan Amiri, Du Phan, Fabian Fuchs, Tobias Weyand,
  Nenad Tomasev, Alexandra Cordell, Xin Liu, Jonathan Mallinson, Pankaj Joshi,
  Andy Crawford, Arun Suggala, Steve Chien, Nick Fernando, Mariella
  Sanchez-Vargas, Duncan Williams, Phil Crone, Xiyang Luo, Igor Karpov, Jyn
  Shan, Terry Thurk, Robin Strudel, Paul Voigtlaender, Piyush Patil, Tim Dozat,
  Ali Khodaei, Sahil Singla, Piotr Ambroszczyk, Qiyin Wu, Yifan Chang, Brian
  Roark, Chaitra Hegde, Tianli Ding, Angelos Filos, Zhongru Wu, André~Susano
  Pinto, Shuang Liu, Saarthak Khanna, Aditya Pandey, Siobhan Mcloughlin, Qiujia
  Li, Sam Haves, Allan Zhou, Elena Buchatskaya, Isabel Leal, Peter de Boursac,
  Nami Akazawa, Nina Anderson, Terry Chen, Krishna Somandepalli, Chen Liang,
  Sheela Goenka, Stephanie Winkler, Alexander Grushetsky, Yifan Ding, Jamie
  Smith, Fan Ye, Jordi Pont-Tuset, Eric Li, Ruichao Li, Tomer Golany, Dawid
  Wegner, Tao Jiang, Omer Barak, Yuan Shangguan, Eszter Vértes, Renee Wong,
  Jörg Bornschein, Alex Tudor, Michele Bevilacqua, Tom Schaul, Ankit~Singh
  Rawat, Yang Zhao, Kyriakos Axiotis, Lei Meng, Cory McLean, Jonathan Lai,
  Jennifer Beattie, Nate Kushman, Yaxin Liu, Blair Kutzman, Fiona Lang,
  Jingchen Ye, Praneeth Netrapalli, Pushkar Mishra, Myriam Khan, Megha Goel,
  Rob Willoughby, David Tian, Honglei Zhuang, JD Chen, Zak Tsai, Tasos
  Kementsietsidis, Arjun Khare, James Keeling, Keyang Xu, Nathan Waters,
  Florent Altché, Ashok Popat, Bhavishya Mittal, David Saxton, Dalia~El
  Badawy, Michael Mathieu, Zheng Zheng, Hao Zhou, Nishant Ranka, Richard Shin,
  Qingnan Duan, Tim Salimans, Ioana Mihailescu, Uri Shaham, Ming-Wei Chang,
  Yannis Assael, Nishanth Dikkala, Martin Izzard, Vincent Cohen-Addad, Cat
  Graves, Vlad Feinberg, Grace Chung, DJ Strouse, Danny Karmon, Sahand
  Sharifzadeh, Zoe Ashwood, Khiem Pham, Jon Blanton, Alex Vasiloff, Jarred
  Barber, Mark Geller, Aurick Zhou, Fedir Zubach, Tzu-Kuo Huang, Lei Zhang,
  Himanshu Gupta, Matt Young, Julia Proskurnia, Ronny Votel, Valentin Gabeur,
  Gabriel Barcik, Aditya Tripathi, Hongkun Yu, Geng Yan, Beer Changpinyo, Filip
  Pavetić, Amy Coyle, Yasuhisa Fujii, Jorge~Gonzalez Mendez, Tianhao Zhou,
  Harish Rajamani, Blake Hechtman, Eddie Cao, Da-Cheng Juan, Yi-Xuan Tan,
  Valentin Dalibard, Yilun Du, Natalie Clay, Kaisheng Yao, Wenhao Jia, Dimple
  Vijaykumar, Yuxiang Zhou, Xinyi Bai, Wei-Chih Hung, Steven Pecht, Georgi
  Todorov, Nikhil Khadke, Pramod Gupta, Preethi Lahoti, Arnaud Autef, Karthik
  Duddu, James Lee-Thorp, Alexander Bykovsky, Tautvydas Misiunas, Sebastian
  Flennerhag, Santhosh Thangaraj, Jed McGiffin, Zack Nado, Markus Kunesch,
  Andreas Noever, Amir Hertz, Marco Liang, Victor Stone, Evan Palmer, Samira
  Daruki, Arijit Pramanik, Siim Põder, Austin Kyker, Mina Khan, Evgeny
  Sluzhaev, Marvin Ritter, Avraham Ruderman, Wenlei Zhou, Chirag Nagpal, Kiran
  Vodrahalli, George Necula, Paul Barham, Ellie Pavlick, Jay Hartford, Izhak
  Shafran, Long Zhao, Maciej Mikuła, Tom Eccles, Hidetoshi Shimokawa, Kanav
  Garg, Luke Vilnis, Hanwen Chen, Ilia Shumailov, Kuang-Huei Lee, Abdelrahman
  Abdelhamed, Meiyan Xie, Vered Cohen, Ester Hlavnova, Dan Malkin, Chawin
  Sitawarin, James Lottes, Pauline Coquinot, Tianli Yu, Sandeep Kumar, Jingwei
  Zhang, Aroma Mahendru, Zafarali Ahmed, James Martens, Tao Chen, Aviel Boag,
  Daiyi Peng, Coline Devin, Arseniy Klimovskiy, Mary Phuong, Danny Vainstein,
  Jin Xie, Bhuvana Ramabhadran, Nathan Howard, Xinxin Yu, Gitartha Goswami,
  Jingyu Cui, Sam Shleifer, Mario Pinto, Chih-Kuan Yeh, Ming-Hsuan Yang, Sara
  Javanmardi, Dan Ethier, Chace Lee, Jordi Orbay, Suyog Kotecha, Carla
  Bromberg, Pete Shaw, James Thornton, Adi~Gerzi Rosenthal, Shane Gu, Matt
  Thomas, Ian Gemp, Aditya Ayyar, Asahi Ushio, Aarush Selvan, Joel Wee, Chenxi
  Liu, Maryam Majzoubi, Weiren Yu, Jake Abernethy, Tyler Liechty, Renke Pan,
  Hoang Nguyen, Qiong, Hu, Sarah Perrin, Abhinav Arora, Emily Pitler, Weiyi
  Wang, Kaushik Shivakumar, Flavien Prost, Ben Limonchik, Jing Wang, Yi Gao,
  Timothee Cour, Shyamal Buch, Huan Gui, Maria Ivanova, Philipp Neubeck, Kelvin
  Chan, Lucy Kim, Huizhong Chen, Naman Goyal, Da-Woon Chung, Lu Liu, Yao Su,
  Anastasia Petrushkina, Jiajun Shen, Armand Joulin, Yuanzhong Xu, Stein~Xudong
  Lin, Yana Kulizhskaya, Ciprian Chelba, Shobha Vasudevan, Eli Collins,
  Vasilisa Bashlovkina, Tony Lu, Doug Fritz, Jongbin Park, Yanqi Zhou, Chen Su,
  Richard Tanburn, Mikhail Sushkov, Mitchelle Rasquinha, Jinning Li, Jennifer
  Prendki, Yiming Li, Pallavi LV, Shriya Sharma, Hen Fitoussi, Hui Huang,
  Andrew Dai, Phuong Dao, Mike Burrows, Henry Prior, Danfeng Qin, Golan Pundak,
  Lars~Lowe Sjoesund, Art Khurshudov, Zhenkai Zhu, Albert Webson, Elizabeth
  Kemp, Tat Tan, Saurabh Agrawal, Susie Sargsyan, Liqun Cheng, Jim Stephan, Tom
  Kwiatkowski, David Reid, Arunkumar Byravan, Assaf~Hurwitz Michaely, Nicolas
  Heess, Luowei Zhou, Sonam Goenka, Viral Carpenter, Anselm Levskaya, Bo Wang,
  Reed Roberts, Rémi Leblond, Sharat Chikkerur, Stav Ginzburg, Max Chang,
  Robert Riachi, Chuqiao, Xu, Zalán Borsos, Michael Pliskin, Julia Pawar,
  Morgane Lustman, Hannah Kirkwood, Ankit Anand, Aditi Chaudhary, Norbert Kalb,
  Kieran Milan, Sean Augenstein, Anna Goldie, Laurel Prince, Karthik Raman,
  Yanhua Sun, Vivian Xia, Aaron Cohen, Zhouyuan Huo, Josh Camp, Seher Ellis,
  Lukas Zilka, David~Vilar Torres, Lisa Patel, Sho Arora, Betty Chan, Jonas
  Adler, Kareem Ayoub, Jacky Liang, Fayaz Jamil, Jiepu Jiang, Simon
  Baumgartner, Haitian Sun, Yael Karov, Yaroslav Akulov, Hui Zheng, Irene Cai,
  Claudio Fantacci, James Rubin, Alex~Rav Acha, Mengchao Wang, Nina D'Souza,
  Rohit Sathyanarayana, Shengyang Dai, Simon Rowe, Andrey Simanovsky, Omer
  Goldman, Yuheng Kuang, Xiaoyue Pan, Andrew Rosenberg, Tania Rojas-Esponda,
  Praneet Dutta, Amy Zeng, Irina Jurenka, Greg Farquhar, Yamini Bansal, Shariq
  Iqbal, Becca Roelofs, Ga-Young Joung, Parker Beak, Changwan Ryu, Ryan Poplin,
  Yan Wu, Jean-Baptiste Alayrac, Senaka Buthpitiya, Olaf Ronneberger, Caleb
  Habtegebriel, Wei Li, Paul Cavallaro, Aurora Wei, Guy Bensky, Timo Denk,
  Harish Ganapathy, Jeff Stanway, Pratik Joshi, Francesco Bertolini, Jessica
  Lo, Olivia Ma, Zachary Charles, Geta Sampemane, Himanshu Sahni, Xu Chen,
  Harry Askham, David Gaddy, Peter Young, Jiewen Tan, Matan Eyal, Arthur
  Bražinskas, Li Zhong, Zhichun Wu, Mark Epstein, Kai Bailey, Andrew Hard,
  Kamyu Lee, Sasha Goldshtein, Alex Ruiz, Mohammed Badawi, Matthias
  Lochbrunner, JK Kearns, Ashley Brown, Fabio Pardo, Theophane Weber, Haichuan
  Yang, Pan-Pan Jiang, Berkin Akin, Zhao Fu, Marcus Wainwright, Chi Zou, Meenu
  Gaba, Pierre-Antoine Manzagol, Wendy Kan, Yang Song, Karina Zainullina, Rui
  Lin, Jeongwoo Ko, Salil Deshmukh, Apoorv Jindal, James Svensson, Divya Tyam,
  Heri Zhao, Christine Kaeser-Chen, Scott Baird, Pooya Moradi, Jamie Hall,
  Qiuchen Guo, Vincent Tsang, Bowen Liang, Fernando Pereira, Suhas Ganesh, Ivan
  Korotkov, Jakub Adamek, Sridhar Thiagarajan, Vinh Tran, Charles Chen, Chris
  Tar, Sanil Jain, Ishita Dasgupta, Taylan Bilal, David Reitter, Kai Zhao,
  Giulia Vezzani, Yasmin Gehman, Pulkit Mehta, Lauren Beltrone, Xerxes
  Dotiwalla, Sergio Guadarrama, Zaheer Abbas, Stefani Karp, Petko Georgiev,
  Chun-Sung Ferng, Marc Brockschmidt, Liqian Peng, Christoph Hirnschall, Vikas
  Verma, Yingying Bi, Ying Xiao, Avigail Dabush, Kelvin Xu, Phil Wallis,
  Randall Parker, Qifei Wang, Yang Xu, Ilkin Safarli, Dinesh Tewari, Yin Zhang,
  Seungyeon Kim, Andrea Gesmundo, Mackenzie Thomas, Sergey Levi, Ahmed
  Chowdhury, Kanishka Rao, Peter Garst, Sam Conway-Rahman, Helen Ran, Kay
  McKinney, Zhisheng Xiao, Wenhao Yu, Rohan Agrawal, Axel Stjerngren, Catalin
  Ionescu, Jingjing Chen, Vivek Sharma, Justin Chiu, Fei Liu, Ken Franko,
  Clayton Sanford, Xingyu Cai, Paul Michel, Sanjay Ganapathy, Jane Labanowski,
  Zachary Garrett, Ben Vargas, Sean Sun, Bryan Gale, Thomas Buschmann,
  Guillaume Desjardins, Nimesh Ghelani, Palak Jain, Mudit Verma, Chulayuth
  Asawaroengchai, Julian Eisenschlos, Jitendra Harlalka, Hideto Kazawa, Don
  Metzler, Joshua Howland, Ying Jian, Jake Ades, Viral Shah, Tynan Gangwani,
  Seungji Lee, Roman Ring, Steven~M. Hernandez, Dean Reich, Amer Sinha,
  Ashutosh Sathe, Joe Kovac, Ashleah Gill, Ajay Kannan, Andrea D'olimpio,
  Martin Sevenich, Jay Whang, Been Kim, Khe~Chai Sim, Jilin Chen, Jiageng
  Zhang, Shuba Lall, Yossi Matias, Bill Jia, Abe Friesen, Sara Nasso, Ashish
  Thapliyal, Bryan Perozzi, Ting Yu, Anna Shekhawat, Safeen Huda, Peter
  Grabowski, Eric Wang, Ashwin Sreevatsa, Hilal Dib, Mehadi Hassen, Parker
  Schuh, Vedrana Milutinovic, Chris Welty, Michael Quinn, Ali Shah, Bangju
  Wang, Gabe Barth-Maron, Justin Frye, Natalie Axelsson, Tao Zhu, Yukun Ma,
  Irene Giannoumis, Hanie Sedghi, Chang Ye, Yi Luan, Kevin Aydin, Bilva
  Chandra, Vivek Sampathkumar, Ronny Huang, Victor Lavrenko, Ahmed Eleryan, Zhi
  Hong, Steven Hansen, Sara~Mc Carthy, Bidisha Samanta, Domagoj Ćevid, Xin
  Wang, Fangtao Li, Michael Voznesensky, Matt Hoffman, Andreas Terzis, Vikash
  Sehwag, Gil Fidel, Luheng He, Mu Cai, Yanzhang He, Alex Feng, Martin
  Nikoltchev, Samrat Phatale, Jason Chase, Rory Lawton, Ming Zhang, Tom Ouyang,
  Manuel Tragut, Mehdi~Hafezi Manshadi, Arjun Narayanan, Jiaming Shen, Xu Gao,
  Tolga Bolukbasi, Nick Roy, Xin Li, Daniel Golovin, Liviu Panait, Zhen Qin,
  Guangxing Han, Thomas Anthony, Sneha Kudugunta, Viorica Patraucean, Aniket
  Ray, Xinyun Chen, Xiaochen Yang, Tanuj Bhatia, Pranav Talluri, Alex Morris,
  Andrija Ražnatović, Bethanie Brownfield, James An, Sheng Peng, Patrick
  Kane, Ce Zheng, Nico Duduta, Joshua Kessinger, James Noraky, Siqi Liu, Keran
  Rong, Petar Veličković, Keith Rush, Alex Goldin, Fanny Wei, Shiva
  Mohan~Reddy Garlapati, Caroline Pantofaru, Okwan Kwon, Jianmo Ni, Eric
  Noland, Julia~Di Trapani, Françoise Beaufays, Abhijit~Guha Roy, Yinlam Chow,
  Aybuke Turker, Geoffrey Cideron, Lantao Mei, Jon Clark, Qingyun Dou, Matko
  Bošnjak, Ralph Leith, Yuqing Du, Amir Yazdanbakhsh, Milad Nasr, Chester
  Kwak, Suraj~Satishkumar Sheth, Alex Kaskasoli, Ankesh Anand, Balaji
  Lakshminarayanan, Sammy Jerome, David Bieber, Chun-Te Chu, Alexandre Senges,
  Tianxiao Shen, Mukund Sridhar, Ndaba Ndebele, Benjamin Beyret, Shakir
  Mohamed, Mia Chen, Markus Freitag, Jiaxian Guo, Luyang Liu, Paul Roit, Heng
  Chen, Shen Yan, Tom Stone, JD Co-Reyes, Jeremy Cole, Salvatore Scellato,
  Shekoofeh Azizi, Hadi Hashemi, Alicia Jin, Anand Iyer, Marcella Valentine,
  András György, Arun Ahuja, Daniel~Hernandez Diaz, Chen-Yu Lee, Nathan
  Clement, Weize Kong, Drew Garmon, Ishaan Watts, Kush Bhatia, Khyatti Gupta,
  Matt Miecnikowski, Hugo Vallet, Ankur Taly, Edward Loper, Saket Joshi, James
  Atwood, Jo Chick, Mark Collier, Fotis Iliopoulos, Ryan Trostle, Beliz Gunel,
  Ramiro Leal-Cavazos, Arnar~Mar Hrafnkelsson, Michael Guzman, Xiaoen Ju, Andy
  Forbes, Jesse Emond, Kushal Chauhan, Ben Caine, Li Xiao, Wenjun Zeng,
  Alexandre Moufarek, Daniel Murphy, Maya Meng, Nitish Gupta, Felix Riedel,
  Anil Das, Elijah Lawal, Shashi Narayan, Tiberiu Sosea, James Swirhun, Linda
  Friso, Behnam Neyshabur, Jing Lu, Sertan Girgin, Michael Wunder, Edouard
  Yvinec, Aroonalok Pyne, Victor Carbune, Shruti Rijhwani, Yang Guo, Tulsee
  Doshi, Anton Briukhov, Max Bain, Ayal Hitron, Xuanhui Wang, Ashish Gupta, Ke
  Chen, Cosmo Du, Weiyang Zhang, Dhruv Shah, Arjun Akula, Max Dylla, Ashyana
  Kachra, Weicheng Kuo, Tingting Zou, Lily Wang, Luyao Xu, Jifan Zhu, Justin
  Snyder, Sachit Menon, Orhan Firat, Igor Mordatch, Yuan Yuan, Natalia
  Ponomareva, Rory Blevins, Lawrence Moore, Weijun Wang, Phil Chen, Martin
  Scholz, Artur Dwornik, Jason Lin, Sicheng Li, Diego Antognini, Te I, Xiaodan
  Song, Matt Miller, Uday Kalra, Adam Raveret, Oscar Akerlund, Felix Wu, Andrew
  Nystrom, Namrata Godbole, Tianqi Liu, Hannah DeBalsi, Jewel Zhao, Buhuang
  Liu, Avi Caciularu, Lauren Lax, Urvashi Khandelwal, Victoria Langston, Eric
  Bailey, Silvio Lattanzi, Yufei Wang, Neel Kovelamudi, Sneha Mondal, Guru
  Guruganesh, Nan Hua, Ofir Roval, Paweł Wesołowski, Rishikesh Ingale,
  Jonathan Halcrow, Tim Sohn, Christof Angermueller, Bahram Raad, Eli
  Stickgold, Eva Lu, Alec Kosik, Jing Xie, Timothy Lillicrap, Austin Huang,
  Lydia~Lihui Zhang, Dominik Paulus, Clement Farabet, Alex Wertheim, Bing Wang,
  Rishabh Joshi, Chu ling Ko, Yonghui Wu, Shubham Agrawal, Lily Lin, XiangHai
  Sheng, Peter Sung, Tyler Breland-King, Christina Butterfield, Swapnil Gawde,
  Sumeet Singh, Qiao Zhang, Raj Apte, Shilpa Shetty, Adrian Hutter, Tao Li,
  Elizabeth Salesky, Federico Lebron, Jonni Kanerva, Michela Paganini, Arthur
  Nguyen, Rohith Vallu, Jan-Thorsten Peter, Sarmishta Velury, David Kao, Jay
  Hoover, Anna Bortsova, Colton Bishop, Shoshana Jakobovits, Alessandro
  Agostini, Alekh Agarwal, Chang Liu, Charles Kwong, Sasan Tavakkol, Ioana
  Bica, Alex Greve, Anirudh GP, Jake Marcus, Le Hou, Tom Duerig, Rivka
  Moroshko, Dave Lacey, Andy Davis, Julien Amelot, Guohui Wang, Frank Kim,
  Theofilos Strinopoulos, Hui Wan, Charline~Le Lan, Shankar Krishnan, Haotian
  Tang, Peter Humphreys, Junwen Bai, Idan~Heimlich Shtacher, Diego Machado,
  Chenxi Pang, Ken Burke, Dangyi Liu, Renga Aravamudhan, Yue Song, Ed Hirst,
  Abhimanyu Singh, Brendan Jou, Liang Bai, Francesco Piccinno, Chuyuan~Kelly
  Fu, Robin Alazard, Barak Meiri, Daniel Winter, Charlie Chen, Mingda Zhang,
  Jens Heitkaemper, John Lambert, Jinhyuk Lee, Alexander Frömmgen, Sergey
  Rogulenko, Pranav Nair, Paul Niemczyk, Anton Bulyenov, Bibo Xu, Hadar
  Shemtov, Morteza Zadimoghaddam, Serge Toropov, Mateo Wirth, Hanjun Dai,
  Sreenivas Gollapudi, Daniel Zheng, Alex Kurakin, Chansoo Lee, Kalesha
  Bullard, Nicolas Serrano, Ivana Balazevic, Yang Li, Johan Schalkwyk, Mark
  Murphy, Mingyang Zhang, Kevin Sequeira, Romina Datta, Nishant Agrawal,
  Charles Sutton, Nithya Attaluri, Mencher Chiang, Wael Farhan, Gregory
  Thornton, Kate Lin, Travis Choma, Hung Nguyen, Kingshuk Dasgupta, Dirk
  Robinson, Iulia Comşa, Michael Riley, Arjun Pillai, Basil Mustafa, Ben
  Golan, Amir Zandieh, Jean-Baptiste Lespiau, Billy Porter, David Ross,
  Sujeevan Rajayogam, Mohit Agarwal, Subhashini Venugopalan, Bobak Shahriari,
  Qiqi Yan, Hao Xu, Taylor Tobin, Pavel Dubov, Hongzhi Shi, Adrià Recasens,
  Anton Kovsharov, Sebastian Borgeaud, Lucio Dery, Shanthal Vasanth, Elena
  Gribovskaya, Linhai Qiu, Mahdis Mahdieh, Wojtek Skut, Elizabeth Nielsen, CJ
  Zheng, Adams Yu, Carrie~Grimes Bostock, Shaleen Gupta, Aaron Archer, Chris
  Rawles, Elinor Davies, Alexey Svyatkovskiy, Tomy Tsai, Yoni Halpern,
  Christian Reisswig, Bartek Wydrowski, Bo Chang, Joan Puigcerver, Mor~Hazan
  Taege, Jian Li, Eva Schnider, Xinjian Li, Dragos Dena, Yunhan Xu, Umesh
  Telang, Tianze Shi, Heiga Zen, Kyle Kastner, Yeongil Ko, Neesha Subramaniam,
  Aviral Kumar, Pete Blois, Zhuyun Dai, John Wieting, Yifeng Lu, Yoel Zeldes,
  Tian Xie, Anja Hauth, Alexandru Ţifrea, Yuqi Li, Sam El-Husseini, Dan
  Abolafia, Howard Zhou, Wen Ding, Sahra Ghalebikesabi, Carlos Guía, Andrii
  Maksai, Ágoston Weisz, Sercan Arik, Nick Sukhanov, Aga Świetlik, Xuhui Jia,
  Luo Yu, Weiyue Wang, Mark Brand, Dawn Bloxwich, Sean Kirmani, Zhe Chen, Alec
  Go, Pablo Sprechmann, Nithish Kannen, Alen Carin, Paramjit Sandhu, Isabel
  Edkins, Leslie Nooteboom, Jai Gupta, Loren Maggiore, Javad Azizi, Yael
  Pritch, Pengcheng Yin, Mansi Gupta, Danny Tarlow, Duncan Smith, Desi Ivanov,
  Mohammad Babaeizadeh, Ankita Goel, Satish Kambala, Grace Chu, Matej Kastelic,
  Michelle Liu, Hagen Soltau, Austin Stone, Shivani Agrawal, Min Kim, Kedar
  Soparkar, Srinivas Tadepalli, Oskar Bunyan, Rachel Soh, Arvind Kannan, DY
  Kim, Blake~JianHang Chen, Afief Halumi, Sudeshna Roy, Yulong Wang, Olcan
  Sercinoglu, Gena Gibson, Sijal Bhatnagar, Motoki Sano, Daniel von Dincklage,
  Qingchun Ren, Blagoj Mitrevski, Mirek Olšák, Jennifer She, Carl Doersch,
  Jilei, Wang, Bingyuan Liu, Qijun Tan, Tamar Yakar, Tris Warkentin, Alex
  Ramirez, Carl Lebsack, Josh Dillon, Rajiv Mathews, Tom Cobley, Zelin Wu,
  Zhuoyuan Chen, Jon Simon, Swaroop Nath, Tara Sainath, Alexei Bendebury, Ryan
  Julian, Bharath Mankalale, Daria Ćurko, Paulo Zacchello, Adam~R. Brown,
  Kiranbir Sodhia, Heidi Howard, Sergi Caelles, Abhinav Gupta, Gareth Evans,
  Anna Bulanova, Lesley Katzen, Roman Goldenberg, Anton Tsitsulin, Joe Stanton,
  Benoit Schillings, Vitaly Kovalev, Corey Fry, Rushin Shah, Kuo Lin, Shyam
  Upadhyay, Cheng Li, Soroush Radpour, Marcello Maggioni, Jing Xiong, Lukas
  Haas, Jenny Brennan, Aishwarya Kamath, Nikolay Savinov, Arsha Nagrani, Trevor
  Yacovone, Ryan Kappedal, Kostas Andriopoulos, Li Lao, YaGuang Li, Grigory
  Rozhdestvenskiy, Kazuma Hashimoto, Andrew Audibert, Sophia Austin, Daniel
  Rodriguez, Anian Ruoss, Garrett Honke, Deep Karkhanis, Xi Xiong, Qing Wei,
  James Huang, Zhaoqi Leng, Vittal Premachandran, Stan Bileschi, Georgios
  Evangelopoulos, Thomas Mensink, Jay Pavagadhi, Denis Teplyashin, Paul Chang,
  Linting Xue, Garrett Tanzer, Sally Goldman, Kaushal Patel, Shixin Li, Jeremy
  Wiesner, Ivy Zheng, Ian Stewart-Binks, Jie Han, Zhi Li, Liangchen Luo, Karel
  Lenc, Mario Lučić, Fuzhao Xue, Ryan Mullins, Alexey Guseynov, Chung-Ching
  Chang, Isaac Galatzer-Levy, Adam Zhang, Garrett Bingham, Grace Hu, Ale
  Hartman, Yue Ma, Jordan Griffith, Alex Irpan, Carey Radebaugh, Summer Yue,
  Lijie Fan, Victor Ungureanu, Christina Sorokin, Hannah Teufel, Peiran Li,
  Rohan Anil, Dimitris Paparas, Todd Wang, Chu-Cheng Lin, Hui Peng, Megan Shum,
  Goran Petrovic, Demetra Brady, Richard Nguyen, Klaus Macherey, Zhihao Li,
  Harman Singh, Madhavi Yenugula, Mariko Iinuma, Xinyi Chen, Kavya Kopparapu,
  Alexey Stern, Shachi Dave, Chandu Thekkath, Florence Perot, Anurag Kumar,
  Fangda Li, Yang Xiao, Matthew Bilotti, Mohammad~Hossein Bateni, Isaac Noble,
  Lisa Lee, Amelio Vázquez-Reina, Julian Salazar, Xiaomeng Yang, Boyu Wang,
  Ela Gruzewska, Anand Rao, Sindhu Raghuram, Zheng Xu, Eyal Ben-David, Jieru
  Mei, Sid Dalmia, Zhaoyi Zhang, Yuchen Liu, Gagan Bansal, Helena Pankov,
  Steven Schwarcz, Andrea Burns, Christine Chan, Sumit Sanghai, Ricky Liang,
  Ethan Liang, Antoine He, Amy Stuart, Arun Narayanan, Yukun Zhu, Christian
  Frank, Bahar Fatemi, Amit Sabne, Oran Lang, Indro Bhattacharya, Shane Settle,
  Maria Wang, Brendan McMahan, Andrea Tacchetti, Livio~Baldini Soares, Majid
  Hadian, Serkan Cabi, Timothy Chung, Nikita Putikhin, Gang Li, Jeremy Chen,
  Austin Tarango, Henryk Michalewski, Mehran Kazemi, Hussain Masoom, Hila
  Sheftel, Rakesh Shivanna, Archita Vadali, Ramona Comanescu, Doug Reid, Joss
  Moore, Arvind Neelakantan, Michaël Sander, Jonathan Herzig, Aviv Rosenberg,
  Mostafa Dehghani, JD Choi, Michael Fink, Reid Hayes, Eric Ge, Shitao Weng,
  Chia-Hua Ho, John Karro, Kalpesh Krishna, Lam~Nguyen Thiet, Amy Skerry-Ryan,
  Daniel Eppens, Marco Andreetto, Navin Sarma, Silvano Bonacina, Burcu~Karagol
  Ayan, Megha Nawhal, Zhihao Shan, Mike Dusenberry, Shantanu Thakoor, Sagar
  Gubbi, Duc~Dung Nguyen, Reut Tsarfaty, Samuel Albanie, Jovana Mitrović, Meet
  Gandhi, Bo-Juen Chen, Alessandro Epasto, Georgi Stephanov, Ye Jin, Samuel
  Gehman, Aida Amini, Jack Weber, Feryal Behbahani, Shawn Xu, Miltos Allamanis,
  Xi Chen, Myle Ott, Claire Sha, Michal Jastrzebski, Hang Qi, David Greene,
  Xinyi Wu, Abodunrinwa Toki, Daniel Vlasic, Jane Shapiro, Ragha Kotikalapudi,
  Zhe Shen, Takaaki Saeki, Sirui Xie, Albin Cassirer, Shikhar Bharadwaj,
  Tatsuya Kiyono, Srinadh Bhojanapalli, Elan Rosenfeld, Sam Ritter, Jieming
  Mao, João~Gabriel Oliveira, Zoltan Egyed, Bernd Bandemer, Emilio Parisotto,
  Keisuke Kinoshita, Juliette Pluto, Petros Maniatis, Steve Li, Yaohui Guo,
  Golnaz Ghiasi, Jean Tarbouriech, Srimon Chatterjee, Julie Jin, Katrina, Xu,
  Jennimaria Palomaki, Séb Arnold, Madhavi Sewak, Federico Piccinini, Mohit
  Sharma, Ben Albrecht, Sean Purser-haskell, Ashwin Vaswani, Chongyan Chen,
  Matheus Wisniewski, Qin Cao, John Aslanides, Nguyet~Minh Phu, Maximilian
  Sieb, Lauren Agubuzu, Anne Zheng, Daniel Sohn, Marco Selvi, Anders
  Andreassen, Krishan Subudhi, Prem Eruvbetine, Oliver Woodman, Tomas Mery,
  Sebastian Krause, Xiaoqi Ren, Xiao Ma, Jincheng Luo, Dawn Chen, Wei Fan,
  Henry Griffiths, Christian Schuler, Alice Li, Shujian Zhang, Jean-Michel
  Sarr, Shixin Luo, Riccardo Patana, Matthew Watson, Dani Naboulsi, Michael
  Collins, Sailesh Sidhwani, Emiel Hoogeboom, Sharon Silver, Emily Caveness,
  Xiaokai Zhao, Mikel Rodriguez, Maxine Deines, Libin Bai, Patrick Griffin,
  Marco Tagliasacchi, Emily Xue, Spandana~Raj Babbula, Bo Pang, Nan Ding,
  Gloria Shen, Elijah Peake, Remi Crocker, Shubha~Srinivas Raghvendra, Danny
  Swisher, Woohyun Han, Richa Singh, Ling Wu, Vladimir Pchelin, Tsendsuren
  Munkhdalai, Dana Alon, Geoff Bacon, Efren Robles, Jannis Bulian, Melvin
  Johnson, George Powell, Felipe~Tiengo Ferreira, Yaoyiran Li, Frederik
  Benzing, Mihajlo Velimirović, Hubert Soyer, William Kong, Tony, Nguyên,
  Zhen Yang, Jeremiah Liu, Joost van Amersfoort, Daniel Gillick, Baochen Sun,
  Nathalie Rauschmayr, Katie Zhang, Serena Zhan, Tao Zhou, Alexey Frolov,
  Chengrun Yang, Denis Vnukov, Louis Rouillard, Hongji Li, Amol Mandhane, Nova
  Fallen, Rajesh Venkataraman, Clara~Huiyi Hu, Jennifer Brennan, Jenny Lee,
  Jerry Chang, Martin Sundermeyer, Zhufeng Pan, Rosemary Ke, Simon Tong, Alex
  Fabrikant, William Bono, Jindong Gu, Ryan Foley, Yiran Mao, Manolis Delakis,
  Dhruva Bhaswar, Roy Frostig, Nick Li, Avital Zipori, Cath Hope, Olga Kozlova,
  Swaroop Mishra, Josip Djolonga, Craig Schiff, Majd~Al Merey, Eleftheria
  Briakou, Peter Morgan, Andy Wan, Avinatan Hassidim, RJ Skerry-Ryan, Kuntal
  Sengupta, Mary Jasarevic, Praveen Kallakuri, Paige Kunkle, Hannah Brennan,
  Tom Lieber, Hassan Mansoor, Julian Walker, Bing Zhang, Annie Xie, Goran
  Žužić, Adaeze Chukwuka, Alex Druinsky, Donghyun Cho, Rui Yao, Ferjad
  Naeem, Shiraz Butt, Eunyoung Kim, Zhipeng Jia, Mandy Jordan, Adam Lelkes,
  Mark Kurzeja, Sophie Wang, James Zhao, Andrew Over, Abhishek Chakladar,
  Marcel Prasetya, Neha Jha, Sriram Ganapathy, Yale Cong, Prakash Shroff, Carl
  Saroufim, Sobhan Miryoosefi, Mohamed Hammad, Tajwar Nasir, Weijuan Xi, Yang
  Gao, Young Maeng, Ben Hora, Chin-Yi Cheng, Parisa Haghani, Yoad Lewenberg,
  Caden Lu, Martin Matysiak, Naina Raisinghani, Huiyu Wang, Lexi Baugher, Rahul
  Sukthankar, Minh Giang, John Schultz, Noah Fiedel, Minmin Chen, Cheng-Chun
  Lee, Tapomay Dey, Hao Zheng, Shachi Paul, Celine Smith, Andy Ly, Yicheng
  Wang, Rishabh Bansal, Bartek Perz, Susanna Ricco, Stasha Blank, Vaishakh
  Keshava, Deepak Sharma, Marvin Chow, Kunal Lad, Komal Jalan, Simon Osindero,
  Craig Swanson, Jacob Scott, Anastasija Ilić, Xiaowei Li, Siddhartha~Reddy
  Jonnalagadda, Afzal~Shama Soudagar, Yan Xiong, Bat-Orgil Batsaikhan, Daniel
  Jarrett, Naveen Kumar, Maulik Shah, Matt Lawlor, Austin Waters, Mark Graham,
  Rhys May, Sabela Ramos, Sandra Lefdal, Zeynep Cankara, Nacho Cano, Brendan
  O'Donoghue, Jed Borovik, Frederick Liu, Jordan Grimstad, Mahmoud Alnahlawi,
  Katerina Tsihlas, Tom Hudson, Nikolai Grigorev, Yiling Jia, Terry Huang,
  Tobenna~Peter Igwe, Sergei Lebedev, Xiaodan Tang, Igor Krivokon, Frankie
  Garcia, Melissa Tan, Eric Jia, Peter Stys, Shikhar Vashishth, Yu Liang,
  Balaji Venkatraman, Chenjie Gu, Anastasios Kementsietsidis, Chen Zhu,
  Junehyuk Jung, Yunfei Bai, Mohammad~Javad Hosseini, Faruk Ahmed, Aditya
  Gupta, Xin Yuan, Shereen Ashraf, Shitij Nigam, Gautam Vasudevan, Pranjal
  Awasthi, Adi~Mayrav Gilady, Zelda Mariet, Ramy Eskander, Haiguang Li, Hexiang
  Hu, Guillermo Garrido, Philippe Schlattner, George Zhang, Rohun Saxena, Petar
  Dević, Kritika Muralidharan, Ashwin Murthy, Yiqian Zhou, Min Choi, Arissa
  Wongpanich, Zhengdong Wang, Premal Shah, Yuntao Xu, Yiling Huang, Stephen
  Spencer, Alice Chen, James Cohan, Junjie Wang, Jonathan Tompson, Junru Wu,
  Ruba Haroun, Haiqiong Li, Blanca Huergo, Fan Yang, Tongxin Yin, James Wendt,
  Michael Bendersky, Rahma Chaabouni, Javier Snaider, Johan Ferret, Abhishek
  Jindal, Tara Thompson, Andrew Xue, Will Bishop, Shubham~Milind Phal, Archit
  Sharma, Yunhsuan Sung, Prabakar Radhakrishnan, Mo Shomrat, Reeve Ingle,
  Roopali Vij, Justin Gilmer, Mihai~Dorin Istin, Sam Sobell, Yang Lu, Emily
  Nottage, Dorsa Sadigh, Jeremiah Willcock, Tingnan Zhang, Steve Xu, Sasha
  Brown, Katherine Lee, Gary Wang, Yun Zhu, Yi Tay, Cheolmin Kim, Audrey
  Gutierrez, Abhanshu Sharma, Yongqin Xian, Sungyong Seo, Claire Cui, Elena
  Pochernina, Cip Baetu, Krzysztof Jastrzębski, Mimi Ly, Mohamed Elhawaty, Dan
  Suh, Eren Sezener, Pidong Wang, Nancy Yuen, George Tucker, Jiahao Cai,
  Zuguang Yang, Cindy Wang, Alex Muzio, Hai Qian, Jae Yoo, Derek Lockhart,
  Kevin~R. McKee, Mandy Guo, Malika Mehrotra, Artur Mendonça, Sanket~Vaibhav
  Mehta, Sherry Ben, Chetan Tekur, Jiaqi Mu, Muye Zhu, Victoria Krakovna,
  Hongrae Lee, AJ Maschinot, Sébastien Cevey, HyunJeong Choe, Aijun Bai, Hansa
  Srinivasan, Derek Gasaway, Nick Young, Patrick Siegler, Dan Holtmann-Rice,
  Vihari Piratla, Kate Baumli, Roey Yogev, Alex Hofer, Hado van Hasselt,
  Svetlana Grant, Yuri Chervonyi, David Silver, Andrew Hogue, Ayushi Agarwal,
  Kathie Wang, Preeti Singh, Four Flynn, Josh Lipschultz, Robert David,
  Lizzetth Bellot, Yao-Yuan Yang, Long Le, Filippo Graziano, Kate Olszewska,
  Kevin Hui, Akanksha Maurya, Nikos Parotsidis, Weijie Chen, Tayo Oguntebi, Joe
  Kelley, Anirudh Baddepudi, Johannes Mauerer, Gregory Shaw, Alex Siegman, Lin
  Yang, Shravya Shetty, Subhrajit Roy, Yunting Song, Wojciech Stokowiec, Ryan
  Burnell, Omkar Savant, Robert Busa-Fekete, Jin Miao, Samrat Ghosh, Liam
  MacDermed, Phillip Lippe, Mikhail Dektiarev, Zach Behrman, Fabian Mentzer,
  Kelvin Nguyen, Meng Wei, Siddharth Verma, Chris Knutsen, Sudeep Dasari,
  Zhipeng Yan, Petr Mitrichev, Xingyu Wang, Virat Shejwalkar, Jacob Austin,
  Srinivas Sunkara, Navneet Potti, Yan Virin, Christian Wright, Gaël Liu,
  Oriana Riva, Etienne Pot, Greg Kochanski, Quoc Le, Gargi Balasubramaniam,
  Arka Dhar, Yuguo Liao, Adam Bloniarz, Divyansh Shukla, Elizabeth Cole, Jong
  Lee, Sheng Zhang, Sushant Kafle, Siddharth Vashishtha, Parsa Mahmoudieh,
  Grace Chen, Raphael Hoffmann, Pranesh Srinivasan, Agustin~Dal Lago, Yoav~Ben
  Shalom, Zi Wang, Michael Elabd, Anuj Sharma, Junhyuk Oh, Suraj Kothawade,
  Maigo Le, Marianne Monteiro, Shentao Yang, Kaiz Alarakyia, Robert Geirhos,
  Diana Mincu, Håvard Garnes, Hayato Kobayashi, Soroosh Mariooryad, Kacper
  Krasowiak, Zhixin, Lai, Shibl Mourad, Mingqiu Wang, Fan Bu, Ophir Aharoni,
  Guanjie Chen, Abhimanyu Goyal, Vadim Zubov, Ankur Bapna, Elahe Dabir, Nisarg
  Kothari, Kay Lamerigts, Nicola~De Cao, Jeremy Shar, Christopher Yew, Nitish
  Kulkarni, Dre Mahaarachchi, Mandar Joshi, Zhenhai Zhu, Jared Lichtarge,
  Yichao Zhou, Hannah Muckenhirn, Vittorio Selo, Oriol Vinyals, Peter Chen,
  Anthony Brohan, Vaibhav Mehta, Sarah Cogan, Ruth Wang, Ty Geri, Wei-Jen Ko,
  Wei Chen, Fabio Viola, Keshav Shivam, Lisa Wang, Madeleine~Clare Elish,
  Raluca~Ada Popa, Sébastien Pereira, Jianqiao Liu, Raphael Koster, Donnie
  Kim, Gufeng Zhang, Sayna Ebrahimi, Partha Talukdar, Yanyan Zheng, Petra
  Poklukar, Ales Mikhalap, Dale Johnson, Anitha Vijayakumar, Mark Omernick,
  Matt Dibb, Ayush Dubey, Qiong Hu, Apurv Suman, Vaibhav Aggarwal, Ilya
  Kornakov, Fei Xia, Wing Lowe, Alexey Kolganov, Ted Xiao, Vitaly Nikolaev,
  Steven Hemingray, Bonnie Li, Joana Iljazi, Mikołaj Rybiński, Ballie Sandhu,
  Peggy Lu, Thang Luong, Rodolphe Jenatton, Vineetha Govindaraj, Hui, Li,
  Gabriel Dulac-Arnold, Wonpyo Park, Henry Wang, Abhinit Modi, Jean
  Pouget-Abadie, Kristina Greller, Rahul Gupta, Robert Berry, Prajit
  Ramachandran, Jinyu Xie, Liam McCafferty, Jianling Wang, Kilol Gupta,
  Hyeontaek Lim, Blaž Bratanič, Andy Brock, Ilia Akolzin, Jim Sproch, Dan
  Karliner, Duhyeon Kim, Adrian Goedeckemeyer, Noam Shazeer, Cordelia Schmid,
  Daniele Calandriello, Parul Bhatia, Krzysztof Choromanski, Ceslee Montgomery,
  Dheeru Dua, Ana Ramalho, Helen King, Yue Gao, Lynn Nguyen, David Lindner,
  Divya Pitta, Oleaser Johnson, Khalid Salama, Diego Ardila, Michael Han, Erin
  Farnese, Seth Odoom, Ziyue Wang, Xiangzhuo Ding, Norman Rink, Ray Smith,
  Harshal~Tushar Lehri, Eden Cohen, Neera Vats, Tong He, Parthasarathy
  Gopavarapu, Adam Paszke, Miteyan Patel, Wouter~Van Gansbeke, Lucia Loher,
  Luis Castro, Maria Voitovich, Tamara von Glehn, Nelson George, Simon Niklaus,
  Zach Eaton-Rosen, Nemanja Rakićević, Erik Jue, Sagi Perel, Carrie Zhang,
  Yuval Bahat, Angéline Pouget, Zhi Xing, Fantine Huot, Ashish Shenoy, Taylor
  Bos, Vincent Coriou, Bryan Richter, Natasha Noy, Yaqing Wang, Santiago
  Ontanon, Siyang Qin, Gleb Makarchuk, Demis Hassabis, Zhuowan Li, Mandar
  Sharma, Kumaran Venkatesan, Iurii Kemaev, Roxanne Daniel, Shiyu Huang, Saloni
  Shah, Octavio Ponce, Warren, Chen, Manaal Faruqui, Jialin Wu, Slavica
  Andačić, Szabolcs Payrits, Daniel McDuff, Tom Hume, Yuan Cao, MH Tessler,
  Qingze Wang, Yinan Wang, Ivor Rendulic, Eirikur Agustsson, Matthew Johnson,
  Tanya Lando, Andrew Howard, Sri Gayatri~Sundara Padmanabhan, Mayank Daswani,
  Andrea Banino, Michael Kilgore, Jonathan Heek, Ziwei Ji, Alvaro Caceres,
  Conglong Li, Nora Kassner, Alexey Vlaskin, Zeyu Liu, Alex Grills, Yanhan Hou,
  Roykrong Sukkerd, Gowoon Cheon, Nishita Shetty, Larisa Markeeva, Piotr
  Stanczyk, Tejas Iyer, Yuan Gong, Shawn Gao, Keerthana Gopalakrishnan, Tim
  Blyth, Malcolm Reynolds, Avishkar Bhoopchand, Misha Bilenko, Dero Gharibian,
  Vicky Zayats, Aleksandra Faust, Abhinav Singh, Min Ma, Hongyang Jiao,
  Sudheendra Vijayanarasimhan, Lora Aroyo, Vikas Yadav, Sarah Chakera, Ashwin
  Kakarla, Vilobh Meshram, Karol Gregor, Gabriela Botea, Evan Senter, Dawei
  Jia, Geza Kovacs, Neha Sharma, Sebastien Baur, Kai Kang, Yifan He, Lin Zhuo,
  Marija Kostelac, Itay Laish, Songyou Peng, Louis O'Bryan, Daniel Kasenberg,
  Girish~Ramchandra Rao, Edouard Leurent, Biao Zhang, Sage Stevens, Ana
  Salazar, Ye Zhang, Ivan Lobov, Jake Walker, Allen Porter, Morgan Redshaw, Han
  Ke, Abhishek Rao, Alex Lee, Hoi Lam, Michael Moffitt, Jaeyoun Kim, Siyuan
  Qiao, Terry Koo, Robert Dadashi, Xinying Song, Mukund Sundararajan, Peng Xu,
  Chizu Kawamoto, Yan Zhong, Clara Barbu, Apoorv Reddy, Mauro Verzetti, Leon
  Li, George Papamakarios, Hanna Klimczak-Plucińska, Mary Cassin, Koray
  Kavukcuoglu, Rigel Swavely, Alain Vaucher, Jeffrey Zhao, Ross Hemsley,
  Michael Tschannen, Heming Ge, Gaurav Menghani, Yang Yu, Natalie Ha, Wei He,
  Xiao Wu, Maggie Song, Rachel Sterneck, Stefan Zinke, Dan~A. Calian, Annie
  Marsden, Alejandro~Cruzado Ruiz, Matteo Hessel, Almog Gueta, Benjamin Lee,
  Brian Farris, Manish Gupta, Yunjie Li, Mohammad Saleh, Vedant Misra, Kefan
  Xiao, Piermaria Mendolicchio, Gavin Buttimore, Varvara Krayvanova, Nigamaa
  Nayakanti, Matthew Wiethoff, Yash Pande, Azalia Mirhoseini, Ni Lao, Jasmine
  Liu, Yiqing Hua, Angie Chen, Yury Malkov, Dmitry Kalashnikov, Shubham Gupta,
  Kartik Audhkhasi, Yuexiang Zhai, Sudhindra Kopalle, Prateek Jain, Eran Ofek,
  Clemens Meyer, Khuslen Baatarsukh, Hana Strejček, Jun Qian, James Freedman,
  Ricardo Figueira, Michal Sokolik, Olivier Bachem, Raymond Lin, Dia Kharrat,
  Chris Hidey, Pingmei Xu, Dennis Duan, Yin Li, Muge Ersoy, Richard Everett,
  Kevin Cen, Rebeca Santamaria-Fernandez, Amir Taubenfeld, Ian Mackinnon, Linda
  Deng, Polina Zablotskaia, Shashank Viswanadha, Shivanker Goel, Damion Yates,
  Yunxiao Deng, Peter Choy, Mingqing Chen, Abhishek Sinha, Alex Mossin, Yiming
  Wang, Arthur Szlam, Susan Hao, Paul~Kishan Rubenstein, Metin Toksoz-Exley,
  Miranda Aperghis, Yin Zhong, Junwhan Ahn, Michael Isard, Olivier Lacombe,
  Florian Luisier, Chrysovalantis Anastasiou, Yogesh Kalley, Utsav Prabhu, Emma
  Dunleavy, Shaan Bijwadia, Justin Mao-Jones, Kelly Chen, Rama Pasumarthi,
  Emily Wood, Adil Dostmohamed, Nate Hurley, Jiri Simsa, Alicia Parrish, Mantas
  Pajarskas, Matt Harvey, Ondrej Skopek, Yony Kochinski, Javier Rey, Verena
  Rieser, Denny Zhou, Sun~Jae Lee, Trilok Acharya, Guowang Li, Joe Jiang,
  Xiaofan Zhang, Bryant Gipson, Ethan Mahintorabi, Marco Gelmi, Nima
  Khajehnouri, Angel Yeh, Kayi Lee, Loic Matthey, Leslie Baker, Trang Pham, Han
  Fu, Alex Pak, Prakhar Gupta, Cristina Vasconcelos, Adam Sadovsky, Brian
  Walker, Sissie Hsiao, Patrik Zochbauer, Andreea Marzoca, Noam Velan, Junhao
  Zeng, Gilles Baechler, Danny Driess, Divya Jain, Yanping Huang, Lizzie Tao,
  John Maggs, Nir Levine, Jon Schneider, Erika Gemzer, Samuel Petit, Shan Han,
  Zach Fisher, Dustin Zelle, Courtney Biles, Eugene Ie, Asya Fadeeva, Casper
  Liu, Juliana~Vicente Franco, Adrian Collister, Hao Zhang, Renshen Wang,
  Ruizhe Zhao, Leandro Kieliger, Kurt Shuster, Rui Zhu, Boqing Gong, Lawrence
  Chan, Ruoxi Sun, Sujoy Basu, Roland Zimmermann, Jamie Hayes, Abhishek Bapna,
  Jasper Snoek, Weel Yang, Puranjay Datta, Jad~Al Abdallah, Kevin Kilgour, Lu
  Li, SQ Mah, Yennie Jun, Morgane Rivière, Abhijit Karmarkar, Tammo Spalink,
  Tao Huang, Lucas Gonzalez, Duc-Hieu Tran, Averi Nowak, John Palowitch, Martin
  Chadwick, Ellie Talius, Harsh Mehta, Thibault Sellam, Philipp Fränken,
  Massimo Nicosia, Kyle He, Aditya Kini, David Amos, Sugato Basu, Harrison
  Jobe, Eleni Shaw, Qiantong Xu, Colin Evans, Daisuke Ikeda, Chaochao Yan,
  Larry Jin, Lun Wang, Sachin Yadav, Ilia Labzovsky, Ramesh Sampath, Ada Ma,
  Candice Schumann, Aditya Siddhant, Rohin Shah, John Youssef, Rishabh Agarwal,
  Natalie Dabney, Alessio Tonioni, Moran Ambar, Jing Li, Isabelle Guyon, Benny
  Li, David Soergel, Boya Fang, Georgi Karadzhov, Cristian Udrescu, Trieu
  Trinh, Vikas Raunak, Seb Noury, Dee Guo, Sonal Gupta, Mara Finkelstein, Denis
  Petek, Lihao Liang, Greg Billock, Pei Sun, David Wood, Yiwen Song, Xiaobin
  Yu, Tatiana Matejovicova, Regev Cohen, Kalyan Andra, David D'Ambrosio, Zhiwei
  Deng, Vincent Nallatamby, Ebrahim Songhori, Rumen Dangovski, Andrew Lampinen,
  Pankil Botadra, Adam Hillier, Jiawei Cao, Nagabhushan Baddi, Adhi Kuncoro,
  Toshihiro Yoshino, Ankit Bhagatwala, Marcáurelio Ranzato, Rylan Schaeffer,
  Tianlin Liu, Shuai Ye, Obaid Sarvana, John Nham, Chenkai Kuang, Isabel Gao,
  Jinoo Baek, Shubham Mittal, Ayzaan Wahid, Anita Gergely, Bin Ni, Josh
  Feldman, Carrie Muir, Pascal Lamblin, Wolfgang Macherey, Ethan Dyer, Logan
  Kilpatrick, Víctor Campos, Mukul Bhutani, Stanislav Fort, Yanif Ahmad,
  Aliaksei Severyn, Kleopatra Chatziprimou, Oleksandr Ferludin, Mason Dimarco,
  Aditya Kusupati, Joe Heyward, Dan Bahir, Kevin Villela, Katie Millican, Dror
  Marcus, Sanaz Bahargam, Caglar Unlu, Nicholas Roth, Zichuan Wei, Siddharth
  Gopal, Deepanway Ghoshal, Edward Lee, Sharon Lin, Jennie Lees, Dayeong Lee,
  Anahita Hosseini, Connie Fan, Seth Neel, Marcus Wu, Yasemin Altun, Honglong
  Cai, Enrique Piqueras, Josh Woodward, Alessandro Bissacco, Salem Haykal,
  Mahyar Bordbar, Prasha Sundaram, Sarah Hodkinson, Daniel Toyama, George
  Polovets, Austin Myers, Anu Sinha, Tomer Levinboim, Kashyap Krishnakumar,
  Rachita Chhaparia, Tatiana Sholokhova, Nitesh~Bharadwaj Gundavarapu, Ganesh
  Jawahar, Haroon Qureshi, Jieru Hu, Nikola Momchev, Matthew Rahtz, Renjie Wu,
  Aishwarya~P S, Kedar Dhamdhere, Meiqi Guo, Umang Gupta, Ali Eslami, Mariano
  Schain, Michiel Blokzijl, David Welling, Dave Orr, Levent Bolelli, Nicolas
  Perez-Nieves, Mikhail Sirotenko, Aman Prasad, Arjun Kar, Borja De~Balle
  Pigem, Tayfun Terzi, Gellért Weisz, Dipankar Ghosh, Aditi Mavalankar, Dhruv
  Madeka, Kaspar Daugaard, Hartwig Adam, Viraj Shah, Dana Berman, Maggie Tran,
  Steven Baker, Ewa Andrejczuk, Grishma Chole, Ganna Raboshchuk, Mahdi
  Mirzazadeh, Thais Kagohara, Shimu Wu, Christian Schallhart, Bernett Orlando,
  Chen Wang, Alban Rrustemi, Hao Xiong, Hao Liu, Arpi Vezer, Nolan Ramsden,
  Shuo yiin Chang, Sidharth Mudgal, Yan Li, Nino Vieillard, Yedid Hoshen,
  Farooq Ahmad, Ambrose Slone, Amy Hua, Natan Potikha, Mirko Rossini, Jon
  Stritar, Sushant Prakash, Zifeng Wang, Xuanyi Dong, Alireza Nazari, Efrat
  Nehoran, Kaan Tekelioglu, Yinxiao Li, Kartikeya Badola, Tom Funkhouser,
  Yuanzhen Li, Varun Yerram, Ramya Ganeshan, Daniel Formoso, Karol Langner,
  Tian Shi, Huijian Li, Yumeya Yamamori, Amayika Panda, Alaa Saade,
  Angelo~Scorza Scarpati, Chris Breaux, CJ Carey, Zongwei Zhou, Cho-Jui Hsieh,
  Sophie Bridgers, Alena Butryna, Nishesh Gupta, Vaibhav Tulsyan, Sanghyun Woo,
  Evgenii Eltyshev, Will Grathwohl, Chanel Parks, Seth Benjamin, Rina
  Panigrahy, Shenil Dodhia, Daniel~De Freitas, Chris Sauer, Will Song, Ferran
  Alet, Jackson Tolins, Cosmin Paduraru, Xingyi Zhou, Brian Albert, Zizhao
  Zhang, Lei Shu, Mudit Bansal, Sarah Nguyen, Amir Globerson, Owen Xiao, James
  Manyika, Tom Hennigan, Rong Rong, Josip Matak, Anton Bakalov, Ankur Sharma,
  Danila Sinopalnikov, Andrew Pierson, Stephen Roller, Geoff Brown, Mingcen
  Gao, Toshiyuki Fukuzawa, Amin Ghafouri, Kenny Vassigh, Iain Barr, Zhicheng
  Wang, Anna Korsun, Rajesh Jayaram, Lijie Ren, Tim Zaman, Samira Khan, Yana
  Lunts, Dan Deutsch, Dave Uthus, Nitzan Katz, Masha Samsikova, Amr Khalifa,
  Nikhil Sethi, Jiao Sun, Luming Tang, Uri Alon, Xianghong Luo, Dian Yu,
  Abhishek Nayyar, Bryce Petrini, Will Truong, Vincent Hellendoorn, Nikolai
  Chinaev, Chris Alberti, Wei Wang, Jingcao Hu, Vahab Mirrokni, Ananth
  Balashankar, Avia Aharon, Aahil Mehta, Ahmet Iscen, Joseph Kready, Lucas
  Manning, Anhad Mohananey, Yuankai Chen, Anshuman Tripathi, Allen Wu, Igor
  Petrovski, Dawsen Hwang, Martin Baeuml, Shreyas Chandrakaladharan, Yuan Liu,
  Rey Coaguila, Maxwell Chen, Sally Ma, Pouya Tafti, Susheel Tatineni, Terry
  Spitz, Jiayu Ye, Paul Vicol, Mihaela Rosca, Adrià Puigdomènech, Zohar
  Yahav, Sanjay Ghemawat, Hanzhao Lin, Phoebe Kirk, Zaid Nabulsi, Sergey Brin,
  Bernd Bohnet, Ken Caluwaerts, Aditya~Srikanth Veerubhotla, Dan Zheng, Zihang
  Dai, Petre Petrov, Yichong Xu, Ramin Mehran, Zhuo Xu, Luisa Zintgraf, Jiho
  Choi, Spurthi~Amba Hombaiah, Romal Thoppilan, Sashank Reddi, Lukasz Lew, Li
  Li, Kellie Webster, KP Sawhney, Lampros Lamprou, Siamak Shakeri, Mayank
  Lunayach, Jianmin Chen, Sumit Bagri, Alex Salcianu, Ying Chen, Yani Donchev,
  Charlotte Magister, Signe Nørly, Vitor Rodrigues, Tomas Izo, Hila Noga, Joe
  Zou, Thomas Köppe, Wenxuan Zhou, Kenton Lee, Xiangzhu Long, Danielle
  Eisenbud, Anthony Chen, Connor Schenck, Chi~Ming To, Peilin Zhong, Emanuel
  Taropa, Minh Truong, Omer Levy, Danilo Martins, Zhiyuan Zhang, Christopher
  Semturs, Kelvin Zhang, Alex Yakubovich, Pol Moreno, Lara McConnaughey, Di Lu,
  Sam Redmond, Lotte Weerts, Yonatan Bitton, Tiziana Refice, Nicolas Lacasse,
  Arthur Conmy, Corentin Tallec, Julian Odell, Hannah Forbes-Pollard, Arkadiusz
  Socala, Jonathan Hoech, Pushmeet Kohli, Alanna Walton, Rui Wang, Mikita
  Sazanovich, Kexin Zhu, Andrei Kapishnikov, Rich Galt, Matthew Denton, Ben
  Murdoch, Caitlin Sikora, Kareem Mohamed, Wei Wei, Uri First, Tim McConnell,
  Luis~C. Cobo, James Qin, Thi Avrahami, Daniel Balle, Yu Watanabe, Annie
  Louis, Adam Kraft, Setareh Ariafar, Yiming Gu, Eugénie Rives, Charles Yoon,
  Andrei Rusu, James Cobon-Kerr, Chris Hahn, Jiaming Luo, Yuvein, Zhu, Niharika
  Ahuja, Rodrigo Benenson, Raphaël~Lopez Kaufman, Honglin Yu, Lloyd Hightower,
  Junlin Zhang, Darren Ni, Lisa~Anne Hendricks, Gabby Wang, Gal Yona, Lalit
  Jain, Pablo Barrio, Surya Bhupatiraju, Siva Velusamy, Allan Dafoe, Sebastian
  Riedel, Tara Thomas, Zhe Yuan, Mathias Bellaiche, Sheena Panthaplackel,
  Klemen Kloboves, Sarthak Jauhari, Canfer Akbulut, Todor Davchev, Evgeny
  Gladchenko, David Madras, Aleksandr Chuklin, Tyrone Hill, Quan Yuan, Mukundan
  Madhavan, Luke Leonhard, Dylan Scandinaro, Qihang Chen, Ning Niu, Arthur
  Douillard, Bogdan Damoc, Yasumasa Onoe, Fabian Pedregosa, Fred Bertsch, Chas
  Leichner, Joseph Pagadora, Jonathan Malmaud, Sameera Ponda, Andy Twigg,
  Oleksii Duzhyi, Jingwei Shen, Miaosen Wang, Roopal Garg, Jing Chen, Utku
  Evci, Jonathan Lee, Leon Liu, Koji Kojima, Masa Yamaguchi, Arunkumar
  Rajendran, AJ Piergiovanni, Vinodh~Kumar Rajendran, Marco Fornoni, Gabriel
  Ibagon, Harry Ragan, Sadh~MNM Khan, John Blitzer, Andrew Bunner, Guan Sun,
  Takahiro Kosakai, Scott Lundberg, Ndidi Elue, Kelvin Guu, SK Park, Jane Park,
  Arunachalam Narayanaswamy, Chengda Wu, Jayaram Mudigonda, Trevor Cohn,
  Hairong Mu, Ravi Kumar, Laura Graesser, Yichi Zhang, Richard Killam, Vincent
  Zhuang, Mai Giménez, Wael~Al Jishi, Ruy Ley-Wild, Alex Zhai, Kazuki Osawa,
  Diego Cedillo, Jialu Liu, Mayank Upadhyay, Marcin Sieniek, Roshan Sharma, Tom
  Paine, Anelia Angelova, Sravanti Addepalli, Carolina Parada, Kingshuk
  Majumder, Avery Lamp, Sanjiv Kumar, Xiang Deng, Artiom Myaskovsky, Tea
  Sabolić, Jeffrey Dudek, Sarah York, Félix de Chaumont~Quitry, Jiazhong Nie,
  Dee Cattle, Alok Gunjan, Bilal Piot, Waleed Khawaja, Seojin Bang, Simon Wang,
  Siavash Khodadadeh, Raghavender R, Praynaa Rawlani, Richard Powell, Kevin
  Lee, Johannes Griesser, GS Oh, Cesar Magalhaes, Yujia Li, Simon Tokumine,
  Hadas~Natalie Vogel, Dennis Hsu, Arturo BC, Disha Jindal, Matan Cohen, Zi
  Yang, Junwei Yuan, Dario de Cesare, Tony Bruguier, Jun Xu, Monica Roy, Alon
  Jacovi, Dan Belov, Rahul Arya, Phoenix Meadowlark, Shlomi Cohen-Ganor,
  Wenting Ye, Patrick Morris-Suzuki, Praseem Banzal, Gan Song, Pranavaraj
  Ponnuramu, Fred Zhang, George Scrivener, Salah Zaiem, Alif~Raditya Rochman,
  Kehang Han, Badih Ghazi, Kate Lee, Shahar Drath, Daniel Suo, Antonious
  Girgis, Pradeep Shenoy, Duy Nguyen, Douglas Eck, Somit Gupta, Le Yan, Joao
  Carreira, Anmol Gulati, Ruoxin Sang, Daniil Mirylenka, Emma Cooney, Edward
  Chou, Mingyang Ling, Cindy Fan, Ben Coleman, Guilherme Tubone, Ravin Kumar,
  Jason Baldridge, Felix Hernandez-Campos, Angeliki Lazaridou, James Besley,
  Itay Yona, Neslihan Bulut, Quentin Wellens, AJ Pierigiovanni, Jasmine George,
  Richard Green, Pu Han, Connie Tao, Geoff Clark, Chong You, Abbas Abdolmaleki,
  Justin Fu, Tongzhou Chen, Ashwin Chaugule, Angad Chandorkar, Altaf Rahman,
  Will Thompson, Penporn Koanantakool, Mike Bernico, Jie Ren, Andrey Vlasov,
  Sergei Vassilvitskii, Maciej Kula, Yizhong Liang, Dahun Kim, Yangsibo Huang,
  Chengxi Ye, Dmitry Lepikhin, and Wesley Helmholz.
\newblock Gemini 2.5: Pushing the frontier with advanced reasoning,
  multimodality, long context, and next generation agentic capabilities, 2025.

\bibitem[Dai et~al.(2026)Dai, Jiao, Lin, and
  Zhang]{dai2026egocotbenchbenchmarkinggroundedverifiable}
Yang Dai, Dian Jiao, Tianwei Lin, and Wenqiao Zhang.
\newblock Egocot-bench: Benchmarking grounded and verifiable operation-centric
  chain of thought reasoning for mllms, 2026.

\bibitem[Endsley(1995)]{Endsley_1995}
Mica~R. Endsley.
\newblock Toward a theory of situation awareness in dynamic systems.
\newblock \emph{Human Factors: The Journal of the Human Factors and Ergonomics
  Society}, 37\penalty0 (1):\penalty0 32–64, 1995.

\bibitem[Fiorini and Shiller(1998)]{Fiorini_1998}
Paolo Fiorini and Zvi Shiller.
\newblock Motion planning in dynamic environments using velocity obstacles.
\newblock \emph{The International Journal of Robotics Research}, 17\penalty0
  (7):\penalty0 760–772, 1998.

\bibitem[Forte et~al.(2026)Forte, Lando, and
  Furnari]{forte2026egostreamdiagnosticbenchmarkstreaming}
Rosario Forte, Giuseppe Lando, and Antonino Furnari.
\newblock Egostream: A diagnostic benchmark for streaming episodic memory in
  egocentric vision, 2026.

\bibitem[Fu et~al.(2025)Fu, Dai, Luo, Li, Ren, Zhang, Wang, Zhou, Shen, Zhang,
  Chen, Li, Lin, Zhao, Li, Xu, Zheng, Chen, Shan, He, and Sun]{Fu_2025_CVPR}
Chaoyou Fu, Yuhan Dai, Yongdong Luo, Lei Li, Shuhuai Ren, Renrui Zhang, Zihan
  Wang, Chenyu Zhou, Yunhang Shen, Mengdan Zhang, Peixian Chen, Yanwei Li,
  Shaohui Lin, Sirui Zhao, Ke Li, Tong Xu, Xiawu Zheng, Enhong Chen, Caifeng
  Shan, Ran He, and Xing Sun.
\newblock Video-mme: The first-ever comprehensive evaluation benchmark of
  multi-modal llms in video analysis.
\newblock In \emph{Proceedings of the IEEE/CVF Conference on Computer Vision
  and Pattern Recognition (CVPR)}, pages 24108--24118, 2025.

\bibitem[Fu et~al.(2026)Fu, Yuan, Dong, Zhang, Shen, Hu, Li, Su, Long, Xie,
  Xie, Zheng, Yang, Cao, Wu, Liu, Sun, Shan, and
  He]{fu2026videommev2stagebenchmarkscomprehensive}
Chaoyou Fu, Haozhi Yuan, Yuhao Dong, Yi-Fan Zhang, Yunhang Shen, Xiaoxing Hu,
  Xueying Li, Jinsen Su, Chengwu Long, Xiaoyao Xie, Yongkang Xie, Xiawu Zheng,
  Xue Yang, Haoyu Cao, Yunsheng Wu, Ziwei Liu, Xing Sun, Caifeng Shan, and Ran
  He.
\newblock Video-mme-v2: Towards the next stage in benchmarks for comprehensive
  video understanding, 2026.

\bibitem[Gao et~al.(2026)Gao, Li, Cheng, Xu, Haworth, Law, and
  Ye]{gao2026unidriveunifiedvisionlanguagegrounding}
Xiaowei Gao, Pengxiang Li, Yitai Cheng, Ruihan Xu, James Haworth, Stephen Law,
  and Yun Ye.
\newblock Unidrive: A unified vision-language and grounding framework for
  interpretable risk understanding in autonomous driving, 2026.

\bibitem[Gong et~al.(2026)Gong, Yan, Kang, Zheng, Ruan, Lu, Zhang, Sato, and
  Huang]{gong2026vinci2}
Sitong Gong, Tianyu Yan, Caixin Kang, Bo Zheng, Xiang Ruan, Huchuan Lu, Kaipeng
  Zhang, Yoichi Sato, and Yifei Huang.
\newblock Vinci2: Providing proactive assistance in continuous egocentric
  videos.
\newblock In \emph{European Conference on Computer Vision (ECCV)}, 2026.

\bibitem[Grattafiori et~al.(2024)Grattafiori, Dubey, Jauhri, Pandey, Kadian,
  Al-Dahle, Letman, Mathur, Schelten, Vaughan, Yang, Fan, Goyal, Hartshorn,
  Yang, Mitra, Sravankumar, Korenev, Hinsvark, Rao, Zhang, Rodriguez,
  Gregerson, Spataru, Roziere, Biron, Tang, Chern, Caucheteux, Nayak, Bi,
  Marra, McConnell, Keller, Touret, Wu, Wong, Ferrer, Nikolaidis, Allonsius,
  Song, Pintz, Livshits, Wyatt, Esiobu, Choudhary, Mahajan, Garcia-Olano,
  Perino, Hupkes, Lakomkin, AlBadawy, Lobanova, Dinan, Smith, Radenovic,
  Guzmán, Zhang, Synnaeve, Lee, Anderson, Thattai, Nail, Mialon, Pang,
  Cucurell, Nguyen, Korevaar, Xu, Touvron, Zarov, Ibarra, Kloumann, Misra,
  Evtimov, Zhang, Copet, Lee, Geffert, Vranes, Park, Mahadeokar, Shah, van~der
  Linde, Billock, Hong, Lee, Fu, Chi, Huang, Liu, Wang, Yu, Bitton, Spisak,
  Park, Rocca, Johnstun, Saxe, Jia, Alwala, Prasad, Upasani, Plawiak, Li,
  Heafield, Stone, El-Arini, Iyer, Malik, Chiu, Bhalla, Lakhotia,
  Rantala-Yeary, van~der Maaten, Chen, Tan, Jenkins, Martin, Madaan, Malo,
  Blecher, Landzaat, de~Oliveira, Muzzi, Pasupuleti, Singh, Paluri, Kardas,
  Tsimpoukelli, Oldham, Rita, Pavlova, Kambadur, Lewis, Si, Singh, Hassan,
  Goyal, Torabi, Bashlykov, Bogoychev, Chatterji, Zhang, Duchenne, Çelebi,
  Alrassy, Zhang, Li, Vasic, Weng, Bhargava, Dubal, Krishnan, Koura, Xu, He,
  Dong, Srinivasan, Ganapathy, Calderer, Cabral, Stojnic, Raileanu, Maheswari,
  Girdhar, Patel, Sauvestre, Polidoro, Sumbaly, Taylor, Silva, Hou, Wang,
  Hosseini, Chennabasappa, Singh, Bell, Kim, Edunov, Nie, Narang, Raparthy,
  Shen, Wan, Bhosale, Zhang, Vandenhende, Batra, Whitman, Sootla, Collot,
  Gururangan, Borodinsky, Herman, Fowler, Sheasha, Georgiou, Scialom,
  Speckbacher, Mihaylov, Xiao, Karn, Goswami, Gupta, Ramanathan, Kerkez,
  Gonguet, Do, Vogeti, Albiero, Petrovic, Chu, Xiong, Fu, Meers, Martinet,
  Wang, Wang, Tan, Xia, Xie, Jia, Wang, Goldschlag, Gaur, Babaei, Wen, Song,
  Zhang, Li, Mao, Coudert, Yan, Chen, Papakipos, Singh, Srivastava, Jain,
  Kelsey, Shajnfeld, Gangidi, Victoria, Goldstand, Menon, Sharma, Boesenberg,
  Baevski, Feinstein, Kallet, Sangani, Teo, Yunus, Lupu, Alvarado, Caples, Gu,
  Ho, Poulton, Ryan, Ramchandani, Dong, Franco, Goyal, Saraf, Chowdhury,
  Gabriel, Bharambe, Eisenman, Yazdan, James, Maurer, Leonhardi, Huang, Loyd,
  Paola, Paranjape, Liu, Wu, Ni, Hancock, Wasti, Spence, Stojkovic, Gamido,
  Montalvo, Parker, Burton, Mejia, Liu, Wang, Kim, Zhou, Hu, Chu, Cai, Tindal,
  Feichtenhofer, Gao, Civin, Beaty, Kreymer, Li, Adkins, Xu, Testuggine, David,
  Parikh, Liskovich, Foss, Wang, Le, Holland, Dowling, Jamil, Montgomery,
  Presani, Hahn, Wood, Le, Brinkman, Arcaute, Dunbar, Smothers, Sun, Kreuk,
  Tian, Kokkinos, Ozgenel, Caggioni, Kanayet, Seide, Florez, Schwarz, Badeer,
  Swee, Halpern, Herman, Sizov, Guangyi, Zhang, Lakshminarayanan, Inan,
  Shojanazeri, Zou, Wang, Zha, Habeeb, Rudolph, Suk, Aspegren, Goldman, Zhan,
  Damlaj, Molybog, Tufanov, Leontiadis, Veliche, Gat, Weissman, Geboski, Kohli,
  Lam, Asher, Gaya, Marcus, Tang, Chan, Zhen, Reizenstein, Teboul, Zhong, Jin,
  Yang, Cummings, Carvill, Shepard, McPhie, Torres, Ginsburg, Wang, Wu, U,
  Saxena, Khandelwal, Zand, Matosich, Veeraraghavan, Michelena, Li, Jagadeesh,
  Huang, Chawla, Huang, Chen, Garg, A, Silva, Bell, Zhang, Guo, Yu, Moshkovich,
  Wehrstedt, Khabsa, Avalani, Bhatt, Mankus, Hasson, Lennie, Reso, Groshev,
  Naumov, Lathi, Keneally, Liu, Seltzer, Valko, Restrepo, Patel, Vyatskov,
  Samvelyan, Clark, Macey, Wang, Hermoso, Metanat, Rastegari, Bansal,
  Santhanam, Parks, White, Bawa, Singhal, Egebo, Usunier, Mehta, Laptev, Dong,
  Cheng, Chernoguz, Hart, Salpekar, Kalinli, Kent, Parekh, Saab, Balaji,
  Rittner, Bontrager, Roux, Dollar, Zvyagina, Ratanchandani, Yuvraj, Liang,
  Alao, Rodriguez, Ayub, Murthy, Nayani, Mitra, Parthasarathy, Li, Hogan,
  Battey, Wang, Howes, Rinott, Mehta, Siby, Bondu, Datta, Chugh, Hunt, Dhillon,
  Sidorov, Pan, Mahajan, Verma, Yamamoto, Ramaswamy, Lindsay, Lindsay, Feng,
  Lin, Zha, Patil, Shankar, Zhang, Zhang, Wang, Agarwal, Sajuyigbe, Chintala,
  Max, Chen, Kehoe, Satterfield, Govindaprasad, Gupta, Deng, Cho, Virk,
  Subramanian, Choudhury, Goldman, Remez, Glaser, Best, Koehler, Robinson, Li,
  Zhang, Matthews, Chou, Shaked, Vontimitta, Ajayi, Montanez, Mohan, Kumar,
  Mangla, Ionescu, Poenaru, Mihailescu, Ivanov, Li, Wang, Jiang, Bouaziz,
  Constable, Tang, Wu, Wang, Wu, Gao, Kleinman, Chen, Hu, Jia, Qi, Li, Zhang,
  Zhang, Adi, Nam, Yu, Wang, Zhao, Hao, Qian, Li, He, Rait, DeVito, Rosnbrick,
  Wen, Yang, Zhao, and Ma]{grattafiori2024llama3herdmodels}
Aaron Grattafiori, Abhimanyu Dubey, Abhinav Jauhri, Abhinav Pandey, Abhishek
  Kadian, Ahmad Al-Dahle, Aiesha Letman, Akhil Mathur, Alan Schelten, Alex
  Vaughan, Amy Yang, Angela Fan, Anirudh Goyal, Anthony Hartshorn, Aobo Yang,
  Archi Mitra, Archie Sravankumar, Artem Korenev, Arthur Hinsvark, Arun Rao,
  Aston Zhang, Aurelien Rodriguez, Austen Gregerson, Ava Spataru, Baptiste
  Roziere, Bethany Biron, Binh Tang, Bobbie Chern, Charlotte Caucheteux, Chaya
  Nayak, Chloe Bi, Chris Marra, Chris McConnell, Christian Keller, Christophe
  Touret, Chunyang Wu, Corinne Wong, Cristian~Canton Ferrer, Cyrus Nikolaidis,
  Damien Allonsius, Daniel Song, Danielle Pintz, Danny Livshits, Danny Wyatt,
  David Esiobu, Dhruv Choudhary, Dhruv Mahajan, Diego Garcia-Olano, Diego
  Perino, Dieuwke Hupkes, Egor Lakomkin, Ehab AlBadawy, Elina Lobanova, Emily
  Dinan, Eric~Michael Smith, Filip Radenovic, Francisco Guzmán, Frank Zhang,
  Gabriel Synnaeve, Gabrielle Lee, Georgia~Lewis Anderson, Govind Thattai,
  Graeme Nail, Gregoire Mialon, Guan Pang, Guillem Cucurell, Hailey Nguyen,
  Hannah Korevaar, Hu Xu, Hugo Touvron, Iliyan Zarov, Imanol~Arrieta Ibarra,
  Isabel Kloumann, Ishan Misra, Ivan Evtimov, Jack Zhang, Jade Copet, Jaewon
  Lee, Jan Geffert, Jana Vranes, Jason Park, Jay Mahadeokar, Jeet Shah, Jelmer
  van~der Linde, Jennifer Billock, Jenny Hong, Jenya Lee, Jeremy Fu, Jianfeng
  Chi, Jianyu Huang, Jiawen Liu, Jie Wang, Jiecao Yu, Joanna Bitton, Joe
  Spisak, Jongsoo Park, Joseph Rocca, Joshua Johnstun, Joshua Saxe, Junteng
  Jia, Kalyan~Vasuden Alwala, Karthik Prasad, Kartikeya Upasani, Kate Plawiak,
  Ke Li, Kenneth Heafield, Kevin Stone, Khalid El-Arini, Krithika Iyer, Kshitiz
  Malik, Kuenley Chiu, Kunal Bhalla, Kushal Lakhotia, Lauren Rantala-Yeary,
  Laurens van~der Maaten, Lawrence Chen, Liang Tan, Liz Jenkins, Louis Martin,
  Lovish Madaan, Lubo Malo, Lukas Blecher, Lukas Landzaat, Luke de Oliveira,
  Madeline Muzzi, Mahesh Pasupuleti, Mannat Singh, Manohar Paluri, Marcin
  Kardas, Maria Tsimpoukelli, Mathew Oldham, Mathieu Rita, Maya Pavlova,
  Melanie Kambadur, Mike Lewis, Min Si, Mitesh~Kumar Singh, Mona Hassan, Naman
  Goyal, Narjes Torabi, Nikolay Bashlykov, Nikolay Bogoychev, Niladri
  Chatterji, Ning Zhang, Olivier Duchenne, Onur Çelebi, Patrick Alrassy,
  Pengchuan Zhang, Pengwei Li, Petar Vasic, Peter Weng, Prajjwal Bhargava,
  Pratik Dubal, Praveen Krishnan, Punit~Singh Koura, Puxin Xu, Qing He,
  Qingxiao Dong, Ragavan Srinivasan, Raj Ganapathy, Ramon Calderer,
  Ricardo~Silveira Cabral, Robert Stojnic, Roberta Raileanu, Rohan Maheswari,
  Rohit Girdhar, Rohit Patel, Romain Sauvestre, Ronnie Polidoro, Roshan
  Sumbaly, Ross Taylor, Ruan Silva, Rui Hou, Rui Wang, Saghar Hosseini, Sahana
  Chennabasappa, Sanjay Singh, Sean Bell, Seohyun~Sonia Kim, Sergey Edunov,
  Shaoliang Nie, Sharan Narang, Sharath Raparthy, Sheng Shen, Shengye Wan,
  Shruti Bhosale, Shun Zhang, Simon Vandenhende, Soumya Batra, Spencer Whitman,
  Sten Sootla, Stephane Collot, Suchin Gururangan, Sydney Borodinsky, Tamar
  Herman, Tara Fowler, Tarek Sheasha, Thomas Georgiou, Thomas Scialom, Tobias
  Speckbacher, Todor Mihaylov, Tong Xiao, Ujjwal Karn, Vedanuj Goswami, Vibhor
  Gupta, Vignesh Ramanathan, Viktor Kerkez, Vincent Gonguet, Virginie Do, Vish
  Vogeti, Vítor Albiero, Vladan Petrovic, Weiwei Chu, Wenhan Xiong, Wenyin Fu,
  Whitney Meers, Xavier Martinet, Xiaodong Wang, Xiaofang Wang, Xiaoqing~Ellen
  Tan, Xide Xia, Xinfeng Xie, Xuchao Jia, Xuewei Wang, Yaelle Goldschlag,
  Yashesh Gaur, Yasmine Babaei, Yi Wen, Yiwen Song, Yuchen Zhang, Yue Li,
  Yuning Mao, Zacharie~Delpierre Coudert, Zheng Yan, Zhengxing Chen, Zoe
  Papakipos, Aaditya Singh, Aayushi Srivastava, Abha Jain, Adam Kelsey, Adam
  Shajnfeld, Adithya Gangidi, Adolfo Victoria, Ahuva Goldstand, Ajay Menon,
  Ajay Sharma, Alex Boesenberg, Alexei Baevski, Allie Feinstein, Amanda Kallet,
  Amit Sangani, Amos Teo, Anam Yunus, Andrei Lupu, Andres Alvarado, Andrew
  Caples, Andrew Gu, Andrew Ho, Andrew Poulton, Andrew Ryan, Ankit Ramchandani,
  Annie Dong, Annie Franco, Anuj Goyal, Aparajita Saraf, Arkabandhu Chowdhury,
  Ashley Gabriel, Ashwin Bharambe, Assaf Eisenman, Azadeh Yazdan, Beau James,
  Ben Maurer, Benjamin Leonhardi, Bernie Huang, Beth Loyd, Beto~De Paola,
  Bhargavi Paranjape, Bing Liu, Bo Wu, Boyu Ni, Braden Hancock, Bram Wasti,
  Brandon Spence, Brani Stojkovic, Brian Gamido, Britt Montalvo, Carl Parker,
  Carly Burton, Catalina Mejia, Ce Liu, Changhan Wang, Changkyu Kim, Chao Zhou,
  Chester Hu, Ching-Hsiang Chu, Chris Cai, Chris Tindal, Christoph
  Feichtenhofer, Cynthia Gao, Damon Civin, Dana Beaty, Daniel Kreymer, Daniel
  Li, David Adkins, David Xu, Davide Testuggine, Delia David, Devi Parikh,
  Diana Liskovich, Didem Foss, Dingkang Wang, Duc Le, Dustin Holland, Edward
  Dowling, Eissa Jamil, Elaine Montgomery, Eleonora Presani, Emily Hahn, Emily
  Wood, Eric-Tuan Le, Erik Brinkman, Esteban Arcaute, Evan Dunbar, Evan
  Smothers, Fei Sun, Felix Kreuk, Feng Tian, Filippos Kokkinos, Firat Ozgenel,
  Francesco Caggioni, Frank Kanayet, Frank Seide, Gabriela~Medina Florez,
  Gabriella Schwarz, Gada Badeer, Georgia Swee, Gil Halpern, Grant Herman,
  Grigory Sizov, Guangyi, Zhang, Guna Lakshminarayanan, Hakan Inan, Hamid
  Shojanazeri, Han Zou, Hannah Wang, Hanwen Zha, Haroun Habeeb, Harrison
  Rudolph, Helen Suk, Henry Aspegren, Hunter Goldman, Hongyuan Zhan, Ibrahim
  Damlaj, Igor Molybog, Igor Tufanov, Ilias Leontiadis, Irina-Elena Veliche,
  Itai Gat, Jake Weissman, James Geboski, James Kohli, Janice Lam, Japhet
  Asher, Jean-Baptiste Gaya, Jeff Marcus, Jeff Tang, Jennifer Chan, Jenny Zhen,
  Jeremy Reizenstein, Jeremy Teboul, Jessica Zhong, Jian Jin, Jingyi Yang, Joe
  Cummings, Jon Carvill, Jon Shepard, Jonathan McPhie, Jonathan Torres, Josh
  Ginsburg, Junjie Wang, Kai Wu, Kam~Hou U, Karan Saxena, Kartikay Khandelwal,
  Katayoun Zand, Kathy Matosich, Kaushik Veeraraghavan, Kelly Michelena, Keqian
  Li, Kiran Jagadeesh, Kun Huang, Kunal Chawla, Kyle Huang, Lailin Chen,
  Lakshya Garg, Lavender A, Leandro Silva, Lee Bell, Lei Zhang, Liangpeng Guo,
  Licheng Yu, Liron Moshkovich, Luca Wehrstedt, Madian Khabsa, Manav Avalani,
  Manish Bhatt, Martynas Mankus, Matan Hasson, Matthew Lennie, Matthias Reso,
  Maxim Groshev, Maxim Naumov, Maya Lathi, Meghan Keneally, Miao Liu,
  Michael~L. Seltzer, Michal Valko, Michelle Restrepo, Mihir Patel, Mik
  Vyatskov, Mikayel Samvelyan, Mike Clark, Mike Macey, Mike Wang, Miquel~Jubert
  Hermoso, Mo Metanat, Mohammad Rastegari, Munish Bansal, Nandhini Santhanam,
  Natascha Parks, Natasha White, Navyata Bawa, Nayan Singhal, Nick Egebo,
  Nicolas Usunier, Nikhil Mehta, Nikolay~Pavlovich Laptev, Ning Dong, Norman
  Cheng, Oleg Chernoguz, Olivia Hart, Omkar Salpekar, Ozlem Kalinli, Parkin
  Kent, Parth Parekh, Paul Saab, Pavan Balaji, Pedro Rittner, Philip Bontrager,
  Pierre Roux, Piotr Dollar, Polina Zvyagina, Prashant Ratanchandani, Pritish
  Yuvraj, Qian Liang, Rachad Alao, Rachel Rodriguez, Rafi Ayub, Raghotham
  Murthy, Raghu Nayani, Rahul Mitra, Rangaprabhu Parthasarathy, Raymond Li,
  Rebekkah Hogan, Robin Battey, Rocky Wang, Russ Howes, Ruty Rinott, Sachin
  Mehta, Sachin Siby, Sai~Jayesh Bondu, Samyak Datta, Sara Chugh, Sara Hunt,
  Sargun Dhillon, Sasha Sidorov, Satadru Pan, Saurabh Mahajan, Saurabh Verma,
  Seiji Yamamoto, Sharadh Ramaswamy, Shaun Lindsay, Shaun Lindsay, Sheng Feng,
  Shenghao Lin, Shengxin~Cindy Zha, Shishir Patil, Shiva Shankar, Shuqiang
  Zhang, Shuqiang Zhang, Sinong Wang, Sneha Agarwal, Soji Sajuyigbe, Soumith
  Chintala, Stephanie Max, Stephen Chen, Steve Kehoe, Steve Satterfield,
  Sudarshan Govindaprasad, Sumit Gupta, Summer Deng, Sungmin Cho, Sunny Virk,
  Suraj Subramanian, Sy Choudhury, Sydney Goldman, Tal Remez, Tamar Glaser,
  Tamara Best, Thilo Koehler, Thomas Robinson, Tianhe Li, Tianjun Zhang, Tim
  Matthews, Timothy Chou, Tzook Shaked, Varun Vontimitta, Victoria Ajayi,
  Victoria Montanez, Vijai Mohan, Vinay~Satish Kumar, Vishal Mangla, Vlad
  Ionescu, Vlad Poenaru, Vlad~Tiberiu Mihailescu, Vladimir Ivanov, Wei Li,
  Wenchen Wang, Wenwen Jiang, Wes Bouaziz, Will Constable, Xiaocheng Tang,
  Xiaojian Wu, Xiaolan Wang, Xilun Wu, Xinbo Gao, Yaniv Kleinman, Yanjun Chen,
  Ye Hu, Ye Jia, Ye Qi, Yenda Li, Yilin Zhang, Ying Zhang, Yossi Adi, Youngjin
  Nam, Yu, Wang, Yu Zhao, Yuchen Hao, Yundi Qian, Yunlu Li, Yuzi He, Zach Rait,
  Zachary DeVito, Zef Rosnbrick, Zhaoduo Wen, Zhenyu Yang, Zhiwei Zhao, and
  Zhiyu Ma.
\newblock The llama 3 herd of models, 2024.

\bibitem[Grauman et~al.(2022)Grauman, Westbury, Byrne, Chavis, Furnari,
  Girdhar, Hamburger, Jiang, Liu, Liu, Martin, Nagarajan, Radosavovic,
  Ramakrishnan, Ryan, Sharma, Wray, Xu, Xu, Zhao, Bansal, Batra, Cartillier,
  Crane, Do, Doulaty, Erapalli, Feichtenhofer, Fragomeni, Fu, Gebreselasie,
  Gonz\'alez, Hillis, Huang, Huang, Jia, Khoo, Kol\'a\v{r}, Kottur, Kumar,
  Landini, Li, Li, Li, Mangalam, Modhugu, Munro, Murrell, Nishiyasu, Price,
  Ruiz, Ramazanova, Sari, Somasundaram, Southerland, Sugano, Tao, Vo, Wang, Wu,
  Yagi, Zhao, Zhu, Arbel\'aez, Crandall, Damen, Farinella, Fuegen, Ghanem,
  Ithapu, Jawahar, Joo, Kitani, Li, Newcombe, Oliva, Park, Rehg, Sato, Shi,
  Shou, Torralba, Torresani, Yan, and Malik]{Grauman_2022_CVPR}
Kristen Grauman, Andrew Westbury, Eugene Byrne, Zachary Chavis, Antonino
  Furnari, Rohit Girdhar, Jackson Hamburger, Hao Jiang, Miao Liu, Xingyu Liu,
  Miguel Martin, Tushar Nagarajan, Ilija Radosavovic, Santhosh~Kumar
  Ramakrishnan, Fiona Ryan, Jayant Sharma, Michael Wray, Mengmeng Xu,
  Eric~Zhongcong Xu, Chen Zhao, Siddhant Bansal, Dhruv Batra, Vincent
  Cartillier, Sean Crane, Tien Do, Morrie Doulaty, Akshay Erapalli, Christoph
  Feichtenhofer, Adriano Fragomeni, Qichen Fu, Abrham Gebreselasie, Cristina
  Gonz\'alez, James Hillis, Xuhua Huang, Yifei Huang, Wenqi Jia, Weslie Khoo,
  J\'achym Kol\'a\v{r}, Satwik Kottur, Anurag Kumar, Federico Landini, Chao Li,
  Yanghao Li, Zhenqiang Li, Karttikeya Mangalam, Raghava Modhugu, Jonathan
  Munro, Tullie Murrell, Takumi Nishiyasu, Will Price, Paola Ruiz, Merey
  Ramazanova, Leda Sari, Kiran Somasundaram, Audrey Southerland, Yusuke Sugano,
  Ruijie Tao, Minh Vo, Yuchen Wang, Xindi Wu, Takuma Yagi, Ziwei Zhao, Yunyi
  Zhu, Pablo Arbel\'aez, David Crandall, Dima Damen, Giovanni~Maria Farinella,
  Christian Fuegen, Bernard Ghanem, Vamsi~Krishna Ithapu, C.~V. Jawahar,
  Hanbyul Joo, Kris Kitani, Haizhou Li, Richard Newcombe, Aude Oliva, Hyun~Soo
  Park, James~M. Rehg, Yoichi Sato, Jianbo Shi, Mike~Zheng Shou, Antonio
  Torralba, Lorenzo Torresani, Mingfei Yan, and Jitendra Malik.
\newblock Ego4d: Around the world in 3,000 hours of egocentric video.
\newblock In \emph{Proceedings of the IEEE/CVF Conference on Computer Vision
  and Pattern Recognition (CVPR)}, pages 18995--19012, 2022.

\bibitem[Green(2000)]{Green_2000}
Marc Green.
\newblock “how long does it take to stop?” methodological analysis of
  driver perception-brake times.
\newblock \emph{Transportation Human Factors}, 2\penalty0 (3):\penalty0
  195–216, 2000.

\bibitem[Hong et~al.(2025)Hong, Cheng, Yang, Wang, Wang, Gu, Huang, Dong, and
  Tang]{Hong_2025_CVPR}
Wenyi Hong, Yean Cheng, Zhuoyi Yang, Weihan Wang, Lefan Wang, Xiaotao Gu, Shiyu
  Huang, Yuxiao Dong, and Jie Tang.
\newblock Motionbench: Benchmarking and improving fine-grained video motion
  understanding for vision language models.
\newblock In \emph{Proceedings of the IEEE/CVF Conference on Computer Vision
  and Pattern Recognition (CVPR)}, pages 8450--8460, 2025.

\bibitem[Huang et~al.(2026)Huang, Chen, Chen, Liu, and
  Lin]{huang2026egvqabenchmarkingverifiablevideo}
Linpeng Huang, Weixing Chen, Zexin Chen, Yang Liu, and Liang Lin.
\newblock Eg-vqa: Benchmarking verifiable video question answering with
  grounded temporal evidence, 2026.

\bibitem[Jung et~al.(2026)Jung, Lee, Kim, Han, and
  Oh]{jung2026egogapbenchbenchmarkingegocentricaction}
Jihyeok Jung, Jeewu Lee, Sanghyeop Kim, Chanhee Han, and Seong~Joon Oh.
\newblock Egogapbench: Benchmarking egocentric action selection in multi-agent
  scenes, 2026.

\bibitem[Karamouzas et~al.(2014)Karamouzas, Skinner, and
  Guy]{PhysRevLett.113.238701}
Ioannis Karamouzas, Brian Skinner, and Stephen~J. Guy.
\newblock Universal power law governing pedestrian interactions.
\newblock \emph{Phys. Rev. Lett.}, 113:\penalty0 238701, 2014.

\bibitem[Kung et~al.(2024)Kung, Yang, Pao, Lu, Chen, Lu, and Chen]{10610270}
Chi-Hsi Kung, Chieh-Chi Yang, Pang-Yuan Pao, Shu-Wei Lu, Pin-Lun Chen,
  Hsin-Cheng Lu, and Yi-Ting Chen.
\newblock Riskbench: A scenario-based benchmark for risk identification.
\newblock In \emph{2024 IEEE International Conference on Robotics and
  Automation (ICRA)}, pages 14800--14807, 2024.

\bibitem[Lei et~al.(2026)Lei, Li, Zhang, Hua, Li, and
  Liu]{lei2026egosatcomprehensivebenchmarkegocentric}
Yijia Lei, Jinzhao Li, Yichi Zhang, Jiacheng Hua, Yin Li, and Miao Liu.
\newblock Egosat: A comprehensive benchmark of egocentric streaming interaction
  understanding, 2026.

\bibitem[Li et~al.(2026)Li, Han, Hsu, Liang, Dhawan, Wu, Yang, and
  Wang]{li2026sawbench}
Chuhan Li, Rilyn~R. Han, Joy Hsu, Yongyuan Liang, Rajiv Dhawan, Jiajun Wu,
  Ming-Hsuan Yang, and Xin~Eric Wang.
\newblock {SAW}-bench: Learning situated awareness in the real world.
\newblock In \emph{Forty-third International Conference on Machine Learning},
  2026.

\bibitem[Li et~al.(2024)Li, Wang, He, Li, Wang, Liu, Wang, Xu, Chen, Luo, Wang,
  and Qiao]{Li_2024_CVPR}
Kunchang Li, Yali Wang, Yinan He, Yizhuo Li, Yi Wang, Yi Liu, Zun Wang, Jilan
  Xu, Guo Chen, Ping Luo, Limin Wang, and Yu Qiao.
\newblock Mvbench: A comprehensive multi-modal video understanding benchmark.
\newblock In \emph{Proceedings of the IEEE/CVF Conference on Computer Vision
  and Pattern Recognition (CVPR)}, pages 22195--22206, 2024.

\bibitem[Liu et~al.(2024)Liu, Li, Liu, Wang, Ren, Li, Chen, Sun, and
  Hou]{liu-etal-2024-tempcompass}
Yuanxin Liu, Shicheng Li, Yi Liu, Yuxiang Wang, Shuhuai Ren, Lei Li, Sishuo
  Chen, Xu Sun, and Lu Hou.
\newblock {T}emp{C}ompass: Do video {LLM}s really understand videos?
\newblock In \emph{Findings of the Association for Computational Linguistics:
  ACL 2024}, pages 8731--8772, Bangkok, Thailand, 2024. Association for
  Computational Linguistics.

\bibitem[Liu et~al.(2026)Liu, Guo, Li, Zhen, He, Ji, Ren, Zhang, Lu, and
  Liu]{liu2026thinkingstreamingvideo}
Zikang Liu, Longteng Guo, Handong Li, Ru Zhen, Xingjian He, Ruyi Ji, Xiaoming
  Ren, Yanhao Zhang, Haonan Lu, and Jing Liu.
\newblock Thinking in streaming video, 2026.

\bibitem[Malla et~al.(2023)Malla, Choi, Dwivedi, Choi, and Li]{Malla_2023_WACV}
Srikanth Malla, Chiho Choi, Isht Dwivedi, Joon~Hee Choi, and Jiachen Li.
\newblock Drama: Joint risk localization and captioning in driving.
\newblock In \emph{Proceedings of the IEEE/CVF Winter Conference on
  Applications of Computer Vision (WACV)}, pages 1043--1052, 2023.

\bibitem[Mangalam et~al.(2023)Mangalam, Akshulakov, and
  Malik]{NEURIPS2023_90ce332a}
Karttikeya Mangalam, Raiymbek Akshulakov, and Jitendra Malik.
\newblock Egoschema: A diagnostic benchmark for very long-form video language
  understanding.
\newblock In \emph{Advances in Neural Information Processing Systems}, pages
  46212--46244. Curran Associates, Inc., 2023.

\bibitem[Nagrani et~al.(2026)Nagrani, Uijlings, Buch, Weyand, Vijayanarasimhan,
  Hu, Mehran, A~Ross, and Schmid]{Nagrani_2026_CVPR}
Arsha Nagrani, Jasper Uijlings, Shyamal Buch, Tobias Weyand, Sudheendra
  Vijayanarasimhan, Bo Hu, Ramin Mehran, David A~Ross, and Cordelia Schmid.
\newblock Minerva-ego: Spatiotemporal hints for egocentric video understanding.
\newblock In \emph{Proceedings of the IEEE/CVF Conference on Computer Vision
  and Pattern Recognition (CVPR)}, pages 38859--38869, 2026.

\bibitem[Pundlik et~al.(2021)Pundlik, Baliutaviciute, Moharrer, Bowers, and
  Luo]{pundlik2021homeuse}
Shrinivas Pundlik, Vilte Baliutaviciute, Mojtaba Moharrer, Alex~R. Bowers, and
  Gang Luo.
\newblock Home-use evaluation of a wearable collision warning device for
  individuals with severe vision impairments: A randomized clinical trial.
\newblock \emph{JAMA Ophthalmology}, 139\penalty0 (9):\penalty0 998--1005,
  2021.

\bibitem[{Qwen Team}(2026)]{qwen38}
{Qwen Team}.
\newblock Qwen3.8-max: A new bar for coding and cowork, 2026.

\bibitem[Ran et~al.(2026)Ran, Ou, Li, Tong, Guo, Guo, Wang, and
  Lu]{ran2026egoprobenchbenchmarkingpersonalizedproactive}
Dongchuan Ran, Linyu Ou, Xueheng Li, Wenwen Tong, Chenxu Guo, Hewei Guo,
  Kaibing Wang, and Lewei Lu.
\newblock Egopro-bench: Benchmarking personalized proactive interaction in
  egocentric video streams, 2026.

\bibitem[Shangguan et~al.(2025)Shangguan, Li, Ding, Zheng, Zhao, Fitzgerald,
  and Cohan]{ICLR2025_16ba99f2}
Ziyao Shangguan, Chuhan Li, Yuxuan Ding, Yanan Zheng, Yilun Zhao, Tesca
  Fitzgerald, and Arman Cohan.
\newblock Tomato: Assessing visual temporal reasoning capabilities in
  multimodal foundation models.
\newblock In \emph{International Conference on Learning Representations}, pages
  7593--7734, 2025.

\bibitem[Singh et~al.(2026)Singh, Fry, Perelman, Tart, Ganesh, El-Kishky,
  McLaughlin, Low, Ostrow, Ananthram, Nathan, Luo, Helyar, Madry, Efremov,
  Spyra, Baker-Whitcomb, Beutel, Karpenko, Makelov, Neitz, Wei, Barr,
  Kirchmeyer, Ivanov, Christakis, Gillespie, Tam, Bennett, Wan, Huang,
  Sandjideh, Yang, Kumar, Saraiva, Vallone, Gheorghe, Garcia, Braunstein, Liu,
  Schmidt, Mereskin, Mishchenko, Applebaum, Rogerson, Rajan, Wei, Kotha,
  Srivastava, Agrawal, Vijayvergiya, Tyra, Nair, Nayak, Eggers, Ji, Hoover,
  Chen, Chen, Barak, Minaiev, Hao, Baker, Lightcap, McKinzie, Wang, Quinn,
  Fioca, Hsu, Yang, Yu, Zhang, Brenner, Zetino, Raymond, Lugaresi, Paz, Hudson,
  Whitney, Li, Chen, Cole, Voss, Ding, Shen, Huang, Colby, Hallacy, Koch, Lu,
  Kaplan, Kim, Minott-Henriques, Frey, Yu, Czarnecki, Reid, Wei, Decareaux,
  Scheau, Zhang, Forbes, Tang, Goldberg, Roberts, Palmie, Kappler, Levine,
  Wright, Leo, Lin, Robinson, Grabb, Chen, Lim, Salama, Bhattacharjee, Tsipras,
  Li, Yu, Strouse, Williams, Hunn, Bayes, Arbus, Akyurek, Le, Widmann, Yani,
  Proehl, Sert, Cheung, Schwartz, Han, Jiang, Mitchell, Sigler, Wallace,
  Ritter, Kavanaugh, Mays, Nikishin, Li, Such, de~Avila Belbute~Peres, Raso,
  Bekerman, Tsimpourlas, Chantzis, Song, Zhang, Raila, McGrath, Briggs, Yang,
  Parascandolo, Chabot, Kim, Zhao, Valiant, Leclerc, Salman, Wang, Sheng,
  Jiang, Wang, Jin, Sikchi, Schmidt, Aspegren, Chen, Qiu, Lightman, Covert,
  Kivlichan, Silber, Sohl, Hammoud, Clavera, Lan, Akkaya, Kostrikov, Kofman,
  Etinger, Singal, Hehir, Huh, Pan, Wilczynski, Pachocki, Lee, Quinn, Kiros,
  Kalra, Samaroo, Wang, Wolfe, Chen, Wang, Harb, Han, Wang, Zhao, Chen, Yang,
  Tworek, Chand, Landon, Liang, Lin, Liu, Wang, Tang, Yin, Jang, Morris, Flynn,
  Ferstad, Heidecke, Fishbein, Hallman, Grant, Chien, Gordon, Park, Liss,
  Kraaijeveld, Guay, Mo, Lawson, McGrath, Vendrow, Jiao, Lee, Steele, Wang,
  Mao, Chen, Hayashi, Xiao, Salahi, Wu, Sekhri, Sharma, Singhal, Li, Nguyen,
  Gu-Lemberg, King, Liu, Stone, Yu, Ying, Georgiev, Lim, Tirumala, Miller,
  Ahmad, Lv, Clare, Fauconnet, Itow, Yang, Romaniuk, Anise, Byron, Pathak,
  Maksin, Lo, Ho, Jing, Wu, Xiong, Mamitsuka, Yang, McCallum, Held, Bourgeois,
  Engstrom, Kuhn, Feuvrier, Zhang, Switzer, Kondraciuk, Kaiser, Joglekar,
  Singh, Shah, Stratta, Williams, Chen, Sun, Cayton, Li, Zhang, Aljubeh,
  Nichols, Haines, Schwarzer, Gupta, Shah, Guan, Huang, Dong, Wang, Glaese,
  Carroll, Lampe, Malek, Sharman, Zhang, Wang, Pokrass, Florian, Pavlov, Wang,
  Chen, Wang, Feng, Bavarian, Lin, Abdool, Rohaninejad, Soto, Staudacher,
  LaFontaine, Marwell, Liu, Preston, Turley, Ansman, Blades, Pancha, Mikhaylin,
  Felix, Handa, Rai, Keskar, Brown, Nachum, Boiko, Murk, Watkins, Gleeson,
  Mishkin, Lesiewicz, Baltescu, Belov, Zhokhov, Pronin, Guo, Thacker, Liu,
  Yuan, Liu, Dias, Puckett, Arora, Mullapudi, Gaon, Miyara, Song, Aggarwal,
  Marsan, Yemiru, Xiong, Kshirsagar, Nuttall, Tsiupa, Eldan, Wang, James, Ziv,
  Shu, Nigmatullin, Jain, Talaie, Altman, Arnesen, Toizer, Toyer, Miserendino,
  Agarwal, Yoo, Heon, Ethersmith, Grove, Taylor, Bubeck, Banesiu, Amdo, Zhao,
  Wu, Santurkar, Zhao, Chaudhuri, Krishnaswamy, Shuaiqi, Xia, Cheng, Anadkat,
  Fishman, Tobin, Fu, Jain, Mei, Egoian, Kim, Golden, Mah, Lin, Imm, Sharpe,
  Yadlowsky, Choudhry, Eum, Sanjeev, Khan, Stramer, Wang, Xin, Gogineni,
  Christianson, Sanders, Patwardhan, Degry, Shadwell, Fu, Gao, Garipov,
  Sriskandarajah, Sherbakov, Korbak, Kaftan, Hiratsuka, Wang, Song, Zhao,
  Peterson, Kharitonov, Chernova, Kosaraju, Kuo, Pong, Verma, Petrov, Jiang,
  Zhang, Zhou, Xie, Zhan, McCabe, DePue, Ellsworth, Bain, Thompson, Chen, Qi,
  Xiang, Shi, Dubois, Yu, Khakbaz, Wu, Qian, Lee, Chen, Zhang, Xiong, Tian,
  Cha, Bai, Yang, Yuan, Li, Zhang, Yang, Jin, Jiang, Wang, Wang, Liu,
  Stubenvoll, Dou, Wu, and Wang]{singh2026openaigpt5card}
Aaditya Singh, Adam Fry, Adam Perelman, Adam Tart, Adi Ganesh, Ahmed El-Kishky,
  Aidan McLaughlin, Aiden Low, AJ Ostrow, Akhila Ananthram, Akshay Nathan, Alan
  Luo, Alec Helyar, Aleksander Madry, Aleksandr Efremov, Aleksandra Spyra, Alex
  Baker-Whitcomb, Alex Beutel, Alex Karpenko, Alex Makelov, Alex Neitz, Alex
  Wei, Alexandra Barr, Alexandre Kirchmeyer, Alexey Ivanov, Alexi Christakis,
  Alistair Gillespie, Allison Tam, Ally Bennett, Alvin Wan, Alyssa Huang,
  Amy~McDonald Sandjideh, Amy Yang, Ananya Kumar, Andre Saraiva, Andrea
  Vallone, Andrei Gheorghe, Andres~Garcia Garcia, Andrew Braunstein, Andrew
  Liu, Andrew Schmidt, Andrey Mereskin, Andrey Mishchenko, Andy Applebaum, Andy
  Rogerson, Ann Rajan, Annie Wei, Anoop Kotha, Anubha Srivastava, Anushree
  Agrawal, Arun Vijayvergiya, Ashley Tyra, Ashvin Nair, Avi Nayak, Ben Eggers,
  Bessie Ji, Beth Hoover, Bill Chen, Blair Chen, Boaz Barak, Borys Minaiev,
  Botao Hao, Bowen Baker, Brad Lightcap, Brandon McKinzie, Brandon Wang,
  Brendan Quinn, Brian Fioca, Brian Hsu, Brian Yang, Brian Yu, Brian Zhang,
  Brittany Brenner, Callie~Riggins Zetino, Cameron Raymond, Camillo Lugaresi,
  Carolina Paz, Cary Hudson, Cedric Whitney, Chak Li, Charles Chen, Charlotte
  Cole, Chelsea Voss, Chen Ding, Chen Shen, Chengdu Huang, Chris Colby, Chris
  Hallacy, Chris Koch, Chris Lu, Christina Kaplan, Christina Kim, CJ
  Minott-Henriques, Cliff Frey, Cody Yu, Coley Czarnecki, Colin Reid, Colin
  Wei, Cory Decareaux, Cristina Scheau, Cyril Zhang, Cyrus Forbes, Da Tang,
  Dakota Goldberg, Dan Roberts, Dana Palmie, Daniel Kappler, Daniel Levine,
  Daniel Wright, Dave Leo, David Lin, David Robinson, Declan Grabb, Derek Chen,
  Derek Lim, Derek Salama, Dibya Bhattacharjee, Dimitris Tsipras, Dinghua Li,
  Dingli Yu, DJ Strouse, Drew Williams, Dylan Hunn, Ed Bayes, Edwin Arbus, Ekin
  Akyurek, Elaine~Ya Le, Elana Widmann, Eli Yani, Elizabeth Proehl, Enis Sert,
  Enoch Cheung, Eri Schwartz, Eric Han, Eric Jiang, Eric Mitchell, Eric Sigler,
  Eric Wallace, Erik Ritter, Erin Kavanaugh, Evan Mays, Evgenii Nikishin,
  Fangyuan Li, Felipe~Petroski Such, Filipe de Avila Belbute~Peres, Filippo
  Raso, Florent Bekerman, Foivos Tsimpourlas, Fotis Chantzis, Francis Song,
  Francis Zhang, Gaby Raila, Garrett McGrath, Gary Briggs, Gary Yang,
  Giambattista Parascandolo, Gildas Chabot, Grace Kim, Grace Zhao, Gregory
  Valiant, Guillaume Leclerc, Hadi Salman, Hanson Wang, Hao Sheng, Haoming
  Jiang, Haoyu Wang, Haozhun Jin, Harshit Sikchi, Heather Schmidt, Henry
  Aspegren, Honglin Chen, Huida Qiu, Hunter Lightman, Ian Covert, Ian
  Kivlichan, Ian Silber, Ian Sohl, Ibrahim Hammoud, Ignasi Clavera, Ikai Lan,
  Ilge Akkaya, Ilya Kostrikov, Irina Kofman, Isak Etinger, Ishaan Singal,
  Jackie Hehir, Jacob Huh, Jacqueline Pan, Jake Wilczynski, Jakub Pachocki,
  James Lee, James Quinn, Jamie Kiros, Janvi Kalra, Jasmyn Samaroo, Jason Wang,
  Jason Wolfe, Jay Chen, Jay Wang, Jean Harb, Jeffrey Han, Jeffrey Wang,
  Jennifer Zhao, Jeremy Chen, Jerene Yang, Jerry Tworek, Jesse Chand, Jessica
  Landon, Jessica Liang, Ji Lin, Jiancheng Liu, Jianfeng Wang, Jie Tang, Jihan
  Yin, Joanne Jang, Joel Morris, Joey Flynn, Johannes Ferstad, Johannes
  Heidecke, John Fishbein, John Hallman, Jonah Grant, Jonathan Chien, Jonathan
  Gordon, Jongsoo Park, Jordan Liss, Jos Kraaijeveld, Joseph Guay, Joseph Mo,
  Josh Lawson, Josh McGrath, Joshua Vendrow, Joy Jiao, Julian Lee, Julie
  Steele, Julie Wang, Junhua Mao, Kai Chen, Kai Hayashi, Kai Xiao, Kamyar
  Salahi, Kan Wu, Karan Sekhri, Karan Sharma, Karan Singhal, Karen Li, Kenny
  Nguyen, Keren Gu-Lemberg, Kevin King, Kevin Liu, Kevin Stone, Kevin Yu,
  Kristen Ying, Kristian Georgiev, Kristie Lim, Kushal Tirumala, Kyle Miller,
  Lama Ahmad, Larry Lv, Laura Clare, Laurance Fauconnet, Lauren Itow, Lauren
  Yang, Laurentia Romaniuk, Leah Anise, Lee Byron, Leher Pathak, Leon Maksin,
  Leyan Lo, Leyton Ho, Li Jing, Liang Wu, Liang Xiong, Lien Mamitsuka, Lin
  Yang, Lindsay McCallum, Lindsey Held, Liz Bourgeois, Logan Engstrom, Lorenz
  Kuhn, Louis Feuvrier, Lu Zhang, Lucas Switzer, Lukas Kondraciuk, Lukasz
  Kaiser, Manas Joglekar, Mandeep Singh, Mandip Shah, Manuka Stratta, Marcus
  Williams, Mark Chen, Mark Sun, Marselus Cayton, Martin Li, Marvin Zhang,
  Marwan Aljubeh, Matt Nichols, Matthew Haines, Max Schwarzer, Mayank Gupta,
  Meghan Shah, Melody~Y. Guan, Melody Huang, Meng Dong, Mengqing Wang, Mia
  Glaese, Micah Carroll, Michael Lampe, Michael Malek, Michael Sharman, Michael
  Zhang, Michele Wang, Michelle Pokrass, Mihai Florian, Mikhail Pavlov, Miles
  Wang, Ming Chen, Mingxuan Wang, Minnia Feng, Mo Bavarian, Molly Lin, Moose
  Abdool, Mostafa Rohaninejad, Nacho Soto, Natalie Staudacher, Natan
  LaFontaine, Nathan Marwell, Nelson Liu, Nick Preston, Nick Turley, Nicklas
  Ansman, Nicole Blades, Nikil Pancha, Nikita Mikhaylin, Niko Felix, Nikunj
  Handa, Nishant Rai, Nitish Keskar, Noam Brown, Ofir Nachum, Oleg Boiko, Oleg
  Murk, Olivia Watkins, Oona Gleeson, Pamela Mishkin, Patryk Lesiewicz, Paul
  Baltescu, Pavel Belov, Peter Zhokhov, Philip Pronin, Phillip Guo, Phoebe
  Thacker, Qi Liu, Qiming Yuan, Qinghua Liu, Rachel Dias, Rachel Puckett, Rahul
  Arora, Ravi~Teja Mullapudi, Raz Gaon, Reah Miyara, Rennie Song, Rishabh
  Aggarwal, RJ Marsan, Robel Yemiru, Robert Xiong, Rohan Kshirsagar, Rohan
  Nuttall, Roman Tsiupa, Ronen Eldan, Rose Wang, Roshan James, Roy Ziv, Rui
  Shu, Ruslan Nigmatullin, Saachi Jain, Saam Talaie, Sam Altman, Sam Arnesen,
  Sam Toizer, Sam Toyer, Samuel Miserendino, Sandhini Agarwal, Sarah Yoo,
  Savannah Heon, Scott Ethersmith, Sean Grove, Sean Taylor, Sebastien Bubeck,
  Sever Banesiu, Shaokyi Amdo, Shengjia Zhao, Sherwin Wu, Shibani Santurkar,
  Shiyu Zhao, Shraman~Ray Chaudhuri, Shreyas Krishnaswamy, Shuaiqi, Xia,
  Shuyang Cheng, Shyamal Anadkat, Simón~Posada Fishman, Simon Tobin, Siyuan
  Fu, Somay Jain, Song Mei, Sonya Egoian, Spencer Kim, Spug Golden, SQ Mah,
  Steph Lin, Stephen Imm, Steve Sharpe, Steve Yadlowsky, Sulman Choudhry,
  Sungwon Eum, Suvansh Sanjeev, Tabarak Khan, Tal Stramer, Tao Wang, Tao Xin,
  Tarun Gogineni, Taya Christianson, Ted Sanders, Tejal Patwardhan, Thomas
  Degry, Thomas Shadwell, Tianfu Fu, Tianshi Gao, Timur Garipov, Tina
  Sriskandarajah, Toki Sherbakov, Tomek Korbak, Tomer Kaftan, Tomo Hiratsuka,
  Tongzhou Wang, Tony Song, Tony Zhao, Troy Peterson, Val Kharitonov, Victoria
  Chernova, Vineet Kosaraju, Vishal Kuo, Vitchyr Pong, Vivek Verma, Vlad
  Petrov, Wanning Jiang, Weixing Zhang, Wenda Zhou, Wenlei Xie, Wenting Zhan,
  Wes McCabe, Will DePue, Will Ellsworth, Wulfie Bain, Wyatt Thompson,
  Xiangning Chen, Xiangyu Qi, Xin Xiang, Xinwei Shi, Yann Dubois, Yaodong Yu,
  Yara Khakbaz, Yifan Wu, Yilei Qian, Yin~Tat Lee, Yinbo Chen, Yizhen Zhang,
  Yizhong Xiong, Yonglong Tian, Young Cha, Yu Bai, Yu Yang, Yuan Yuan, Yuanzhi
  Li, Yufeng Zhang, Yuguang Yang, Yujia Jin, Yun Jiang, Yunyun Wang, Yushi
  Wang, Yutian Liu, Zach Stubenvoll, Zehao Dou, Zheng Wu, and Zhigang Wang.
\newblock Openai gpt-5 system card, 2026.

\bibitem[Tang et~al.(2026)Tang, Wang, Lin, Li, Salvo, and
  Zhang]{tang2026egoeverythingbenchmarkhumanbehavior}
Qiance Tang, Ziqi Wang, Jieyu Lin, Ziyun Li, Barbara~De Salvo, and Sai~Qian
  Zhang.
\newblock Egoeverything: A benchmark for human behavior inspired long context
  egocentric video understanding in ar environment, 2026.

\bibitem[Xu et~al.(2026)Xu, Yang, Jiang, Liu, Hou, Fan, Chen, Yan, and
  Jiang]{xu2026pro2assistcontinuousstepawareproactive}
Lilin Xu, Bufang Yang, Siyang Jiang, Kaiwei Liu, Kaiyuan Hou, Yuang Fan,
  Hongkai Chen, Zhenyu Yan, and Xiaofan Jiang.
\newblock Pro$^2$assist: Continuous step-aware proactive assistance with
  multi-modal egocentric perception for long-horizon procedural tasks, 2026.

\bibitem[Yang et~al.(2025)Yang, Tamkittikhun, Hamilton-Fletcher, Ramdhanie, Vu,
  Beheshti, Hudson, Vedanthan, Riewpaiboon, Mongkolwat, Feng, and
  Rizzo]{yang2025unav}
Anbang Yang, Nattachart Tamkittikhun, Giles Hamilton-Fletcher, Vinay Ramdhanie,
  Thu Vu, Mahya Beheshti, Todd Hudson, Rajesh Vedanthan, Wachara Riewpaiboon,
  Pattanasak Mongkolwat, Chen Feng, and John-Ross Rizzo.
\newblock Evaluating the efficacy of {UNav}: A computer vision-based navigation
  aid for persons with blindness or low vision.
\newblock \emph{Assistive Technology}, 37\penalty0 (6):\penalty0 459--473,
  2025.
\newblock PMID: 39137956.

\bibitem[Yao et~al.(2026)Yao, Gu, Deng, Dai, Fan, Zhang, Huang, Fan, and
  Zhang]{ICLR2026_18374cc2}
Jiali Yao, Xin Gu, Xinran Deng, Mengrui Dai, Bing Fan, Zhipeng Zhang, Yan
  Huang, Heng Fan, and Libo Zhang.
\newblock Omnistvg: Toward spatio-temporal omni-object video grounding.
\newblock In \emph{International Conference on Learning Representations}, pages
  14584--14605, 2026.

\bibitem[Zhang et~al.(2026{\natexlab{a}})Zhang, Huang, Dong, Fang, Chen, Zhu,
  Chen, Zhang, and Li]{zhang2026gstbenchvlmsdevelopglobal}
Qifeng Zhang, Kaixiang Huang, Heng Dong, Huang Fang, Junting Chen, Junjie Zhu,
  Yonghang Chen, Zhiyu Zhang, and Wei Li.
\newblock Gst-bench: Can vlms develop global spatial awareness from video?,
  2026{\natexlab{a}}.

\bibitem[Zhang et~al.(2026{\natexlab{b}})Zhang, Li, Zhu, Wang, Wu, Yu, Chu, Lu,
  and Jia]{zhang2026streamarenacontinuousinteractivelonghorizon}
Xichen Zhang, Guankai Li, Yinghao Zhu, Shijian Wang, Sitong Wu, Shaozuo Yu,
  Meng Chu, Yuan Lu, and Jiaya Jia.
\newblock Streamarena: Toward continuous, interactive, and long-horizon agentic
  streaming video understanding, 2026{\natexlab{b}}.

\bibitem[Zhao et~al.(2026)Zhao, Xie, Yuan, Hu, Guo, Yang, and
  He]{zhao2026pasbenchvideostreamingvideobenchmark}
Yusong Zhao, Yuejin Xie, Youliang Yuan, Junjie Hu, Jitian Guo, Yujiu Yang, and
  Pinjia He.
\newblock Pasbench-video: A streaming video benchmark for proactive safety
  warning, 2026.

\bibitem[Zhou et~al.(2025{\natexlab{a}})Zhou, Vilesov, He, Wan, Zhang,
  Nagachandra, Chang, Chen, Wang, and Kadambi]{Zhou_2025_VLM4D}
Shijie Zhou, Alexander Vilesov, Xuehai He, Ziyu Wan, Shuwang Zhang, Aditya
  Nagachandra, Di Chang, Dongdong Chen, Xin~Eric Wang, and Achuta Kadambi.
\newblock Vlm4d: Towards spatiotemporal awareness in vision language models.
\newblock In \emph{Proceedings of the IEEE/CVF International Conference on
  Computer Vision (ICCV)}, pages 8600--8612, 2025{\natexlab{a}}.

\bibitem[Zhou et~al.(2025{\natexlab{b}})Zhou, Tang, Xiao, Lin, Liu, Guo, Fei,
  Xia, and Gou]{Zhou_2025_WWW}
Yuchen Zhou, Jiayu Tang, Xiaoyan Xiao, Yueyao Lin, Linkai Liu, Zipeng Guo, Hao
  Fei, Xiaobo Xia, and Chao Gou.
\newblock Where, what, why: Towards explainable driver attention prediction.
\newblock In \emph{Proceedings of the IEEE/CVF International Conference on
  Computer Vision (ICCV)}, pages 2675--2685, 2025{\natexlab{b}}.

\end{thebibliography}
}

\clearpage
\onecolumn
\setcounter{page}{1}
\renewcommand{\thepage}{S\arabic{page}}
\begingroup
\centering\Large
\textbf{\thetitle}\\[0.5em]
Supplementary Material\par\vspace{1em}
\endgroup
\appendix
\raggedbottom
\makeatletter
\renewcommand{\theHtable}{supp.\arabic{LT@tables}}
\patchcmd{\LT@output}{\vss}{\vfil}{}{}
\patchcmd{\LT@output}{\vss}{\vfil}{}{}
\makeatother
\setcounter{topnumber}{4}
\setcounter{bottomnumber}{2}
\setcounter{totalnumber}{6}
\renewcommand{\topfraction}{0.95}
\renewcommand{\bottomfraction}{0.90}
\renewcommand{\textfraction}{0.05}
\renewcommand{\floatpagefraction}{0.75}
\setlength{\floatsep}{12pt plus 2pt minus 2pt}
\setlength{\textfloatsep}{14pt plus 2pt minus 2pt}
\setlength{\intextsep}{12pt plus 2pt minus 2pt}

\section{Benchmark Details}
\label{app:benchmark_details}

\subsection{Data Composition and Temporal Sampling}
Each \rev{evaluation clip} serves as the unit of visual observation and question construction. The resulting question set spans five tasks, covering wearer motion, external-object motion, passage-relation judgments, and autonomous discovery of relevant entities. Table~\ref{tab:task_distribution} summarizes the question distribution. The first four rows form the MCQ evaluation; the final row is one FFG task evaluated with Discovery, Exact and Labels. Its empty and nonempty references share the same inclusion criteria and scoring rule.

\rev{The benchmark contains 2,201 distinct questions. To test the effect of visual history, we asked the same 293 questions from 54 \rev{evaluation clips} with and without the preceding clip (Appendix~\ref{app:history_full}).}

\par\addvspace{8pt}\begingroup

\color{RevisionBlue}
\centering

\begingroup
\EgoTableSetup{9}{2.4}
\begin{longtable}{@{\hspace{\tabcolsep}}>{\raggedright\arraybackslash}p{\dimexpr0.761489473\linewidth-2.284468418\tabcolsep\relax}>{\centering\arraybackslash}p{\dimexpr0.238510527\linewidth-0.715531582\tabcolsep\relax}@{}}
\caption{\rev{Distribution of the 2,201 questions across the five tasks. All 574 FFG questions form one task, with 211 empty and 363 nonempty reference sets.}}\label{tab:task_distribution}\\

\arrayrulecolor{black!70}
\toprule
{\fontsize{8.5}{10.03}\selectfont\bfseries \textbf{Task}} & {\fontsize{8.5}{10.03}\selectfont\bfseries \textbf{Questions}} \\
\midrule
\endfirsthead
\caption[]{(continued)}\\

\arrayrulecolor{black!70}
\toprule
{\fontsize{8.5}{10.03}\selectfont\bfseries \textbf{Task}} & {\fontsize{8.5}{10.03}\selectfont\bfseries \textbf{Questions}} \\
\midrule
\endhead
\midrule
\endfoot
\endlastfoot

\rev{Wearer translation} & \rev{267} \\
\rev{Turning} & \rev{260} \\
\rev{External-object motion} & \rev{425} \\
\rev{Specified-object passage relation} & \rev{675} \\
\rev{Relevant-entity discovery (FFG)} & \rev{574} \\
\rev{\textbf{Total}} & \rev{\textbf{2,201}} \\
\bottomrule
\end{longtable}
\endgroup

\endgroup\par\addvspace{8pt}

\rev{Sampling targeted common walking environments. Object and interaction counts describe the evaluated material and may overlap across categories. Details appear in Tables~\ref{tab:objects} and~\ref{tab:interactions}.}

\subsection{Additional Relevance and Attention Definitions}
Passage relevance is determined by the spatiotemporal relationship between the wearer and an external entity. Approaching along separate paths with sufficient clearance, or crossing the same path at clearly different times, does not by itself establish a conflict. Conversely, a stationary structure may be relevant if it restricts the wearer's forward passage. Entities are defined according to the space they physically occupy: a rider and the vehicle being ridden are treated as a single unit, whereas independent objects are assessed separately.

To assess whether models understand how passage constraints evolve over time, we compare the attention required by the same object at two time points within a clip. Required attention refers to the need for observation, preparation, or action adjustment arising from the passage relation, rather than visual salience or subjective anxiety. An object is classified as irrelevant only when it clearly fails to meet the inclusion criteria at both time points. Cases that cannot be judged reliably retain an insufficient-evidence label. An object leaving the field of view is not, by itself, treated as evidence that the required attention has decreased.

\subsection{Labels and Evidence Boundaries}
External-object motion has four labels: stationary, moving, motion state changed, and uncertain. A state change requires a confirmed start or stop, not a direction change alone. Passage-relation answers distinguish higher or lower attention requirements, irrelevance, and insufficient evidence. FFG returns qualifying entities and their change labels, or an empty set when none qualifies. A change label cannot justify including an irrelevant object.

Qualification requires concrete visual evidence of physical conflict, insufficient passage space, or an avoidance adjustment that must begin within two seconds. Visibility, approach, motion, and salience are not sufficient. Nor is an inability to rule out hypothetical danger. Riders and their vehicles form one occupied unit. Traffic rules, signal colors, paths separated by adequate space, and crossings at clearly different times do not substitute for visual evidence of a conflict. In this benchmark, no relevant object has an unchanged attention requirement across the queried time pair; insufficient evidence means the visual evidence cannot resolve the comparison. Both conditions permit all allowed current frames, rather than only endpoint images. History adds only the immediately preceding clip, never earlier answers.

\FloatBarrier
\section{\rev{Model and Evaluation Protocols}}
\label{app:evaluation_protocols}
{\color{RevisionBlue}
\FloatBarrier
\subsection{Model Configurations}
\label{app:model_configurations}
\rev{Table~\ref{tab:model_configurations} summarizes all 14 visual configurations: eight with chronological frames and six with native video. Each model receives only the observations permitted for that question and a task-specific response format.}

\par\addvspace{8pt}\begingroup

\color{RevisionBlue}
\centering

\begingroup
\footnotesize
\setlength{\tabcolsep}{3pt}
\renewcommand{\arraystretch}{1.38}
\setlength{\extrarowheight}{2pt}
\definecolor{ConfigHeader}{HTML}{DDE6ED}
\definecolor{ConfigBlue}{HTML}{EDF4FA}
\definecolor{ConfigLavender}{HTML}{F2EFF8}
\definecolor{ConfigSand}{HTML}{FBF4E8}
\definecolor{ConfigMint}{HTML}{ECF5F0}
\begingroup
\EgoTableSetup{8}{1.8}
\begin{longtable}{@{\hspace{\tabcolsep}}>{\raggedright\arraybackslash}p{\dimexpr0.196929682\linewidth-2.953945234\tabcolsep\relax}>{\raggedright\arraybackslash}p{\dimexpr0.254002132\linewidth-3.810031985\tabcolsep\relax}>{\centering\arraybackslash}p{\dimexpr0.127987566\linewidth-1.919813486\tabcolsep\relax}>{\centering\arraybackslash}p{\dimexpr0.072910597\linewidth-1.093658958\tabcolsep\relax}>{\centering\arraybackslash}p{\dimexpr0.070903842\linewidth-1.063557628\tabcolsep\relax}>{\centering\arraybackslash}p{\dimexpr0.074941530\linewidth-1.124122950\tabcolsep\relax}>{\centering\arraybackslash}p{\dimexpr0.100473259\linewidth-1.507098884\tabcolsep\relax}>{\centering\arraybackslash}p{\dimexpr0.101851392\linewidth-1.527770874\tabcolsep\relax}@{}}
\caption{\textbf{Model configurations for evaluation.} Settings are based on evaluation records and official model or provider documentation. Unset parameters use defaults where applicable. \textbf{Temp.}: temperature.}\label{tab:model_configurations}\\

\arrayrulecolor{black!70}
\toprule
{\fontsize{7.5}{8.85}\selectfont\bfseries \shortstack[c]{\strut\textbf{Model}\\\strut}} & {\fontsize{7.5}{8.85}\selectfont\bfseries \shortstack[c]{\strut\textbf{Model}\\\strut\textbf{identifier}}} & {\fontsize{7.5}{8.85}\selectfont\bfseries \shortstack[c]{\strut\textbf{Max output}\\\strut\textbf{tokens}}} & {\fontsize{7.5}{8.85}\selectfont\bfseries \shortstack[c]{\strut\textbf{Temp.}\\\strut}} & {\fontsize{7.5}{8.85}\selectfont\bfseries \shortstack[c]{\strut\textbf{Top-P}\\\strut}} & {\fontsize{7.5}{8.85}\selectfont\bfseries \shortstack[c]{\strut\textbf{Top-K}\\\strut}} & {\fontsize{7.5}{8.85}\selectfont\bfseries \shortstack[c]{\strut\textbf{Visual}\\\strut\textbf{input}}} & {\fontsize{7.5}{8.85}\selectfont\bfseries \shortstack[c]{\strut\textbf{Thinking}\\\strut\textbf{setting}}} \\
\midrule
\endfirsthead
\caption[]{(continued)}\\

\arrayrulecolor{black!70}
\toprule
{\fontsize{7.5}{8.85}\selectfont\bfseries \shortstack[c]{\strut\textbf{Model}\\\strut}} & {\fontsize{7.5}{8.85}\selectfont\bfseries \shortstack[c]{\strut\textbf{Model}\\\strut\textbf{identifier}}} & {\fontsize{7.5}{8.85}\selectfont\bfseries \shortstack[c]{\strut\textbf{Max output}\\\strut\textbf{tokens}}} & {\fontsize{7.5}{8.85}\selectfont\bfseries \shortstack[c]{\strut\textbf{Temp.}\\\strut}} & {\fontsize{7.5}{8.85}\selectfont\bfseries \shortstack[c]{\strut\textbf{Top-P}\\\strut}} & {\fontsize{7.5}{8.85}\selectfont\bfseries \shortstack[c]{\strut\textbf{Top-K}\\\strut}} & {\fontsize{7.5}{8.85}\selectfont\bfseries \shortstack[c]{\strut\textbf{Visual}\\\strut\textbf{input}}} & {\fontsize{7.5}{8.85}\selectfont\bfseries \shortstack[c]{\strut\textbf{Thinking}\\\strut\textbf{setting}}} \\
\midrule
\endhead
\midrule
\endfoot
\endlastfoot

\rowcolor{EgoTableNeutral}\multicolumn{8}{l}{\textbf{Chronological frames: GPT}} \\
GPT-6 Astra$^{*}$ & \texttt{gpt-6-astra} & -- & -- & -- & -- & 2 fps$^{a}$ & Medium \\
GPT-5.6 Sol & \texttt{gpt-5.6-sol} & -- & -- & -- & -- & 2 fps$^{a}$ & Medium \\
\midrule
\rowcolor{EgoTableNeutral}\multicolumn{8}{l}{\textbf{Chronological frames: Gemini}} \\
Gemini 3.8 Flash & \texttt{gemini-3.8-flash} & 16,384 & 1.0 & 0.95$^{c}$ & 64$^{c}$ & 2 fps$^{a}$ & Medium \\
\shortstack[l]{Gemini 3.1 Pro\\(Preview)} & \shortstack[l]{\texttt{gemini-3.1-}\\\texttt{pro-preview}} & 16,384 & 1.0 & 0.95$^{c}$ & 64$^{c}$ & 2 fps$^{a}$ & Medium \\
\midrule
\rowcolor{EgoTableNeutral}\multicolumn{8}{l}{\textbf{Chronological frames: Llama}} \\
Llama 4 Maverick & \shortstack[l]{\texttt{meta-llama/}\\\texttt{llama-4-maverick}} & 4,096 & 1.0$^{d}$ & 1.0$^{d}$ & 0$^{d}$ & 2 fps$^{a}$ & -- \\
Llama 4 Scout & \shortstack[l]{\texttt{meta-llama/}\\\texttt{llama-4-scout}} & 4,096 & 1.0$^{d}$ & 1.0$^{d}$ & 0$^{d}$ & 2 fps$^{a}$ & -- \\
\midrule
\rowcolor{EgoTableNeutral}\multicolumn{8}{l}{\textbf{Chronological frames: Qwen}} \\
\shortstack[l]{Qwen3.8 Max\\(0902)} & \shortstack[l]{\texttt{qwen/}\\\texttt{qwen3.8-max-0902}} & 16,384 & 0.6$^{d}$ & 0.95$^{d}$ & 20$^{d}$ & 2 fps$^{a}$ & Medium \\
Qwen3.8 Flash & \shortstack[l]{\texttt{qwen/}\\\texttt{qwen3.8-flash}} & 16,384 & 0.6$^{d}$ & 0.95$^{d}$ & 20$^{d}$ & 2 fps$^{a}$ & On$^{b}$ \\
\midrule
\rowcolor{EgoTableNeutral}\multicolumn{8}{l}{\textbf{Native video}} \\
\shortstack[l]{Qwen3.8 Max\\(0902)} & \shortstack[l]{\texttt{qwen/}\\\texttt{qwen3.8-max-0902}} & 16,384 & 0 & 0.95 & 20 & Video & Minimal \\
GLM-5.3 Flash & \shortstack[l]{\texttt{z-ai/}\\\texttt{glm-5.3-flash}} & 16,384 & 0 & 0.95 & -- & Video & Low \\
MiniMax M3 & \shortstack[l]{\texttt{minimax/}\\\texttt{minimax-m3}} & 16,384 & 0 & 0.95 & -- & Video & Disabled \\
Gemma 4 31B & \shortstack[l]{\texttt{google/}\\\texttt{gemma-4-31b-it}} & 16,384 & 0 & 0.95 & 64 & Video & Disabled \\
\shortstack[l]{Qwen3.5-397B\\(A17B)} & \shortstack[l]{\texttt{qwen/}\\\texttt{qwen3.5-397b-a17b}} & 16,384 & 0 & 0.95 & 20 & Video & Disabled \\
Qwen3.8 Flash & \shortstack[l]{\texttt{qwen/}\\\texttt{qwen3.8-flash}} & 16,384 & 0 & 0.95 & 20 & Video & Disabled \\
\bottomrule
\end{longtable}
\endgroup
\endgroup
\par\smallskip
\begin{minipage}{\linewidth}\footnotesize
In Table~\ref{tab:model_configurations}, \texttt{--} denotes an unset or inapplicable parameter. $^{*}$GPT-6 does not support temperature or top-p adjustment. GPT sampling values and output caps were not recorded. $^{a}$Nominal 2 fps; clearer replacement frames retain their actual timestamps. $^{b}$Qwen3.8 Flash uses enabled reasoning without a specified effort level. $^{c}$Defaults from saved Google model metadata. $^{d}$Provider defaults documented on September 22, 2026; effective runtime values were not returned. Top-K=0 disables top-k sampling.

GPT models use medium reasoning effort. Gemini models use high media resolution and do not return reasoning traces. The frame-based Llama and Qwen configurations use their respective provider defaults for temperature, top-p, and top-k. Native-video rows use the recorded request parameters; Video denotes direct clip input, not 2 fps frame sampling.

\FloatBarrier
\end{minipage}

\endgroup\par\addvspace{8pt}

\subsection{Scoring, Entity Matching and Uncertainty}
\label{app:scoring_methods}
One turning question has no answer option matching its corrected reference and is excluded for all respondents. All 14 visual configurations and both human participants were evaluated on the same 2,201 questions. Each question contributes once to its task. The five tasks are wearer translation, viewpoint and path turning, external-object motion, passage relations for specified objects, and relevant-entity discovery (FFG). All 574 free responses form a single FFG task with the same scoring rules.

\paragraph{Visual inputs and response coverage.} Eight configurations use chronological frames. Six accept the clip as native video with temperature 0, top-$p$ 0.95, and a 16,384-token output cap. Native-video Qwen3.8 Max uses minimal reasoning, GLM-5.3 Flash uses low reasoning, and reasoning is disabled in the other four. FP8 serving routes included Parasail for frame-based Maverick and DeepInfra for native-video Qwen3.5-397B-A17B. The two groups answer the same questions, but differ in both visual input and inference settings.

Of 30,814 model--question combinations, 30,796 produced a scorable response. The other 18 count as zero on question-level metrics and contribute no matched entity to Labels. After a connection failure or an invalid response format, we retained the first response meeting the scoring requirements, regardless of correctness. Where a response unambiguously named an answer option, we mapped it to that option. For FFG format corrections, the program supplied the pair identifier and two times fixed by the question; the model still had to name the entities and assign their change labels. Models did not see reference answers or candidate entities.

\paragraph{How are 95\% confidence intervals computed?} We estimate sampling uncertainty by grouped resampling of the 284 evaluation clips. Sampling is stratified by collection and keeps clips from the same recording and all their questions together. Within each collection, groups are drawn with replacement at the observed group count. We use the same 10,000 draws across models and tasks; in each draw we recompute the score or the paired difference. The approximate pointwise 95\% confidence interval is the observed score $\pm 1.96$ times the standard deviation of these estimates, restricted to the possible score range. Four further sets of 10,000 draws with different seeds check numerical stability. For example, Astra's Discovery score on the 573 questions with an identifiable score is $55.7\%\pm6.7$ percentage points. For human scores, the sampled evaluation clips vary across draws but the two participants stay the same.

\paragraph{How are free descriptions matched?} The first scorer compares each description with the visible entities without seeing the model name. Appearance, position and trajectory help distinguish two people or objects. A second scorer reviews ambiguous cases with the first; both revisit the footage and record their decision. If a description clearly denotes a group, each identifiable member is scored against its own reference label using the predicted change for the group. A description of one person whose identity is ambiguous cannot count as two matches. For a detailed identity review covering 717 distinct descriptions on 63 questions, both the model name and its predicted change label were hidden during matching. The scoring materials will include the available correspondences or match counts and the scripts used to compute the scores.

\paragraph{What if an object still cannot be identified?} Table~\ref{tab:nohistory} reports a point estimate only from cases whose score is fixed for that metric (denominators in Table~\ref{tab:ffg_determined}). Table~\ref{tab:full_bounds} then includes \emph{all} 574 FFG questions and asks how far the score could change under matches consistent with the footage. For Astra, only one Discovery score is unresolved. There are three predictions and three reference entities: two matches are certain, and the third is ambiguous. If the third prediction is extra, Eq.~\ref{eq:object_score} gives $(2-1)/3=1/3$; if it matches, the score is 1. Including that question puts Astra's Discovery at 55.69--55.81\%. These are bounds on how the same answers can be scored, not a sampling confidence interval. For Discovery and Exact, we average the lowest and highest feasible question scores separately. For Labels, we bound the pooled ratio of correct labels to matched entities, using the same feasible matches for its numerator and denominator.

\FloatBarrier
\subsection{Human and Control Protocols}
\label{app:human_protocol}
After a separate practice session, both participants independently answered all 2,201 questions. Like the models, they received the same FFG questions and observation windows without an object list or reference answers. Each person wrote their own description of every relevant object and its change label, or returned an empty set. Descriptions were matched to reference entities only after the answers had been submitted. The two-person mean weights the participants equally.

The question-only condition supplies GPT-5.6 Sol with the MCQ text and answer options without visual input. The always-empty baseline returns no entity on every FFG question and is evaluated with the same scoring rules.

}

\FloatBarrier
\section{\rev{Complete Evaluation Results}}
\label{app:expanded_results}
\label{app:coverage}
\FloatBarrier
\subsection{Performance without Visual History}
\label{app:complete_nohistory}
\rev{Tables~\ref{tab:historical_model} and~\ref{tab:nohistory_coverage} report the same full benchmark as Table~\ref{tab:nohistory}: 2,201 questions across wearer translation, turning, object motion, passage relations, and relevant-entity discovery. The four multiple-choice tasks contain 267, 260, 425, and 675 questions; all 574 free-response questions use one FFG task and one set of scoring definitions. Each formal question contributes once. Unavailable responses count as zero on question-level metrics.}

\par\addvspace{8pt}\begingroup

\noindent\begin{minipage}{\linewidth}\normalfont\normalsize\raggedright
\textbf{Configuration-level scores.} Table~\ref{tab:historical_model} lists the complete no-history scores for every visual configuration. It keeps the four MCQ tasks separate and reports all three FFG metrics, so that an overall score is not mistaken for uniform performance across tasks. The frame and native-video groups differ in both input format and inference settings.
\end{minipage}\par\smallskip

\color{RevisionBlue}
\centering\color{RevisionBlue}

\begingroup
\EgoTableSetup{8.5}{1.8}
\begin{longtable}{@{\hspace{\tabcolsep}}>{\raggedright\arraybackslash}p{\dimexpr0.201914871\linewidth-3.432552802\tabcolsep\relax}>{\columncolor{black!5}\centering\arraybackslash}p{\dimexpr0.086248776\linewidth-1.466229186\tabcolsep\relax}>{\centering\arraybackslash}p{\dimexpr0.107004598\linewidth-1.819078167\tabcolsep\relax}>{\centering\arraybackslash}p{\dimexpr0.107079438\linewidth-1.820350443\tabcolsep\relax}>{\centering\arraybackslash}p{\dimexpr0.097325200\linewidth-1.654528401\tabcolsep\relax}>{\centering\arraybackslash}p{\dimexpr0.108426571\linewidth-1.843251707\tabcolsep\relax}>{\centering\arraybackslash}p{\dimexpr0.079313537\linewidth-1.348330131\tabcolsep\relax}>{\centering\arraybackslash}p{\dimexpr0.123644181\linewidth-2.101951081\tabcolsep\relax}>{\centering\arraybackslash}p{\dimexpr0.089042828\linewidth-1.513728082\tabcolsep\relax}@{}}
\caption{\rev{Visual-model scores on the full 2,201-question benchmark (\%). Values reproduce Table~\ref{tab:nohistory}; FFG uses the same 574 questions. Point-estimate denominators and identity bounds are in Tables~\ref{tab:ffg_determined} and~\ref{tab:full_bounds}.}}\label{tab:historical_model}\\

\arrayrulecolor{black!70}
\toprule
{\fontsize{8}{9.44}\selectfont\bfseries Model} & {\fontsize{8}{9.44}\selectfont\bfseries \shortstack{MCQ\\overall}} & {\fontsize{8}{9.44}\selectfont\bfseries \shortstack{Wearer\\translation}} & {\fontsize{8}{9.44}\selectfont\bfseries Turning} & {\fontsize{8}{9.44}\selectfont\bfseries \shortstack{Object\\motion}} & {\fontsize{8}{9.44}\selectfont\bfseries \shortstack{Passage\\relation}} & {\fontsize{8}{9.44}\selectfont\bfseries Exact} & {\fontsize{8}{9.44}\selectfont\bfseries Discovery} & {\fontsize{8}{9.44}\selectfont\bfseries Labels} \\
\midrule
\endfirsthead
\caption[]{(continued)}\\

\arrayrulecolor{black!70}
\toprule
{\fontsize{8}{9.44}\selectfont\bfseries Model} & {\fontsize{8}{9.44}\selectfont\bfseries \shortstack{MCQ\\overall}} & {\fontsize{8}{9.44}\selectfont\bfseries \shortstack{Wearer\\translation}} & {\fontsize{8}{9.44}\selectfont\bfseries Turning} & {\fontsize{8}{9.44}\selectfont\bfseries \shortstack{Object\\motion}} & {\fontsize{8}{9.44}\selectfont\bfseries \shortstack{Passage\\relation}} & {\fontsize{8}{9.44}\selectfont\bfseries Exact} & {\fontsize{8}{9.44}\selectfont\bfseries Discovery} & {\fontsize{8}{9.44}\selectfont\bfseries Labels} \\
\midrule
\endhead
\midrule
\endfoot
\endlastfoot

\rowcolor{EgoTableFrames}\multicolumn{9}{l}{\textbf{Chronological frames}} \\
GPT-6 Astra & \textbf{65.52} & \textbf{86.14} & 43.85 & \textbf{91.29} & \textbf{49.48} & \textbf{48.26} & \textbf{55.73} & \textbf{82.97} \\
GPT-5.6 Sol & 57.47 & 61.05 & 33.85 & 86.59 & 46.81 & 35.54 & 44.56 & 74.46 \\
Gemini 3.8 Flash & \underline{60.97} & 66.67 & \underline{45.77} & 87.29 & \underline{48.00} & \underline{43.80} & \underline{52.11} & \underline{76.09} \\
Gemini 3.1 Pro & 57.47 & 69.66 & 28.46 & 84.24 & 46.96 & 34.21 & 45.40 & 67.27 \\
Llama 4 Maverick & 47.20 & 39.33 & 38.08 & 80.00 & 33.19 & 14.44 & 29.46 & 38.46 \\
Llama 4 Scout & 44.38 & 36.33 & 35.00 & 73.88 & 32.59 & 28.70 & 39.91 & 49.59 \\
Qwen3.8 Max & 58.39 & 74.91 & 45.00 & \underline{87.53} & 38.67 & 33.45 & 43.88 & 71.48 \\
Qwen3.8 Flash & 54.46 & 64.04 & 30.77 & 85.41 & 40.30 & 29.49 & 40.17 & 68.97 \\
\midrule
\rowcolor{EgoTableVideo}\multicolumn{9}{l}{\textbf{Native video}} \\
Qwen3.8 Max & 58.57 & \underline{79.03} & 40.38 & 84.00 & 41.48 & 28.97 & 40.08 & 59.16 \\
GLM-5.3 Flash & 55.13 & 74.91 & 28.08 & 81.65 & 41.04 & 29.82 & 39.65 & 48.55 \\
MiniMax M3 & 52.49 & 66.29 & 34.23 & 81.41 & 35.85 & 26.18 & 31.68 & 50.00 \\
Gemma 4 31B & 54.76 & 57.30 & \textbf{55.00} & 81.18 & 37.04 & 28.80 & 37.00 & 54.74 \\
Qwen3.5-397B & 55.87 & 77.15 & 44.62 & 79.06 & 37.19 & 24.30 & 29.20 & 47.54 \\
Qwen3.8 Flash & 56.36 & 77.53 & 36.92 & 83.29 & 38.52 & 29.07 & 34.82 & 52.63 \\
\bottomrule
\end{longtable}
\endgroup

\endgroup\par\addvspace{8pt}
\par\addvspace{8pt}\begingroup

\noindent\begin{minipage}{\linewidth}\normalfont\normalsize\raggedright
\textbf{Response coverage and error counts.} Table~\ref{tab:nohistory_coverage} separates unavailable responses from errors in scorable answers. Panel (a) gives correct/planned MCQs and response coverage. Panel (b), continued below, separates entity matching and labeling errors.
\end{minipage}\par\smallskip

\color{RevisionBlue}
\centering\color{RevisionBlue}

\textbf{(a) Question and response counts}\par\smallskip
\begingroup
\EgoTableSetup{8.5}{1.8}
\begin{longtable}{@{\hspace{\tabcolsep}}>{\raggedright\arraybackslash}p{\dimexpr0.191337132\linewidth-3.252731248\tabcolsep\relax}>{\centering\arraybackslash}p{\dimexpr0.101398935\linewidth-1.723781902\tabcolsep\relax}>{\centering\arraybackslash}p{\dimexpr0.101469854\linewidth-1.724987526\tabcolsep\relax}>{\centering\arraybackslash}p{\dimexpr0.092226613\linewidth-1.567852423\tabcolsep\relax}>{\centering\arraybackslash}p{\dimexpr0.102746415\linewidth-1.746689059\tabcolsep\relax}>{\columncolor{black!5}\centering\arraybackslash}p{\dimexpr0.113246807\linewidth-1.925195721\tabcolsep\relax}>{\centering\arraybackslash}p{\dimexpr0.088129302\linewidth-1.498198129\tabcolsep\relax}>{\centering\arraybackslash}p{\dimexpr0.113246807\linewidth-1.925195721\tabcolsep\relax}>{\centering\arraybackslash}p{\dimexpr0.096198134\linewidth-1.635368271\tabcolsep\relax}@{}}
\caption{Response counts for all 2,201 questions. Panel (a): correct/planned MCQs, valid/planned responses and unavailable responses. Entity counts continue in panel (b).}\label{tab:nohistory_coverage}\\

\arrayrulecolor{black!70}
\toprule
{\fontsize{8}{9.44}\selectfont\bfseries Model} & {\fontsize{8}{9.44}\selectfont\bfseries \shortstack{Wearer\\translation}} & {\fontsize{8}{9.44}\selectfont\bfseries Turning} & {\fontsize{8}{9.44}\selectfont\bfseries \shortstack{Object\\motion}} & {\fontsize{8}{9.44}\selectfont\bfseries \shortstack{Passage\\relation}} & {\fontsize{8}{9.44}\selectfont\bfseries \shortstack{MCQ\\correct}} & {\fontsize{8}{9.44}\selectfont\bfseries \shortstack{FFG\\valid}} & {\fontsize{8}{9.44}\selectfont\bfseries \shortstack{All\\valid}} & {\fontsize{8}{9.44}\selectfont\bfseries Missing} \\
\midrule
\endfirsthead
\caption[]{(continued)}\\

\arrayrulecolor{black!70}
\toprule
{\fontsize{8}{9.44}\selectfont\bfseries Model} & {\fontsize{8}{9.44}\selectfont\bfseries \shortstack{Wearer\\translation}} & {\fontsize{8}{9.44}\selectfont\bfseries Turning} & {\fontsize{8}{9.44}\selectfont\bfseries \shortstack{Object\\motion}} & {\fontsize{8}{9.44}\selectfont\bfseries \shortstack{Passage\\relation}} & {\fontsize{8}{9.44}\selectfont\bfseries \shortstack{MCQ\\correct}} & {\fontsize{8}{9.44}\selectfont\bfseries \shortstack{FFG\\valid}} & {\fontsize{8}{9.44}\selectfont\bfseries \shortstack{All\\valid}} & {\fontsize{8}{9.44}\selectfont\bfseries Missing} \\
\midrule
\endhead
\midrule
\endfoot
\endlastfoot

\rowcolor{EgoTableFrames}\multicolumn{9}{l}{\textbf{Chronological frames}} \\
GPT-6 Astra & 230/267 & 114/260 & 388/425 & 334/675 & 1066/1627 & 574/574 & 2201/2201 & 0 \\
GPT-5.6 Sol & 163/267 & 88/260 & 368/425 & 316/675 & 935/1627 & 574/574 & 2201/2201 & 0 \\
Gemini 3.8 Flash & 178/267 & 119/260 & 371/425 & 324/675 & 992/1627 & 574/574 & 2201/2201 & 0 \\
Gemini 3.1 Pro & 186/267 & 74/260 & 358/425 & 317/675 & 935/1627 & 574/574 & 2201/2201 & 0 \\
Llama 4 Maverick & 105/267 & 99/260 & 340/425 & 224/675 & 768/1627 & 573/574 & 2200/2201 & 1 \\
Llama 4 Scout & 97/267 & 91/260 & 314/425 & 220/675 & 722/1627 & 574/574 & 2201/2201 & 0 \\
Qwen3.8 Max & 200/267 & 117/260 & 372/425 & 261/675 & 950/1627 & 572/574 & 2193/2201 & 8 \\
Qwen3.8 Flash & 171/267 & 80/260 & 363/425 & 272/675 & 886/1627 & 570/574 & 2193/2201 & 8 \\
\midrule
\rowcolor{EgoTableVideo}\multicolumn{9}{l}{\textbf{Native video}} \\*
Qwen3.8 Max & 211/267 & 105/260 & 357/425 & 280/675 & 953/1627 & 574/574 & 2201/2201 & 0 \\
GLM-5.3 Flash & 200/267 & 73/260 & 347/425 & 277/675 & 897/1627 & 574/574 & 2201/2201 & 0 \\
MiniMax M3 & 177/267 & 89/260 & 346/425 & 242/675 & 854/1627 & 574/574 & 2201/2201 & 0 \\
Gemma 4 31B & 153/267 & 143/260 & 345/425 & 250/675 & 891/1627 & 573/574 & 2200/2201 & 1 \\
Qwen3.5-397B & 206/267 & 116/260 & 336/425 & 251/675 & 909/1627 & 574/574 & 2201/2201 & 0 \\
Qwen3.8 Flash & 207/267 & 96/260 & 354/425 & 260/675 & 917/1627 & 574/574 & 2201/2201 & 0 \\
\bottomrule
\end{longtable}
\endgroup

\endgroup\par\addvspace{8pt}

\par\addvspace{8pt}\begingroup

\noindent\begin{minipage}{\linewidth}\normalfont\normalsize\raggedright
\textbf{Entity-level coverage.} Panel (b) of Table~\ref{tab:nohistory_coverage} counts outcomes for the same reference entities. Matched entities can receive a wrong change label; omitted reference entities and additional reports remain separate errors. Conditional label accuracy therefore does not measure complete entity coverage.
\end{minipage}\par\smallskip
\centering\color{RevisionBlue}

\textbf{(b) FFG entity counts}\par\smallskip
\begingroup
\EgoTableSetup{9}{2.4}
\addtocounter{table}{-1}
\begin{longtable}{@{\hspace{\tabcolsep}}>{\raggedright\arraybackslash}p{\dimexpr0.264955536\linewidth-2.914510901\tabcolsep\relax}>{\centering\arraybackslash}p{\dimexpr0.135314897\linewidth-1.488463866\tabcolsep\relax}>{\centering\arraybackslash}p{\dimexpr0.135314897\linewidth-1.488463866\tabcolsep\relax}>{\centering\arraybackslash}p{\dimexpr0.135314897\linewidth-1.488463866\tabcolsep\relax}>{\centering\arraybackslash}p{\dimexpr0.141524473\linewidth-1.556769206\tabcolsep\relax}>{\centering\arraybackslash}p{\dimexpr0.187575300\linewidth-2.063328296\tabcolsep\relax}@{}}
\caption{Response counts (continued): FFG outcomes for 431 reference entities across 574 questions, including 211 empty and 363 nonempty references. Ranges retain unresolved identity matches; Labels is conditional accuracy (\%).}\\

\arrayrulecolor{black!70}
\toprule
{\fontsize{8.5}{10.03}\selectfont\bfseries Model} & {\fontsize{8.5}{10.03}\selectfont\bfseries TP} & {\fontsize{8.5}{10.03}\selectfont\bfseries FP} & {\fontsize{8.5}{10.03}\selectfont\bfseries FN} & {\fontsize{8.5}{10.03}\selectfont\bfseries \shortstack{Correct\\labels}} & {\fontsize{8.5}{10.03}\selectfont\bfseries Labels (\%)} \\
\midrule
\endfirsthead
\caption[]{(continued)}\\

\arrayrulecolor{black!70}
\toprule
{\fontsize{8.5}{10.03}\selectfont\bfseries Model} & {\fontsize{8.5}{10.03}\selectfont\bfseries TP} & {\fontsize{8.5}{10.03}\selectfont\bfseries FP} & {\fontsize{8.5}{10.03}\selectfont\bfseries FN} & {\fontsize{8.5}{10.03}\selectfont\bfseries \shortstack{Correct\\labels}} & {\fontsize{8.5}{10.03}\selectfont\bfseries Labels (\%)} \\
\midrule
\endhead
\midrule
\endfoot
\endlastfoot

\rowcolor{EgoTableFrames}\multicolumn{6}{l}{\textbf{Chronological frames}} \\
GPT-6 Astra & 278--279 & 224--225 & 152--153 & 231 & 82.80--83.09 \\
GPT-5.6 Sol & 278--279 & 332--333 & 152--153 & 207--208 & 74.46--74.55 \\
Gemini 3.8 Flash & 184--190 & 132--138 & 241--247 & 140--143 & 74.47--76.47 \\
Gemini 3.1 Pro & 224--239 & 251--266 & 192--207 & 149--160 & 64.22--68.38 \\
Llama 4 Maverick & 130--157 & 305--332 & 274--301 & 50--63 & 33.56--44.06 \\
Llama 4 Scout & 123--143 & 198--218 & 288--308 & 61--67 & 44.53--51.94 \\
Qwen3.8 Max & 258--263 & 319--324 & 168--173 & 184--186 & 70.50--71.54 \\
Qwen3.8 Flash & 235--247 & 336--348 & 184--196 & 161--168 & 65.98--69.42 \\
\midrule
\rowcolor{EgoTableVideo}\multicolumn{6}{l}{\textbf{Native video}} \\*
Qwen3.8 Max & 193--201 & 256--264 & 230--238 & 113--116 & 57.07--59.18 \\
GLM-5.3 Flash & 139--149 & 185--195 & 282--292 & 67--73 & 46.53--50.34 \\
MiniMax M3 & 101--114 & 226--239 & 317--330 & 50--54 & 44.64--51.43 \\
Gemma 4 31B & 95--106 & 136--147 & 325--336 & 52--54 & 49.52--55.67 \\
Qwen3.5-397B & 61--66 & 253--258 & 365--370 & 29--33 & 46.03--50.77 \\
Qwen3.8 Flash & 76--88 & 132--144 & 343--355 & 40--45 & 47.62--55.56 \\
\bottomrule
\end{longtable}
\endgroup

\endgroup\par\addvspace{8pt}

{\color{RevisionBlue}
\par\addvspace{8pt}\begingroup

\noindent\begin{minipage}{\linewidth}\normalfont\normalsize\raggedright
\textbf{Sampling uncertainty in MCQ accuracy.} Table~\ref{tab:mcq_uncertainty} attaches the grouped-resampling intervals from Appendix~\ref{app:scoring_methods} to the MCQ point estimates. Overall uses the pooled question denominator, while task columns use their own denominators. These intervals describe sampling variability of the evaluated clips; a difference between point estimates alone is not a test of a paired difference.
\end{minipage}\par\smallskip

\color{RevisionBlue}

\color{RevisionBlue}
\centering\color{RevisionBlue}

\begingroup\footnotesize
\setlength{\tabcolsep}{3.2pt}\renewcommand{\arraystretch}{1.10}
\begingroup
\EgoTableSetup{9}{2.4}
\begingroup
\EgoTableSetup{9}{2.4}
\begin{longtable}{@{\hspace{\tabcolsep}}>{\raggedright\arraybackslash}p{\dimexpr0.281325291\linewidth-3.094578198\tabcolsep\relax}>{\columncolor{black!5}\centering\arraybackslash}p{\dimexpr0.136747412\linewidth-1.504221531\tabcolsep\relax}>{\centering\arraybackslash}p{\dimexpr0.148346712\linewidth-1.631813834\tabcolsep\relax}>{\centering\arraybackslash}p{\dimexpr0.148451524\linewidth-1.632966762\tabcolsep\relax}>{\centering\arraybackslash}p{\dimexpr0.134790904\linewidth-1.482699941\tabcolsep\relax}>{\centering\arraybackslash}p{\dimexpr0.150338158\linewidth-1.653719734\tabcolsep\relax}@{}}
\caption{\rev{MCQ accuracy with group-resampled 95\% sampling intervals (\%). Overall pools the four task denominators. Wearer translation, turning, object motion and passage relations contain 267, 260, 425 and 675 questions, respectively.}}\label{tab:mcq_uncertainty}\\

\arrayrulecolor{black!70}
\toprule
{\fontsize{8.5}{10.03}\selectfont\bfseries Configuration} & {\fontsize{8.5}{10.03}\selectfont\bfseries Overall} & {\fontsize{8.5}{10.03}\selectfont\bfseries \shortstack{Wearer\\translation}} & {\fontsize{8.5}{10.03}\selectfont\bfseries Turning} & {\fontsize{8.5}{10.03}\selectfont\bfseries \shortstack{Object\\motion}} & {\fontsize{8.5}{10.03}\selectfont\bfseries \shortstack{Passage\\relation}} \\
\midrule
\endfirsthead
\caption[]{(continued)}\\

\arrayrulecolor{black!70}
\toprule
{\fontsize{8.5}{10.03}\selectfont\bfseries Configuration} & {\fontsize{8.5}{10.03}\selectfont\bfseries Overall} & {\fontsize{8.5}{10.03}\selectfont\bfseries \shortstack{Wearer\\translation}} & {\fontsize{8.5}{10.03}\selectfont\bfseries Turning} & {\fontsize{8.5}{10.03}\selectfont\bfseries \shortstack{Object\\motion}} & {\fontsize{8.5}{10.03}\selectfont\bfseries \shortstack{Passage\\relation}} \\
\midrule
\endhead
\midrule
\endfoot
\endlastfoot

\rowcolor{EgoTableFrames}\multicolumn{6}{l}{\textbf{Chronological frames}} \\
GPT-6 Astra & 65.5$\pm$2.7 & 86.1$\pm$5.1 & 43.8$\pm$6.7 & 91.3$\pm$3.1 & 49.5$\pm$3.7 \\
GPT-5.6 Sol & 57.5$\pm$2.9 & 61.0$\pm$7.8 & 33.8$\pm$5.9 & 86.6$\pm$4.7 & 46.8$\pm$4.1 \\
Gemini 3.8 Flash & 61.0$\pm$3.1 & 66.7$\pm$8.4 & 45.8$\pm$6.5 & 87.3$\pm$4.2 & 48.0$\pm$3.9 \\
Gemini 3.1 Pro & 57.5$\pm$2.3 & 69.7$\pm$6.4 & 28.5$\pm$6.4 & 84.2$\pm$4.4 & 47.0$\pm$4.3 \\
Llama 4 Maverick & 47.2$\pm$2.7 & 39.3$\pm$6.5 & 38.1$\pm$7.0 & 80.0$\pm$5.1 & 33.2$\pm$3.7 \\
Llama 4 Scout & 44.4$\pm$3.1 & 36.3$\pm$6.2 & 35.0$\pm$6.6 & 73.9$\pm$5.8 & 32.6$\pm$4.2 \\
Qwen3.8 Max & 58.4$\pm$2.9 & 74.9$\pm$7.0 & 45.0$\pm$6.5 & 87.5$\pm$4.1 & 38.7$\pm$4.2 \\
Qwen3.8 Flash & 54.5$\pm$2.7 & 64.0$\pm$7.2 & 30.8$\pm$5.3 & 85.4$\pm$4.0 & 40.3$\pm$3.8 \\
\midrule
\rowcolor{EgoTableVideo}\multicolumn{6}{l}{\textbf{Native video}} \\
Qwen3.8 Max & 58.6$\pm$2.9 & 79.0$\pm$6.8 & 40.4$\pm$7.6 & 84.0$\pm$4.9 & 41.5$\pm$3.9 \\
GLM-5.3 Flash & 55.1$\pm$2.9 & 74.9$\pm$7.0 & 28.1$\pm$5.9 & 81.6$\pm$5.0 & 41.0$\pm$3.9 \\
MiniMax M3 & 52.5$\pm$2.5 & 66.3$\pm$7.0 & 34.2$\pm$6.1 & 81.4$\pm$5.2 & 35.9$\pm$4.1 \\
Gemma 4 31B & 54.8$\pm$3.2 & 57.3$\pm$8.2 & 55.0$\pm$6.8 & 81.2$\pm$5.5 & 37.0$\pm$4.0 \\
Qwen3.5-397B & 55.9$\pm$3.3 & 77.2$\pm$6.9 & 44.6$\pm$6.8 & 79.1$\pm$4.4 & 37.2$\pm$4.5 \\
Qwen3.8 Flash & 56.4$\pm$2.6 & 77.5$\pm$6.1 & 36.9$\pm$6.7 & 83.3$\pm$4.8 & 38.5$\pm$4.0 \\
\bottomrule
\end{longtable}
\endgroup
\endgroup\endgroup

\endgroup\par\addvspace{8pt}

\par\addvspace{8pt}\begingroup

\noindent\begin{minipage}{\linewidth}\normalfont\normalsize\raggedright
\textbf{Scores under unresolved entity identity.} Table~\ref{tab:full_bounds} includes every FFG question and shows the lowest and highest scores compatible with the remaining identity ambiguities. These ranges describe alternative feasible matches of the same responses, rather than sampling confidence intervals. Empty and Nonempty split Discovery by reference-set status, exposing how correct empty answers and successful discovery contribute to the combined score.
\end{minipage}\par\smallskip

\color{RevisionBlue}
\centering\color{RevisionBlue}

\begingroup\scriptsize
\setlength{\tabcolsep}{3.2pt}\renewcommand{\arraystretch}{1.10}
\begingroup
\EgoTableSetup{8.5}{2.4}
\begin{longtable}{@{\hspace{\tabcolsep}}>{\raggedright\arraybackslash}p{\dimexpr0.220224206\linewidth-2.862914682\tabcolsep\relax}>{\centering\arraybackslash}p{\dimexpr0.070247181\linewidth-0.913213352\tabcolsep\relax}>{\columncolor{black!5}\centering\arraybackslash}p{\dimexpr0.156046923\linewidth-2.028610005\tabcolsep\relax}>{\centering\arraybackslash}p{\dimexpr0.156046923\linewidth-2.028610005\tabcolsep\relax}>{\centering\arraybackslash}p{\dimexpr0.156046923\linewidth-2.028610005\tabcolsep\relax}>{\centering\arraybackslash}p{\dimexpr0.085340919\linewidth-1.109431949\tabcolsep\relax}>{\centering\arraybackslash}p{\dimexpr0.156046923\linewidth-2.028610005\tabcolsep\relax}@{}}
\caption{\rev{FFG scores across all 574 questions (\%). Ranges give the lowest and highest scores consistent with unresolved object identities. Valid counts scorable responses across the 2,201 benchmark questions.}}\label{tab:full_bounds}\\

\arrayrulecolor{black!70}
\toprule
{\fontsize{8}{9.44}\selectfont\bfseries Configuration} & {\fontsize{8}{9.44}\selectfont\bfseries Valid} & {\fontsize{8}{9.44}\selectfont\bfseries Discovery} & {\fontsize{8}{9.44}\selectfont\bfseries Exact} & {\fontsize{8}{9.44}\selectfont\bfseries Labels} & {\fontsize{8}{9.44}\selectfont\bfseries Empty} & {\fontsize{8}{9.44}\selectfont\bfseries Nonempty} \\
\midrule
\endfirsthead
\caption[]{(continued)}\\

\arrayrulecolor{black!70}
\toprule
{\fontsize{8}{9.44}\selectfont\bfseries Configuration} & {\fontsize{8}{9.44}\selectfont\bfseries Valid} & {\fontsize{8}{9.44}\selectfont\bfseries Discovery} & {\fontsize{8}{9.44}\selectfont\bfseries Exact} & {\fontsize{8}{9.44}\selectfont\bfseries Labels} & {\fontsize{8}{9.44}\selectfont\bfseries Empty} & {\fontsize{8}{9.44}\selectfont\bfseries Nonempty} \\
\midrule
\endhead
\midrule
\endfoot
\endlastfoot

\rowcolor{EgoTableFrames}\multicolumn{7}{l}{\textbf{Chronological frames}} \\
GPT-6 Astra & 2201 & 55.69--55.81 & 48.26 & 82.80--83.09 & 69.67 & 47.57--47.75 \\
GPT-5.6 Sol & 2201 & 44.48--44.57 & 35.54 & 74.46--74.55 & 46.92 & 43.07--43.20 \\
Gemini 3.8 Flash & 2201 & 51.74--52.24 & 43.73--43.90 & 74.47--76.47 & 72.04 & 39.94--40.73 \\
Gemini 3.1 Pro & 2201 & 44.45--46.08 & 33.97--34.67 & 64.22--68.38 & 46.45 & 43.30--45.87 \\
Llama 4 Maverick & 2200 & 28.08--31.59 & 14.29--15.33 & 33.56--44.06 & 19.91 & 32.83--38.38 \\
Llama 4 Scout & 2201 & 38.59--41.43 & 28.40--29.44 & 44.53--51.94 & 53.55 & 29.89--34.39 \\
Qwen3.8 Max & 2193 & 43.50--44.05 & 33.45 & 70.50--71.54 & 38.86 & 46.19--47.06 \\
Qwen3.8 Flash & 2193 & 39.40--40.65 & 29.44--29.62 & 65.98--69.42 & 34.12 & 42.47--44.44 \\
\midrule
\rowcolor{EgoTableVideo}\multicolumn{7}{l}{\textbf{Native video}} \\
Qwen3.8 Max & 2201 & 39.66--40.53 & 28.92--29.09 & 57.07--59.18 & 45.97 & 36.00--37.37 \\
GLM-5.3 Flash & 2201 & 39.02--40.30 & 29.62--30.31 & 46.53--50.34 & 57.35 & 28.37--30.39 \\
MiniMax M3 & 2201 & 31.18--32.46 & 26.13--26.31 & 44.64--51.43 & 61.14 & 13.77--15.79 \\
Gemma 4 31B & 2200 & 36.30--37.80 & 28.75--28.92 & 49.52--55.67 & 60.19 & 22.41--24.79 \\
Qwen3.5-397B & 2201 & 28.95--29.53 & 24.22--24.56 & 46.03--50.77 & 55.45 & 13.54--14.46 \\
Qwen3.8 Flash & 2201 & 34.09--35.69 & 28.92--29.44 & 47.62--55.56 & 65.40 & 15.89--18.41 \\
\bottomrule
\end{longtable}
\endgroup\endgroup

\endgroup\par\addvspace{8pt}

\par\addvspace{8pt}\begingroup

\noindent\begin{minipage}{\linewidth}\normalfont\normalsize\raggedright
\textbf{Determined-score estimates and denominators.} Table~\ref{tab:ffg_determined} reports the estimates used in the main result table together with their confidence intervals. The denominator is metric-specific: $n_D$ and $n_E$ count questions with determined scores, whereas $n_L$ counts matched entities in the determined label ratio. This distinction explains why a configuration can have different coverage for Discovery, Exact and Labels.
\end{minipage}\par\smallskip

\color{RevisionBlue}
\centering\color{RevisionBlue}

\begingroup\scriptsize
\setlength{\tabcolsep}{3.2pt}\renewcommand{\arraystretch}{1.10}
\begingroup
\EgoTableSetup{8.5}{2.4}
\begin{longtable}{@{\hspace{\tabcolsep}}>{\raggedright\arraybackslash}p{\dimexpr0.310506624\linewidth-4.036586111\tabcolsep\relax}>{\centering\arraybackslash}p{\dimexpr0.077725629\linewidth-1.010433172\tabcolsep\relax}>{\columncolor{black!5}\centering\arraybackslash}p{\dimexpr0.171402658\linewidth-2.228234558\tabcolsep\relax}>{\centering\arraybackslash}p{\dimexpr0.075616649\linewidth-0.983016436\tabcolsep\relax}>{\centering\arraybackslash}p{\dimexpr0.134380484\linewidth-1.746946291\tabcolsep\relax}>{\centering\arraybackslash}p{\dimexpr0.075616649\linewidth-0.983016436\tabcolsep\relax}>{\centering\arraybackslash}p{\dimexpr0.154751307\linewidth-2.011766996\tabcolsep\relax}@{}}
\caption{\rev{FFG point estimates from Table~\ref{tab:nohistory} and their 95\% confidence intervals from grouped resampling (\%). $n_D$ and $n_E$ count determined question scores for Discovery and Exact. $n_L$ counts matched entities in the determined label ratio.}}\label{tab:ffg_determined}\\

\arrayrulecolor{black!70}
\toprule
{\fontsize{8}{9.44}\selectfont\bfseries Configuration} & {\fontsize{8}{9.44}\selectfont\bfseries $n_D$} & {\fontsize{8}{9.44}\selectfont\bfseries Discovery} & {\fontsize{8}{9.44}\selectfont\bfseries $n_E$} & {\fontsize{8}{9.44}\selectfont\bfseries Exact} & {\fontsize{8}{9.44}\selectfont\bfseries $n_L$} & {\fontsize{8}{9.44}\selectfont\bfseries Labels} \\
\midrule
\endfirsthead
\caption[]{(continued)}\\

\arrayrulecolor{black!70}
\toprule
{\fontsize{8}{9.44}\selectfont\bfseries Configuration} & {\fontsize{8}{9.44}\selectfont\bfseries $n_D$} & {\fontsize{8}{9.44}\selectfont\bfseries Discovery} & {\fontsize{8}{9.44}\selectfont\bfseries $n_E$} & {\fontsize{8}{9.44}\selectfont\bfseries Exact} & {\fontsize{8}{9.44}\selectfont\bfseries $n_L$} & {\fontsize{8}{9.44}\selectfont\bfseries Labels} \\
\midrule
\endhead
\midrule
\endfoot
\endlastfoot

\rowcolor{EgoTableFrames}\multicolumn{7}{l}{\textbf{Chronological frames}} \\
GPT-6 Astra & 573 & 55.7$\pm$6.7 & 574 & 48.3$\pm$6.9 & 276 & 83.0$\pm$7.1 \\
GPT-5.6 Sol & 573 & 44.6$\pm$6.1 & 574 & 35.5$\pm$5.6 & 278 & 74.5$\pm$6.6 \\
Gemini 3.8 Flash & 570 & 52.1$\pm$5.8 & 573 & 43.8$\pm$6.1 & 184 & 76.1$\pm$8.9 \\
Gemini 3.1 Pro & 562 & 45.4$\pm$5.7 & 570 & 34.2$\pm$5.0 & 220 & 67.3$\pm$7.2 \\
Llama 4 Maverick & 547 & 29.5$\pm$5.6 & 568 & 14.4$\pm$4.0 & 130 & 38.5$\pm$5.8 \\
Llama 4 Scout & 555 & 39.9$\pm$4.9 & 568 & 28.7$\pm$4.8 & 123 & 49.6$\pm$5.6 \\
Qwen3.8 Max & 569 & 43.9$\pm$7.1 & 574 & 33.4$\pm$5.8 & 256 & 71.5$\pm$6.7 \\
Qwen3.8 Flash & 563 & 40.2$\pm$5.7 & 573 & 29.5$\pm$5.0 & 232 & 69.0$\pm$7.6 \\
\midrule
\rowcolor{EgoTableVideo}\multicolumn{7}{l}{\textbf{Native video}} \\
Qwen3.8 Max & 568 & 40.1$\pm$5.4 & 573 & 29.0$\pm$4.4 & 191 & 59.2$\pm$8.7 \\
GLM-5.3 Flash & 565 & 39.6$\pm$5.6 & 570 & 29.8$\pm$5.2 & 138 & 48.6$\pm$7.8 \\
MiniMax M3 & 565 & 31.7$\pm$4.6 & 573 & 26.2$\pm$4.7 & 98 & 50.0$\pm$9.6 \\
Gemma 4 31B & 563 & 37.0$\pm$5.0 & 573 & 28.8$\pm$5.1 & 95 & 54.7$\pm$12.3 \\
Qwen3.5-397B & 569 & 29.2$\pm$4.5 & 572 & 24.3$\pm$4.3 & 61 & 47.5$\pm$10.8 \\
Qwen3.8 Flash & 562 & 34.8$\pm$5.2 & 571 & 29.1$\pm$5.1 & 76 & 52.6$\pm$12.0 \\
\bottomrule
\end{longtable}
\endgroup\endgroup

\endgroup\par\addvspace{8pt}

\par\addvspace{8pt}\begingroup

\noindent\begin{minipage}{\linewidth}\normalfont\normalsize\raggedright
\textbf{Entity-level precision and recall.} Table~\ref{tab:prfull} separates the cost of extra reports from the coverage of reference entities. Precision decreases when additional entities are reported, while recall decreases when reference entities are missed. F1 summarizes these two entity-level quantities; it is distinct from the question-averaged Discovery score, whose empty-set rule and false-positive penalty are defined in Eq.~\ref{eq:object_score}.
\end{minipage}\par\smallskip

\color{RevisionBlue}
\centering\color{RevisionBlue}

\begingroup\footnotesize
\setlength{\tabcolsep}{3.2pt}\renewcommand{\arraystretch}{1.10}
\begingroup
\EgoTableSetup{9}{2.4}
\begin{longtable}{@{\hspace{\tabcolsep}}>{\raggedright\arraybackslash}p{\dimexpr0.320076805\linewidth-2.240537636\tabcolsep\relax}>{\centering\arraybackslash}p{\dimexpr0.226641065\linewidth-1.586487455\tabcolsep\relax}>{\centering\arraybackslash}p{\dimexpr0.226641065\linewidth-1.586487455\tabcolsep\relax}>{\centering\arraybackslash}p{\dimexpr0.226641065\linewidth-1.586487455\tabcolsep\relax}@{}}
\caption{\rev{Entity-level micro-precision, recall and F1 (\%) under feasible assignments. The 574 questions have 431 reference entities. An extra report contributes a false positive, including when the reference is empty.}}\label{tab:prfull}\\

\arrayrulecolor{black!70}
\toprule
{\fontsize{8.5}{10.03}\selectfont\bfseries Configuration} & {\fontsize{8.5}{10.03}\selectfont\bfseries Precision} & {\fontsize{8.5}{10.03}\selectfont\bfseries Recall} & {\fontsize{8.5}{10.03}\selectfont\bfseries F1} \\
\midrule
\endfirsthead
\caption[]{(continued)}\\

\arrayrulecolor{black!70}
\toprule
{\fontsize{8.5}{10.03}\selectfont\bfseries Configuration} & {\fontsize{8.5}{10.03}\selectfont\bfseries Precision} & {\fontsize{8.5}{10.03}\selectfont\bfseries Recall} & {\fontsize{8.5}{10.03}\selectfont\bfseries F1} \\
\midrule
\endhead
\midrule
\endfoot
\endlastfoot

\rowcolor{EgoTableFrames}\multicolumn{4}{l}{\textbf{Chronological frames}} \\
GPT-6 Astra & 55.27--55.47 & 64.50--64.73 & 59.53--59.74 \\
GPT-5.6 Sol & 45.50--45.66 & 64.50--64.73 & 53.36--53.55 \\
Gemini 3.8 Flash & 57.14--59.01 & 42.69--44.08 & 48.87--50.46 \\
Gemini 3.1 Pro & 45.71--48.78 & 51.97--55.45 & 48.64--51.90 \\
Llama 4 Maverick & 28.14--33.98 & 30.16--36.43 & 29.12--35.16 \\
Llama 4 Scout & 36.07--41.94 & 28.54--33.18 & 31.87--37.05 \\
Qwen3.8 Max & 44.33--45.19 & 59.86--61.02 & 50.94--51.92 \\
Qwen3.8 Flash & 40.31--42.37 & 54.52--57.31 & 46.35--48.72 \\
\midrule
\rowcolor{EgoTableVideo}\multicolumn{4}{l}{\textbf{Native video}} \\
Qwen3.8 Max & 42.23--43.98 & 44.78--46.64 & 43.47--45.27 \\
GLM-5.3 Flash & 41.62--44.61 & 32.25--34.57 & 36.34--38.95 \\
MiniMax M3 & 29.71--33.53 & 23.43--26.45 & 26.20--29.57 \\
Gemma 4 31B & 39.26--43.80 & 22.04--24.59 & 28.23--31.50 \\
Qwen3.5-397B & 19.12--20.69 & 14.15--15.31 & 16.27--17.60 \\
Qwen3.8 Flash & 34.55--40.00 & 17.63--20.42 & 23.35--27.04 \\
\bottomrule
\end{longtable}
\endgroup\endgroup

\endgroup\par\addvspace{8pt}

\FloatBarrier
\subsection{Human Responses and Control Baselines}\label{app:expanded_humans}
Participant A assigns the correct change to 309 of 311 matched entities, and B to 329 of 337. Their labels agree on 286 of the 295 reference entities they both identify. Label capitalization does not affect scoring.

\par\addvspace{8pt}\begingroup

\noindent\begin{minipage}{\linewidth}\normalfont\normalsize\raggedright
\textbf{Independent human responses.} Table~\ref{tab:human_expanded} reports each participant separately before taking their equally weighted mean. Both participants discover entities without candidate-object cues. Their conditional label scores exclude omitted and extra entities, so these scores should be read together with Discovery and Exact when assessing complete responses.
\end{minipage}\par\smallskip

\color{RevisionBlue}

\color{RevisionBlue}
\centering\color{RevisionBlue}

\begingroup\scriptsize
\setlength{\tabcolsep}{3.2pt}\renewcommand{\arraystretch}{1.10}
\begingroup
\EgoTableSetup{8.5}{1.8}
\begingroup
\EgoTableSetup{8.5}{1.8}
\begin{longtable}{@{\hspace{\tabcolsep}}>{\raggedright\arraybackslash}p{\dimexpr0.200358246\linewidth-3.406090190\tabcolsep\relax}>{\columncolor{black!5}\centering\arraybackslash}p{\dimexpr0.080164855\linewidth-1.362802527\tabcolsep\relax}>{\centering\arraybackslash}p{\dimexpr0.108153139\linewidth-1.838603370\tabcolsep\relax}>{\centering\arraybackslash}p{\dimexpr0.108228782\linewidth-1.839889302\tabcolsep\relax}>{\centering\arraybackslash}p{\dimexpr0.098369847\linewidth-1.672287397\tabcolsep\relax}>{\centering\arraybackslash}p{\dimexpr0.109590375\linewidth-1.863036378\tabcolsep\relax}>{\centering\arraybackslash}p{\dimexpr0.080164855\linewidth-1.362802527\tabcolsep\relax}>{\centering\arraybackslash}p{\dimexpr0.124971325\linewidth-2.124512520\tabcolsep\relax}>{\centering\arraybackslash}p{\dimexpr0.089998576\linewidth-1.529975788\tabcolsep\relax}@{}}
\caption{\rev{Free-response human baseline on 2,201 questions (\%). The 574 FFG questions use the same definitions as the models. Conditional Labels excludes omissions and extras.}}\label{tab:human_expanded}\\

\arrayrulecolor{black!70}
\toprule
{\fontsize{8}{9.44}\selectfont\bfseries Respondent} & {\fontsize{8}{9.44}\selectfont\bfseries MCQ} & {\fontsize{8}{9.44}\selectfont\bfseries \shortstack{Wearer\\translation}} & {\fontsize{8}{9.44}\selectfont\bfseries Turning} & {\fontsize{8}{9.44}\selectfont\bfseries \shortstack{Object\\motion}} & {\fontsize{8}{9.44}\selectfont\bfseries \shortstack{Passage\\relation}} & {\fontsize{8}{9.44}\selectfont\bfseries Exact} & {\fontsize{8}{9.44}\selectfont\bfseries Discovery} & {\fontsize{8}{9.44}\selectfont\bfseries Labels} \\
\midrule
\endfirsthead
\caption[]{(continued)}\\

\arrayrulecolor{black!70}
\toprule
{\fontsize{8}{9.44}\selectfont\bfseries Respondent} & {\fontsize{8}{9.44}\selectfont\bfseries MCQ} & {\fontsize{8}{9.44}\selectfont\bfseries \shortstack{Wearer\\translation}} & {\fontsize{8}{9.44}\selectfont\bfseries Turning} & {\fontsize{8}{9.44}\selectfont\bfseries \shortstack{Object\\motion}} & {\fontsize{8}{9.44}\selectfont\bfseries \shortstack{Passage\\relation}} & {\fontsize{8}{9.44}\selectfont\bfseries Exact} & {\fontsize{8}{9.44}\selectfont\bfseries Discovery} & {\fontsize{8}{9.44}\selectfont\bfseries Labels} \\
\midrule
\endhead
\midrule
\endfoot
\endlastfoot

Participant A & 85.00 & 89.14 & 84.62 & 94.35 & 77.63 & 75.96 & 77.38 & 99.36 \\
Participant B & 86.60 & 90.64 & 81.92 & 93.88 & 82.22 & 79.62 & 81.53 & 97.63 \\
Two-person mean & 85.80 & 89.89 & 83.27 & 94.12 & 79.93 & 77.79 & 79.46 & 98.49 \\
\bottomrule
\end{longtable}
\endgroup
\endgroup\endgroup

\endgroup\par\addvspace{8pt}

\par\addvspace{8pt}\begingroup

\noindent\begin{minipage}{\linewidth}\normalfont\normalsize\raggedright
\textbf{Uncertainty of the human baseline.} Table~\ref{tab:human_sampling} applies the same grouped-resampling procedure to the human responses and separates empty from nonempty FFG questions. The participants stay fixed across resamples. The intervals therefore quantify variation over the sampled clips, rather than uncertainty over a larger population of human respondents.
\end{minipage}\par\smallskip

\color{RevisionBlue}
\centering\color{RevisionBlue}

\begingroup\scriptsize
\setlength{\tabcolsep}{3.2pt}\renewcommand{\arraystretch}{1.10}
\begingroup
\EgoTableSetup{8.5}{2.4}
\begin{longtable}{@{\hspace{\tabcolsep}}>{\raggedright\arraybackslash}p{\dimexpr0.238278024\linewidth-3.097614311\tabcolsep\relax}>{\columncolor{black!5}\centering\arraybackslash}p{\dimexpr0.105037216\linewidth-1.365483805\tabcolsep\relax}>{\centering\arraybackslash}p{\dimexpr0.133975243\linewidth-1.741678160\tabcolsep\relax}>{\centering\arraybackslash}p{\dimexpr0.105037216\linewidth-1.365483805\tabcolsep\relax}>{\centering\arraybackslash}p{\dimexpr0.171976073\linewidth-2.235688945\tabcolsep\relax}>{\centering\arraybackslash}p{\dimexpr0.105037216\linewidth-1.365483805\tabcolsep\relax}>{\centering\arraybackslash}p{\dimexpr0.140659013\linewidth-1.828567168\tabcolsep\relax}@{}}
\caption{\rev{Human estimates with group-resampled 95\% sampling intervals (\%). Empty and Nonempty refer to Discovery on 211 and 363 questions. Labels conditions on matched entities.}}\label{tab:human_sampling}\\

\arrayrulecolor{black!70}
\toprule
{\fontsize{8}{9.44}\selectfont\bfseries Respondent} & {\fontsize{8}{9.44}\selectfont\bfseries MCQ} & {\fontsize{8}{9.44}\selectfont\bfseries Discovery} & {\fontsize{8}{9.44}\selectfont\bfseries Exact} & {\fontsize{8}{9.44}\selectfont\bfseries Labels} & {\fontsize{8}{9.44}\selectfont\bfseries Empty} & {\fontsize{8}{9.44}\selectfont\bfseries Nonempty} \\
\midrule
\endfirsthead
\caption[]{(continued)}\\

\arrayrulecolor{black!70}
\toprule
{\fontsize{8}{9.44}\selectfont\bfseries Respondent} & {\fontsize{8}{9.44}\selectfont\bfseries MCQ} & {\fontsize{8}{9.44}\selectfont\bfseries Discovery} & {\fontsize{8}{9.44}\selectfont\bfseries Exact} & {\fontsize{8}{9.44}\selectfont\bfseries Labels} & {\fontsize{8}{9.44}\selectfont\bfseries Empty} & {\fontsize{8}{9.44}\selectfont\bfseries Nonempty} \\
\midrule
\endhead
\midrule
\endfoot
\endlastfoot

Participant A & 85.0$\pm$1.8 & 77.4$\pm$3.6 & 76.0$\pm$3.8 & [98.5, 100.0] & 82.5$\pm$5.2 & 74.4$\pm$5.0 \\
Participant B & 86.6$\pm$1.6 & 81.5$\pm$2.7 & 79.6$\pm$2.9 & 97.6$\pm$2.2 & 84.8$\pm$4.9 & 79.6$\pm$3.8 \\
Two-person mean & 85.8$\pm$1.4 & 79.5$\pm$3.0 & 77.8$\pm$3.0 & 98.5$\pm$1.1 & 83.6$\pm$4.9 & 77.0$\pm$4.1 \\
\bottomrule
\end{longtable}
\endgroup\endgroup
\par\smallskip\parbox{\linewidth}{\footnotesize Task-specific MCQ scores for the two-person mean (95\% CI, percentage points): wearer translation 89.9$\pm$3.6, turning 83.3$\pm$3.5, object motion 94.1$\pm$2.1, passage relations 79.9$\pm$2.5.}

\endgroup\par\addvspace{8pt}

\par\addvspace{8pt}\begingroup

\noindent\begin{minipage}{\linewidth}\normalfont\normalsize\raggedright
\textbf{Agreement between respondents.} Table~\ref{tab:human_agreement_expanded} provides agreement rates and Cohen\textquotesingle s $\kappa$ for the independent answers. Turning is compared as a semantic sequence rather than as a literal string. The final FFG row includes only jointly identified entities; it measures agreement on their change labels and does not measure agreement on the complete discovered sets.
\end{minipage}\par\smallskip

\color{RevisionBlue}
\centering\color{RevisionBlue}

\begingroup\footnotesize
\setlength{\tabcolsep}{3.2pt}\renewcommand{\arraystretch}{1.10}
\begingroup
\EgoTableSetup{9}{2.4}
\begin{longtable}{@{\hspace{\tabcolsep}}>{\raggedright\arraybackslash}p{\dimexpr0.536136813\linewidth-3.752957691\tabcolsep\relax}>{\centering\arraybackslash}p{\dimexpr0.084034027\linewidth-0.588238191\tabcolsep\relax}>{\centering\arraybackslash}p{\dimexpr0.229801663\linewidth-1.608611640\tabcolsep\relax}>{\centering\arraybackslash}p{\dimexpr0.150027497\linewidth-1.050192478\tabcolsep\relax}@{}}
\caption{\rev{Agreement between independent respondents. Turning is compared using semantic event sequences (head/path, direction and event order). The final row conditions on objects identified by both respondents.}}\label{tab:human_agreement_expanded}\\

\arrayrulecolor{black!70}
\toprule
{\fontsize{8.5}{10.03}\selectfont\bfseries Comparison} & {\fontsize{8.5}{10.03}\selectfont\bfseries Units} & {\fontsize{8.5}{10.03}\selectfont\bfseries Agreement (\%)} & {\fontsize{8.5}{10.03}\selectfont\bfseries Cohen's $\kappa$} \\
\midrule
\endfirsthead
\caption[]{(continued)}\\

\arrayrulecolor{black!70}
\toprule
{\fontsize{8.5}{10.03}\selectfont\bfseries Comparison} & {\fontsize{8.5}{10.03}\selectfont\bfseries Units} & {\fontsize{8.5}{10.03}\selectfont\bfseries Agreement (\%)} & {\fontsize{8.5}{10.03}\selectfont\bfseries Cohen's $\kappa$} \\
\midrule
\endhead
\midrule
\endfoot
\endlastfoot

Wearer translation & 267 & 94.76 & 0.875 \\
Turning sequence & 260 & 82.69 & 0.745 \\
Object motion & 425 & 94.82 & 0.912 \\
Passage relation & 675 & 82.07 & 0.750 \\
FFG empty/nonempty & 574 & 93.38 & 0.866 \\
FFG label on jointly matched entities & 295 & 96.95 & 0.946 \\
\bottomrule
\end{longtable}
\endgroup\endgroup

\endgroup\par\addvspace{8pt}

\par\addvspace{8pt}\begingroup

\noindent\begin{minipage}{\linewidth}\normalfont\normalsize\raggedright
\textbf{Question-only control.} Table~\ref{tab:qonly} compares the same model on the same MCQs with and without visual frames. This control separates performance obtainable from the questions and answer options from performance with visual evidence. The same corrected turning question is excluded in both input conditions, and the task-specific columns show where visual input changes performance.
\end{minipage}\par\smallskip

\color{RevisionBlue}

\color{RevisionBlue}
\centering\color{RevisionBlue}

\begingroup\footnotesize
\setlength{\tabcolsep}{3.2pt}\renewcommand{\arraystretch}{1.10}
\begingroup
\EgoTableSetup{8.5}{2.4}
\begingroup
\EgoTableSetup{8.5}{2.4}
\begin{longtable}{@{\hspace{\tabcolsep}}>{\raggedright\arraybackslash}p{\dimexpr0.174533972\linewidth-2.268941630\tabcolsep\relax}>{\raggedright\arraybackslash}p{\dimexpr0.193781079\linewidth-2.519154023\tabcolsep\relax}>{\columncolor{black!5}\centering\arraybackslash}p{\dimexpr0.108770272\linewidth-1.414013535\tabcolsep\relax}>{\centering\arraybackslash}p{\dimexpr0.133276567\linewidth-1.732595367\tabcolsep\relax}>{\centering\arraybackslash}p{\dimexpr0.133369781\linewidth-1.733807156\tabcolsep\relax}>{\centering\arraybackslash}p{\dimexpr0.121220665\linewidth-1.575868643\tabcolsep\relax}>{\centering\arraybackslash}p{\dimexpr0.135047665\linewidth-1.755619646\tabcolsep\relax}@{}}
\caption{\rev{GPT-5.6 Sol accuracy with group-resampled 95\% confidence intervals on the same 1,627 MCQs (\%). Turning answer options vary across questions. Both input conditions exclude the same turning question whose corrected reference has no matching answer option.}}\label{tab:qonly}\\

\arrayrulecolor{black!70}
\toprule
{\fontsize{8}{9.44}\selectfont\bfseries System} & {\fontsize{8}{9.44}\selectfont\bfseries Input} & {\fontsize{8}{9.44}\selectfont\bfseries MCQ} & {\fontsize{8}{9.44}\selectfont\bfseries \shortstack{Wearer\\translation}} & {\fontsize{8}{9.44}\selectfont\bfseries Turning} & {\fontsize{8}{9.44}\selectfont\bfseries \shortstack{Object\\motion}} & {\fontsize{8}{9.44}\selectfont\bfseries \shortstack{Passage\\relation}} \\
\midrule
\endfirsthead
\caption[]{(continued)}\\

\arrayrulecolor{black!70}
\toprule
{\fontsize{8}{9.44}\selectfont\bfseries System} & {\fontsize{8}{9.44}\selectfont\bfseries Input} & {\fontsize{8}{9.44}\selectfont\bfseries MCQ} & {\fontsize{8}{9.44}\selectfont\bfseries \shortstack{Wearer\\translation}} & {\fontsize{8}{9.44}\selectfont\bfseries Turning} & {\fontsize{8}{9.44}\selectfont\bfseries \shortstack{Object\\motion}} & {\fontsize{8}{9.44}\selectfont\bfseries \shortstack{Passage\\relation}} \\
\midrule
\endhead
\midrule
\endfoot
\endlastfoot

GPT-5.6 Sol & Question only & 45.5$\pm$2.8 & 49.8$\pm$6.8 & 8.5$\pm$3.3 & 78.4$\pm$4.9 & 37.5$\pm$4.5 \\
GPT-5.6 Sol & With frames & 57.5$\pm$2.9 & 61.0$\pm$7.8 & 33.8$\pm$5.9 & 86.6$\pm$4.7 & 46.8$\pm$4.1 \\
\bottomrule
\end{longtable}
\endgroup
\endgroup\endgroup

\endgroup\par\addvspace{8pt}

\par\addvspace{8pt}\begingroup

\noindent\begin{minipage}{\linewidth}\normalfont\normalsize\raggedright
\textbf{Empty-set control.} Table~\ref{tab:emptycontrols} makes the always-empty baseline explicit. It receives full Discovery credit on empty references and none on nonempty references. Reporting both subsets prevents a combined score from obscuring failure to discover entities when an entity is actually present. The human comparison and Astra identity bounds use the same scoring definitions.
\end{minipage}\par\smallskip

\color{RevisionBlue}
\centering\color{RevisionBlue}

\begingroup\small
\setlength{\tabcolsep}{3.2pt}\renewcommand{\arraystretch}{1.10}
\begingroup
\EgoTableSetup{9}{2.4}
\begin{longtable}{@{\hspace{\tabcolsep}}>{\raggedright\arraybackslash}p{\dimexpr0.258136824\linewidth-1.806957765\tabcolsep\relax}>{\centering\arraybackslash}p{\dimexpr0.224095071\linewidth-1.568665496\tabcolsep\relax}>{\centering\arraybackslash}p{\dimexpr0.287002707\linewidth-2.009018950\tabcolsep\relax}>{\columncolor{black!5}\centering\arraybackslash}p{\dimexpr0.230765398\linewidth-1.615357788\tabcolsep\relax}@{}}
\caption{\rev{Discovery by reference-set status (\%). Always-empty Discovery and Exact are 36.8$\pm$6.6 (95\% CI, percentage points). Model bounds reflect unresolved identity. The free-response human mean exceeds Astra's upper bound by $23.65\pm7.45$ percentage points (paired 95\% CI).}}\label{tab:emptycontrols}\\

\arrayrulecolor{black!70}
\toprule
{\fontsize{8.5}{10.03}\selectfont\bfseries System} & {\fontsize{8.5}{10.03}\selectfont\bfseries Empty (211)} & {\fontsize{8.5}{10.03}\selectfont\bfseries Nonempty (363)} & {\fontsize{8.5}{10.03}\selectfont\bfseries All (574)} \\
\midrule
\endfirsthead
\caption[]{(continued)}\\

\arrayrulecolor{black!70}
\toprule
{\fontsize{8.5}{10.03}\selectfont\bfseries System} & {\fontsize{8.5}{10.03}\selectfont\bfseries Empty (211)} & {\fontsize{8.5}{10.03}\selectfont\bfseries Nonempty (363)} & {\fontsize{8.5}{10.03}\selectfont\bfseries All (574)} \\
\midrule
\endhead
\midrule
\endfoot
\endlastfoot

Always empty & 100.00 & 0.00 & 36.76 \\
GPT-6 Astra & 69.67 & 47.57--47.75 & 55.69--55.81 \\
Human mean & 83.65 & 77.02 & 79.46 \\
\bottomrule
\end{longtable}
\endgroup\endgroup

\endgroup\par\addvspace{8pt}

}

\FloatBarrier
\subsection{Visual-History Evaluation}
\label{app:history_full}
The History evaluation covers 293 questions across 54 current \rev{evaluation clips} with an immediately preceding clip. These are the same questions evaluated without history, not additional benchmark questions. In task order, the counts are 55, 53, 64, 67, and 54. Table~\ref{tab:history_full} reports all scorable answers for each model. Missing responses and one Llama 4 Scout response whose entity identity could not be established are excluded; the shared comparison in the main text instead uses the same 257 questions for every model and both conditions.
\par\addvspace{8pt}\begingroup

\noindent\begin{minipage}{\linewidth}\normalfont\normalsize\raggedright
\textbf{Full History coverage.} Table~\ref{tab:history_full} reports every scorable History answer and identifies exclusions in the lower panel. Model-specific denominators can differ in this table. Its auxiliary aggregate combines MCQ correct counts and summed Discovery scores; it should not be confused with the fixed-question paired comparison used in the main text.
\end{minipage}\par\smallskip

\color{RevisionBlue}

\color{RevisionBlue}
\centering\small

\begingroup
\setlength{\tabcolsep}{3pt}
\renewcommand{\arraystretch}{1.12}
\footnotesize\setlength{\tabcolsep}{2pt}
\begingroup
\EgoTableSetup{8.5}{1.8}
\begingroup
\EgoTableSetup{8.5}{1.8}
\begin{longtable}{@{\hspace{\tabcolsep}}>{\raggedright\arraybackslash}p{\dimexpr0.192224877\linewidth-3.267822903\tabcolsep\relax}>{\columncolor{black!5}\centering\arraybackslash}p{\dimexpr0.088538194\linewidth-1.505149299\tabcolsep\relax}>{\centering\arraybackslash}p{\dimexpr0.101869395\linewidth-1.731779710\tabcolsep\relax}>{\centering\arraybackslash}p{\dimexpr0.101940643\linewidth-1.732990929\tabcolsep\relax}>{\centering\arraybackslash}p{\dimexpr0.092654516\linewidth-1.575126768\tabcolsep\relax}>{\centering\arraybackslash}p{\dimexpr0.103223126\linewidth-1.754793150\tabcolsep\relax}>{\centering\arraybackslash}p{\dimexpr0.071564793\linewidth-1.216601483\tabcolsep\relax}>{\centering\arraybackslash}p{\dimexpr0.117710436\linewidth-2.001077413\tabcolsep\relax}>{\centering\arraybackslash}p{\dimexpr0.130274020\linewidth-2.214658345\tabcolsep\relax}@{}}
\caption{History results over each model's scorable set. Score formats follow Table~\ref{tab:history}; label counts appear in Table~\ref{tab:history_ffg_counts}. The lower panel gives the auxiliary aggregate numerator (MCQ correct plus summed Discovery scores) and denominator.}\label{tab:history_full}\\

\arrayrulecolor{black!70}
\toprule
{\fontsize{8}{9.44}\selectfont\bfseries Model} & {\fontsize{8}{9.44}\selectfont\bfseries MCQ} & {\fontsize{8}{9.44}\selectfont\bfseries \shortstack{Wearer\\translation}} & {\fontsize{8}{9.44}\selectfont\bfseries Turning} & {\fontsize{8}{9.44}\selectfont\bfseries \shortstack{Object\\motion}} & {\fontsize{8}{9.44}\selectfont\bfseries \shortstack{Passage\\relation}} & {\fontsize{8}{9.44}\selectfont\bfseries \shortstack{FFG\\Exact}} & {\fontsize{8}{9.44}\selectfont\bfseries Discovery} & {\fontsize{8}{9.44}\selectfont\bfseries Labels (\%)} \\
\midrule
\endfirsthead
\caption[]{(continued)}\\

\arrayrulecolor{black!70}
\toprule
{\fontsize{8}{9.44}\selectfont\bfseries Model} & {\fontsize{8}{9.44}\selectfont\bfseries MCQ} & {\fontsize{8}{9.44}\selectfont\bfseries \shortstack{Wearer\\translation}} & {\fontsize{8}{9.44}\selectfont\bfseries Turning} & {\fontsize{8}{9.44}\selectfont\bfseries \shortstack{Object\\motion}} & {\fontsize{8}{9.44}\selectfont\bfseries \shortstack{Passage\\relation}} & {\fontsize{8}{9.44}\selectfont\bfseries \shortstack{FFG\\Exact}} & {\fontsize{8}{9.44}\selectfont\bfseries Discovery} & {\fontsize{8}{9.44}\selectfont\bfseries Labels (\%)} \\
\midrule
\endhead
\midrule
\endfoot
\endlastfoot

\rowcolor{EgoTableVideo}\multicolumn{9}{l}{\textbf{History: full coverage}} \\
GPT-6 Astra & \textbf{195/239} & \textbf{45/55} & \underline{42/53} & \textbf{61/64} & \textbf{47/67} & \textbf{38/54} & \textbf{41.50/54} & \textbf{78.95} \\
Gemini 3.8 Flash & \underline{179/239} & \underline{42/55} & \textbf{44/53} & 54/64 & \underline{39/67} & \underline{32/54} & \underline{36.50/54} & 64.71 \\
Gemini 3.1 Pro & 147/239 & 34/55 & 23/53 & 56/64 & 34/67 & 25/54 & 28.50/54 & \underline{76.47} \\
GPT-5.6 Sol & 159/239 & 38/55 & 36/53 & \underline{60/64} & 25/67 & 23/54 & 26.00/54 & 70.00 \\
Qwen3.8 Max & 148/232 & \underline{42/55} & 30/51 & 57/64 & 19/62 & 13/51 & 15.00/51 & 60.00 \\
Qwen3.8 Flash & 134/226 & 38/53 & 24/47 & 55/62 & 17/64 & 5/44 & 9.00/44 & 57.14 \\
Llama 4 Scout & 112/239 & 26/55 & 30/53 & 42/64 & 14/67 & 14/51 & 17.50/51 & 60.00 \\
Llama 4 Maverick & 129/239 & 33/55 & 19/53 & 56/64 & 21/67 & 10/54 & 14.00/54 & 63.64 \\
\bottomrule
\end{longtable}
\endgroup
\endgroup
\par\medskip
\begingroup
\EgoTableSetup{9}{2.4}
\begingroup
\EgoTableSetup{9}{2.4}
\addtocounter{table}{-1}
\begin{longtable}{@{\hspace{\tabcolsep}}>{\raggedright\arraybackslash}p{\dimexpr0.291152464\linewidth-2.038067250\tabcolsep\relax}>{\centering\arraybackslash}p{\dimexpr0.207838154\linewidth-1.454867079\tabcolsep\relax}>{\centering\arraybackslash}p{\dimexpr0.169691378\linewidth-1.187839645\tabcolsep\relax}>{\centering\arraybackslash}p{\dimexpr0.331318004\linewidth-2.319226026\tabcolsep\relax}@{}}
\caption[]{(continued)}\\

\arrayrulecolor{black!70}
\toprule
{\fontsize{8.5}{10.03}\selectfont\bfseries Model} & {\fontsize{8.5}{10.03}\selectfont\bfseries Scorable/all} & {\fontsize{8.5}{10.03}\selectfont\bfseries Excluded} & {\fontsize{8.5}{10.03}\selectfont\bfseries Auxiliary aggregate} \\
\midrule
\endfirsthead
\caption[]{(continued)}\\

\arrayrulecolor{black!70}
\toprule
{\fontsize{8.5}{10.03}\selectfont\bfseries Model} & {\fontsize{8.5}{10.03}\selectfont\bfseries Scorable/all} & {\fontsize{8.5}{10.03}\selectfont\bfseries Excluded} & {\fontsize{8.5}{10.03}\selectfont\bfseries Auxiliary aggregate} \\
\midrule
\endhead
\midrule
\endfoot
\endlastfoot

GPT-6 Astra & 293/293 & 0 & 236.50/293 \\
Gemini 3.8 Flash & 293/293 & 0 & 215.50/293 \\
Gemini 3.1 Pro & 293/293 & 0 & 175.50/293 \\
GPT-5.6 Sol & 293/293 & 0 & 185.00/293 \\
Qwen3.8 Max & 283/293 & 10 & 163.00/283 \\
Qwen3.8 Flash & 270/293 & 23 & 143.00/270 \\
Llama 4 Scout & 290/293 & 3 & 129.50/290 \\
Llama 4 Maverick & 293/293 & 0 & 143.00/293 \\
\bottomrule
\end{longtable}
\endgroup
\endgroup
\endgroup

\endgroup\par\addvspace{8pt}

The shared comparison contains 53 wearer translation, 45 turning, 62 external object motion, 58 passage relation, and 39 discovery questions. FFG label accuracy uses successfully matched reference entities separately in each condition. Humans did not complete the History condition. Table~\ref{tab:history} provides absolute scores for both conditions on this shared set; Figure~\ref{fig:history} visualizes their differences. \rev{For each condition, the camera wearers supplied references using the visual evidence available in that condition.} The same condition-specific references apply to every model.
\par\addvspace{8pt}\begingroup

\noindent\begin{minipage}{\linewidth}\normalfont\normalsize\raggedright
\textbf{Fixed-question paired comparison.} Table~\ref{tab:history} holds the shared question set fixed across models and conditions. Correct/scorable fractions make the MCQ and Exact percentages recoverable, while Discovery is reported over the same FFG questions. Label accuracy remains conditional on the entities successfully matched within each condition.
\end{minipage}\par\smallskip

\color{RevisionBlue}

\color{RevisionBlue}
\centering\small

\begingroup
\setlength{\tabcolsep}{3pt}
\renewcommand{\arraystretch}{1.12}
\footnotesize\setlength{\tabcolsep}{2pt}
\begingroup
\EgoTableSetup{8.5}{1.8}
\begingroup
\EgoTableSetup{8.5}{1.8}
\begin{longtable}{@{\hspace{\tabcolsep}}>{\raggedright\arraybackslash}p{\dimexpr0.192224877\linewidth-3.267822903\tabcolsep\relax}>{\columncolor{black!5}\centering\arraybackslash}p{\dimexpr0.088538194\linewidth-1.505149299\tabcolsep\relax}>{\centering\arraybackslash}p{\dimexpr0.101869395\linewidth-1.731779710\tabcolsep\relax}>{\centering\arraybackslash}p{\dimexpr0.101940643\linewidth-1.732990929\tabcolsep\relax}>{\centering\arraybackslash}p{\dimexpr0.092654516\linewidth-1.575126768\tabcolsep\relax}>{\centering\arraybackslash}p{\dimexpr0.103223126\linewidth-1.754793150\tabcolsep\relax}>{\centering\arraybackslash}p{\dimexpr0.071564793\linewidth-1.216601483\tabcolsep\relax}>{\centering\arraybackslash}p{\dimexpr0.117710436\linewidth-2.001077413\tabcolsep\relax}>{\centering\arraybackslash}p{\dimexpr0.130274020\linewidth-2.214658345\tabcolsep\relax}@{}}
\caption{Paired results on 257 questions. MCQ and FFG Exact give correct/scorable counts; Discovery gives summed scores/scorable counts; Labels gives conditional accuracy (\%; counts in Table~\ref{tab:history_ffg_counts}). Bold/underlining mark best/second-best scores per condition.}\label{tab:history}\\

\arrayrulecolor{black!70}
\toprule
{\fontsize{8}{9.44}\selectfont\bfseries Model} & {\fontsize{8}{9.44}\selectfont\bfseries MCQ} & {\fontsize{8}{9.44}\selectfont\bfseries \shortstack{Wearer\\translation}} & {\fontsize{8}{9.44}\selectfont\bfseries Turning} & {\fontsize{8}{9.44}\selectfont\bfseries \shortstack{Object\\motion}} & {\fontsize{8}{9.44}\selectfont\bfseries \shortstack{Passage\\relation}} & {\fontsize{8}{9.44}\selectfont\bfseries \shortstack{FFG\\Exact}} & {\fontsize{8}{9.44}\selectfont\bfseries Discovery} & {\fontsize{8}{9.44}\selectfont\bfseries Labels (\%)} \\
\midrule
\endfirsthead
\caption[]{(continued)}\\

\arrayrulecolor{black!70}
\toprule
{\fontsize{8}{9.44}\selectfont\bfseries Model} & {\fontsize{8}{9.44}\selectfont\bfseries MCQ} & {\fontsize{8}{9.44}\selectfont\bfseries \shortstack{Wearer\\translation}} & {\fontsize{8}{9.44}\selectfont\bfseries Turning} & {\fontsize{8}{9.44}\selectfont\bfseries \shortstack{Object\\motion}} & {\fontsize{8}{9.44}\selectfont\bfseries \shortstack{Passage\\relation}} & {\fontsize{8}{9.44}\selectfont\bfseries \shortstack{FFG\\Exact}} & {\fontsize{8}{9.44}\selectfont\bfseries Discovery} & {\fontsize{8}{9.44}\selectfont\bfseries Labels (\%)} \\
\midrule
\endhead
\midrule
\endfoot
\endlastfoot

\rowcolor{EgoTableFrames}\multicolumn{9}{l}{\textbf{No History}} \\
GPT-6 Astra & \textbf{174/218} & \textbf{46/53} & \underline{34/45} & 54/62 & \textbf{40/58} & \textbf{24/39} & \textbf{27.50/39} & \textbf{69.23} \\
Gemini 3.8 Flash & \underline{163/218} & 37/53 & \textbf{36/45} & \textbf{57/62} & \underline{33/58} & \underline{21/39} & \underline{22.00/39} & \underline{66.67} \\
Gemini 3.1 Pro & 146/218 & 36/53 & 26/45 & \underline{55/62} & 29/58 & 16/39 & 18.50/39 & \textbf{69.23} \\
GPT-5.6 Sol & 143/218 & 33/53 & 30/45 & \textbf{57/62} & 23/58 & 12/39 & 14.00/39 & \underline{66.67} \\
Qwen3.8 Max & 136/218 & \underline{38/53} & 27/45 & 54/62 & 17/58 & 11/39 & 14.00/39 & 45.45 \\
Qwen3.8 Flash & 130/218 & 36/53 & 23/45 & 52/62 & 19/58 & 4/39 & 6.00/39 & 40.00 \\
Llama 4 Scout & 93/218 & 21/53 & 23/45 & 39/62 & 10/58 & 16/39 & 20.00/39 & 50.00 \\
Llama 4 Maverick & 109/218 & 21/53 & 24/45 & 51/62 & 13/58 & 7/39 & 12.00/39 & 37.50 \\
\midrule
\rowcolor{EgoTableVideo}\multicolumn{9}{l}{\textbf{History}} \\
GPT-6 Astra & \textbf{183/218} & \textbf{44/53} & \underline{38/45} & \textbf{59/62} & \textbf{42/58} & \textbf{26/39} & \textbf{29.50/39} & \underline{71.43} \\
Gemini 3.8 Flash & \underline{170/218} & \underline{41/53} & \textbf{40/45} & 53/62 & \underline{36/58} & \underline{24/39} & \underline{27.00/39} & 64.29 \\
Gemini 3.1 Pro & 143/218 & 33/53 & 22/45 & \underline{56/62} & 32/58 & 18/39 & 21.50/39 & \textbf{75.00} \\
GPT-5.6 Sol & 150/218 & 37/53 & 32/45 & \textbf{59/62} & 22/58 & 16/39 & 18.00/39 & 66.67 \\
Qwen3.8 Max & 143/218 & \underline{41/53} & 28/45 & \underline{56/62} & 18/58 & 11/39 & 13.00/39 & 54.55 \\
Qwen3.8 Flash & 134/218 & 38/53 & 24/45 & 55/62 & 17/58 & 4/39 & 7.00/39 & 54.55 \\
Llama 4 Scout & 102/218 & 24/53 & 26/45 & 41/62 & 11/58 & 11/39 & 13.50/39 & 62.50 \\
Llama 4 Maverick & 121/218 & 32/53 & 18/45 & \underline{56/62} & 15/58 & 8/39 & 11.50/39 & 50.00 \\
\bottomrule
\end{longtable}
\endgroup
\endgroup
\endgroup

\endgroup\par\addvspace{8pt}

\par\addvspace{8pt}\begingroup

\noindent\begin{minipage}{\linewidth}\normalfont\normalsize\raggedright
\textbf{Recovering the History FFG scores.} Table~\ref{tab:history_ffg_counts} gives the question counts, summed object scores, exact-response counts and entity counts behind the paired and full History results. The three panels distinguish paired No History, paired History and full History coverage. Empty/nonempty counts refer to the condition-specific reference sets, which can change when the preceding clip supplies additional evidence.
\end{minipage}\par\smallskip

\color{RevisionBlue}
\centering\small

\begingroup
\setlength{\tabcolsep}{3pt}
\renewcommand{\arraystretch}{1.12}
\footnotesize
\begingroup
\EgoTableSetup{8.5}{2.4}
\begin{longtable}{@{\hspace{\tabcolsep}}>{\raggedright\arraybackslash}p{\dimexpr0.265758723\linewidth-3.454863394\tabcolsep\relax}>{\centering\arraybackslash}p{\dimexpr0.053322820\linewidth-0.693196664\tabcolsep\relax}>{\centering\arraybackslash}p{\dimexpr0.252510072\linewidth-3.282630940\tabcolsep\relax}>{\centering\arraybackslash}p{\dimexpr0.086630174\linewidth-1.126192266\tabcolsep\relax}>{\centering\arraybackslash}p{\dimexpr0.088381892\linewidth-1.148964602\tabcolsep\relax}>{\centering\arraybackslash}p{\dimexpr0.152214988\linewidth-1.978794841\tabcolsep\relax}>{\centering\arraybackslash}p{\dimexpr0.101181330\linewidth-1.315357292\tabcolsep\relax}@{}}
\caption{FFG counts for the paired and full History evaluations. $N$ is the number of scorable questions; empty/nonempty refer to reference sets. Discovery $=100\sum s/N$, FFG Exact $=100\,\mathrm{Exact}/N$, and Labels is 100 times the stated correct/assessable fraction.}\label{tab:history_ffg_counts}\\

\arrayrulecolor{black!70}
\toprule
{\fontsize{8}{9.44}\selectfont\bfseries Model} & {\fontsize{8}{9.44}\selectfont\bfseries $N$} & {\fontsize{8}{9.44}\selectfont\bfseries Empty/nonempty} & {\fontsize{8}{9.44}\selectfont\bfseries $\sum s$} & {\fontsize{8}{9.44}\selectfont\bfseries Exact} & {\fontsize{8}{9.44}\selectfont\bfseries TP/FP/FN} & {\fontsize{8}{9.44}\selectfont\bfseries Labels} \\
\midrule
\endfirsthead
\caption[]{(continued)}\\

\arrayrulecolor{black!70}
\toprule
{\fontsize{8}{9.44}\selectfont\bfseries Model} & {\fontsize{8}{9.44}\selectfont\bfseries $N$} & {\fontsize{8}{9.44}\selectfont\bfseries Empty/nonempty} & {\fontsize{8}{9.44}\selectfont\bfseries $\sum s$} & {\fontsize{8}{9.44}\selectfont\bfseries Exact} & {\fontsize{8}{9.44}\selectfont\bfseries TP/FP/FN} & {\fontsize{8}{9.44}\selectfont\bfseries Labels} \\
\midrule
\endhead
\midrule
\endfoot
\endlastfoot

\rowcolor{EgoTableFrames}\multicolumn{7}{l}{\textbf{Paired No History (257 questions)}} \\
GPT-6 Astra & 39 & 24/15 & 27.50 & 24 & 13/9/4 & 9/13 \\
Gemini 3.8 Flash & 39 & 24/15 & 22.00 & 21 & 12/12/5 & 8/12 \\
Gemini 3.1 Pro & 39 & 24/15 & 18.50 & 16 & 13/20/4 & 9/13 \\
GPT-5.6 Sol & 39 & 24/15 & 14.00 & 12 & 15/27/2 & 10/15 \\
Qwen3.8 Max & 39 & 24/15 & 14.00 & 11 & 11/28/6 & 5/11 \\
Qwen3.8 Flash & 39 & 24/15 & 6.00 & 4 & 10/38/7 & 4/10 \\
Llama 4 Scout & 39 & 24/15 & 20.00 & 16 & 8/14/9 & 4/8 \\
Llama 4 Maverick & 39 & 24/15 & 12.00 & 7 & 8/23/9 & 3/8 \\
\midrule
\rowcolor{EgoTableVideo}\multicolumn{7}{l}{\textbf{Paired History (257 questions)}} \\
GPT-6 Astra & 39 & 23/16 & 29.50 & 26 & 14/7/4 & 10/14 \\
Gemini 3.8 Flash & 39 & 23/16 & 27.00 & 24 & 14/8/4 & 9/14 \\
Gemini 3.1 Pro & 39 & 23/16 & 21.50 & 18 & 12/15/6 & 9/12 \\
GPT-5.6 Sol & 39 & 23/16 & 18.00 & 16 & 15/21/3 & 10/15 \\
Qwen3.8 Max & 39 & 23/16 & 13.00 & 11 & 11/29/7 & 6/11 \\
Qwen3.8 Flash & 39 & 23/16 & 7.00 & 4 & 11/40/7 & 6/11 \\
Llama 4 Scout & 39 & 23/16 & 13.50 & 11 & 8/20/10 & 5/8 \\
Llama 4 Maverick & 39 & 23/16 & 11.50 & 8 & 8/24/10 & 4/8 \\
\midrule
\rowcolor{EgoTableVideo}\multicolumn{7}{l}{\textbf{Full History coverage (293 questions)}} \\
GPT-6 Astra & 54 & 33/21 & 41.50 & 38 & 19/9/5 & 15/19 \\
Gemini 3.8 Flash & 54 & 33/21 & 36.50 & 32 & 17/11/7 & 11/17 \\
Gemini 3.1 Pro & 54 & 33/21 & 28.50 & 25 & 17/23/7 & 13/17 \\
GPT-5.6 Sol & 54 & 33/21 & 26.00 & 23 & 20/30/4 & 14/20 \\
Qwen3.8 Max & 51 & 31/20 & 15.00 & 13 & 15/46/8 & 9/15 \\
Qwen3.8 Flash & 44 & 26/18 & 9.00 & 5 & 14/46/7 & 8/14 \\
Llama 4 Scout & 51 & 32/19 & 17.50 & 14 & 10/29/11 & 6/10 \\
Llama 4 Maverick & 54 & 33/21 & 14.00 & 10 & 11/36/13 & 7/11 \\
\bottomrule
\end{longtable}
\endgroup
\endgroup

\endgroup\par\addvspace{8pt}

\input{figure_history}

\FloatBarrier
\section{\rev{Diagnostic Analyses}}
\label{app:diagnostics}
{\color{RevisionBlue}
\FloatBarrier
\subsection{Obstacles and Walking Encounters}\label{app:expanded_coverage}
The six settings describe where the wearer walked. Tables~\ref{tab:objects} and~\ref{tab:interactions} describe what constrained passage. Object categories describe road users, fixed structures, vegetation and terrain constraints encountered in everyday walking. Curbs, steps and ground boundaries are grouped as terrain constraints because elevation changes and edges of walkable surfaces affect foot placement and available passage space. To check coverage, we cross-reference categories against named question targets and reference entities. Object counts cover 260 of the 284 \rev{evaluation clips}. Each category counts a clip once. Categories can overlap. We inspect the constituent targets within merged categories and the distributions of oncoming, crossing, same-direction and overtaking interactions to identify gaps and sparse subtypes concealed by broad category totals. Types occurring in fewer than six clips are combined as Other, counting each clip once within the combined row.

\par\addvspace{8pt}\begingroup

\noindent\begin{minipage}{\linewidth}\normalfont\normalsize\raggedright
\textbf{Object-type coverage.} Table~\ref{tab:objects} counts the evaluated clips containing each object category and gives Astra\textquotesingle s correct/total passage judgments for targets of that type. Categories are nonexclusive, so clip counts should not be summed as a partition of the benchmark. The correct/total column is a descriptive breakdown over the available targets, with different denominators across categories.
\end{minipage}\par\smallskip

\color{RevisionBlue}

\color{RevisionBlue}
\centering\color{RevisionBlue}

\begingroup\footnotesize
\setlength{\tabcolsep}{3.2pt}\renewcommand{\arraystretch}{1.10}
\begingroup
\EgoTableSetup{9}{2.4}
\begingroup
\EgoTableSetup{9}{2.4}
\begin{longtable}{@{\hspace{\tabcolsep}}>{\raggedright\arraybackslash}p{\dimexpr0.563039840\linewidth-2.815199199\tabcolsep\relax}>{\centering\arraybackslash}p{\dimexpr0.081619264\linewidth-0.408096318\tabcolsep\relax}>{\centering\arraybackslash}p{\dimexpr0.355340897\linewidth-1.776704483\tabcolsep\relax}@{}}
\caption{\rev{Coverage of eleven object types in the benchmark, with rare types combined in the last row. The last column counts correct passage-relation answers for Astra among questions targeting that type.}}\label{tab:objects}\\

\arrayrulecolor{black!70}
\toprule
{\fontsize{8.5}{10.03}\selectfont\bfseries Object type} & {\fontsize{8.5}{10.03}\selectfont\bfseries Clips} & {\fontsize{8.5}{10.03}\selectfont\bfseries \shortstack{Astra passage relation\\correct/total}} \\
\midrule
\endfirsthead
\caption[]{(continued)}\\

\arrayrulecolor{black!70}
\toprule
{\fontsize{8.5}{10.03}\selectfont\bfseries Object type} & {\fontsize{8.5}{10.03}\selectfont\bfseries Clips} & {\fontsize{8.5}{10.03}\selectfont\bfseries \shortstack{Astra passage relation\\correct/total}} \\
\midrule
\endhead
\midrule
\endfoot
\endlastfoot

Cars and other road vehicles & 80 & 65/119 \\
Riders with their vehicles & 64 & 55/122 \\
Pedestrians & 50 & 42/98 \\
Two-wheelers assessed separately & 11 & 9/18 \\
Curbs, steps and ground boundaries & 68 & 49/102 \\
Bollards, posts and cones & 54 & 43/84 \\
Trees, branches and hedges & 47 & 40/69 \\
Street furniture, signs and carts & 26 & 21/41 \\
Gates and fences & 11 & 7/14 \\
Other (pushchairs, rocks and similar obstacles) & 5 & 3/8 \\
\bottomrule
\end{longtable}
\endgroup
\endgroup\endgroup

\endgroup\par\addvspace{8pt}

\par\addvspace{8pt}\begingroup

\noindent\begin{minipage}{\linewidth}\normalfont\normalsize\raggedright
\textbf{Interaction-type coverage.} Table~\ref{tab:interactions} distinguishes the walking encounters from the object categories above. The Clips column counts evaluation clips, while Empty and Nonempty count FFG question windows in the clips containing that encounter. A clip may contain several encounter types; these rows likewise describe overlapping coverage rather than disjoint test splits.
\end{minipage}\par\smallskip

\color{RevisionBlue}
\centering\color{RevisionBlue}

\begingroup\footnotesize
\setlength{\tabcolsep}{3.2pt}\renewcommand{\arraystretch}{1.10}
\begingroup
\EgoTableSetup{9}{2.4}
\begin{longtable}{@{\hspace{\tabcolsep}}>{\raggedright\arraybackslash}p{\dimexpr0.704044658\linewidth-4.928312607\tabcolsep\relax}>{\centering\arraybackslash}p{\dimexpr0.070890230\linewidth-0.496231610\tabcolsep\relax}>{\centering\arraybackslash}p{\dimexpr0.090103108\linewidth-0.630721755\tabcolsep\relax}>{\centering\arraybackslash}p{\dimexpr0.134962004\linewidth-0.944734028\tabcolsep\relax}@{}}
\caption{\rev{Interaction coverage in the 145 \rev{evaluation clips} with a nonempty FFG reference in at least one time window. For each identifiable interaction type, counts include all FFG windows in the clips where it occurs. Categories can overlap.}}\label{tab:interactions}\\

\arrayrulecolor{black!70}
\toprule
{\fontsize{8.5}{10.03}\selectfont\bfseries Interaction} & {\fontsize{8.5}{10.03}\selectfont\bfseries Clips} & {\fontsize{8.5}{10.03}\selectfont\bfseries Empty} & {\fontsize{8.5}{10.03}\selectfont\bfseries Nonempty} \\
\midrule
\endfirsthead
\caption[]{(continued)}\\

\arrayrulecolor{black!70}
\toprule
{\fontsize{8.5}{10.03}\selectfont\bfseries Interaction} & {\fontsize{8.5}{10.03}\selectfont\bfseries Clips} & {\fontsize{8.5}{10.03}\selectfont\bfseries Empty} & {\fontsize{8.5}{10.03}\selectfont\bfseries Nonempty} \\
\midrule
\endhead
\midrule
\endfoot
\endlastfoot

Oncoming approach or passing & 36 & 8 & 97 \\
Crossing paths & 23 & 7 & 62 \\
Same-direction travel or following & 12 & 3 & 31 \\
Passing a stationary person, rider or vehicle & 8 & 1 & 25 \\
Space constrained by fixtures or vegetation & 51 & 8 & 138 \\
Negotiating a curb, step or ground boundary & 32 & 6 & 82 \\
Other (overtaking, stopping, parked two-wheelers, gates) & 10 & 2 & 27 \\
\bottomrule
\end{longtable}
\endgroup\endgroup

\endgroup\par\addvspace{8pt}

Astra scores 60.48\% on 62 nonempty questions from clips with crossing paths and 29.83--30.31\% on 138 from clips with fixtures or vegetation constraining space. The ordering holds across all 14 configurations even at opposing identity bounds. These descriptive comparisons use whole-question discovery.

In a subset of 189 \rev{evaluation clips} examined for co-occurring obstacles, 95 include at least two assessed object types, 45 combine people, riders, or road vehicles with infrastructure, vegetation, or terrain, and 23 contain a question requiring several entities in the same response. Examples include a vehicle crossing a pedestrian crossing, a cyclist overtaking the wearer on a shared path, and a gate wing limiting space while the wearer follows a pedestrian through an entrance. These clips show how people, vehicles and fixed obstacles can constrain the same walk. The counts describe the selected clips.

\FloatBarrier
\subsection{Cross-Task Diagnostic Comparisons}
\label{app:cross_task}
The cross-task discovery analysis uses singleton reference sets with an explicit target correspondence and identical ordered comparison times.

\par\addvspace{8pt}\begingroup

\noindent\begin{minipage}{\linewidth}\normalfont\normalsize\raggedright
\textbf{Motion and passage on the same targets.} Table~\ref{tab:targets} pairs independently answered motion and passage questions about the same named target and ordered times. Each row compares accuracies on its own common targets. The stationary/increasing-constraint row isolates cases in which recognizing stationarity alone does not determine whether passage becomes more constrained.
\end{minipage}\par\smallskip

\color{RevisionBlue}

\color{RevisionBlue}
\centering\color{RevisionBlue}

\begingroup\small
\setlength{\tabcolsep}{3.2pt}\renewcommand{\arraystretch}{1.10}
\begingroup
\EgoTableSetup{9}{2.4}
\begingroup
\EgoTableSetup{9}{2.4}
\begin{longtable}{@{\hspace{\tabcolsep}}>{\raggedright\arraybackslash}p{\dimexpr0.394402011\linewidth-2.760814074\tabcolsep\relax}>{\centering\arraybackslash}p{\dimexpr0.180349379\linewidth-1.262445653\tabcolsep\relax}>{\centering\arraybackslash}p{\dimexpr0.201591839\linewidth-1.411142876\tabcolsep\relax}>{\centering\arraybackslash}p{\dimexpr0.223656771\linewidth-1.565597397\tabcolsep\relax}@{}}
\caption{\rev{Astra on named targets with linked, independently answered questions. Links require identical target descriptions in the same clip and time order, with one unique question in each task. The two accuracies use the same targets in each row. Of the 394 comparisons, Astra answers only the motion question correctly in 107 and only the passage question correctly in nine.}}\label{tab:targets}\\

\arrayrulecolor{black!70}
\toprule
{\fontsize{8.5}{10.03}\selectfont\bfseries Target condition} & {\fontsize{8.5}{10.03}\selectfont\bfseries Comparisons} & {\fontsize{8.5}{10.03}\selectfont\bfseries \shortstack{Object motion\\correct}} & {\fontsize{8.5}{10.03}\selectfont\bfseries \shortstack{Passage relation\\correct}} \\
\midrule
\endfirsthead
\caption[]{(continued)}\\

\arrayrulecolor{black!70}
\toprule
{\fontsize{8.5}{10.03}\selectfont\bfseries Target condition} & {\fontsize{8.5}{10.03}\selectfont\bfseries Comparisons} & {\fontsize{8.5}{10.03}\selectfont\bfseries \shortstack{Object motion\\correct}} & {\fontsize{8.5}{10.03}\selectfont\bfseries \shortstack{Passage relation\\correct}} \\
\midrule
\endhead
\midrule
\endfoot
\endlastfoot

All linked targets & 394 & 360 & 262 \\
Stationary, constraint increases & 67 & 67 & 46 \\
Moving, irrelevant at both times & 84 & 82 & 54 \\
\bottomrule
\end{longtable}
\endgroup
\endgroup\endgroup

\endgroup\par\addvspace{8pt}

\par\addvspace{8pt}\begingroup

\noindent\begin{minipage}{\linewidth}\normalfont\normalsize\raggedright
\textbf{Reference-answer distribution.} Table~\ref{tab:classref} shows the reference classes for the passage-relation task. An entity irrelevant at both times is different from a case whose change cannot be resolved from the permitted evidence. Keeping these labels separate is necessary when interpreting errors on reduced constraints or uncertain observations.
\end{minipage}\par\smallskip

\color{RevisionBlue}
\centering

\begingroup
\EgoTableSetup{9}{2.4}
\begin{longtable}{@{\hspace{\tabcolsep}}>{\raggedright\arraybackslash}p{\dimexpr0.7\linewidth-2.1\tabcolsep\relax}>{\centering\arraybackslash}p{\dimexpr0.3\linewidth-0.9\tabcolsep\relax}@{}}
\caption{\rev{Reference answers for the 675 passage-relation questions. Insufficient visual evidence and irrelevance are distinct reference answers.}}\label{tab:classref}\\

\arrayrulecolor{black!70}
\toprule
\textbf{Passage relation} & \textbf{Questions} \\
\midrule
\endfirsthead
\caption[]{(continued)}\\

\arrayrulecolor{black!70}
\toprule
\textbf{Passage relation} & \textbf{Questions} \\
\midrule
\endhead
\midrule
\endfoot
\endlastfoot

Irrelevant at both times & 268 \\
Constraint increases & 181 \\
Constraint decreases & 181 \\
Insufficient evidence & 45 \\
\midrule
\textbf{Total} & \textbf{675} \\
\bottomrule
\end{longtable}
\endgroup

\endgroup\par\addvspace{8pt}

\FloatBarrier
\subsection{Complete Entity Selection Comparison}
\label{app:selection_full}
\input{figure_selection_full}
\rev{Figures~\ref{fig:selection_full} and~\ref{fig:outcomes} use all 574 FFG questions (211 empty and 363 nonempty reference sets), containing 431 reference entities. All 14 visual configurations use this same set. The family averages in Figure~\ref{fig:selection} give equal weight to the two frame-based models in each family; individual configurations are shown here. Discovery, Labels and Exact are three metrics for this single FFG task.}

\par\addvspace{8pt}\begingroup

\noindent\begin{minipage}{\linewidth}\normalfont\normalsize\raggedright
\textbf{Selection errors for individual configurations.} Table~\ref{tab:selection_precision_recall} supplies the counts underlying the selection plots. All rows use the same reference entities; TP and FN describe coverage, while FP records additional reports. The precision and recall ranges retain identity ambiguity rather than selecting a favorable match assignment.
\end{minipage}\par\smallskip

\color{RevisionBlue}
\centering\color{RevisionBlue}

\begingroup
\EgoTableSetup{9}{2.4}
\begin{longtable}{@{\hspace{\tabcolsep}}>{\raggedright\arraybackslash}p{\dimexpr0.246684027\linewidth-2.713524295\tabcolsep\relax}>{\centering\arraybackslash}p{\dimexpr0.125983492\linewidth-1.385818410\tabcolsep\relax}>{\centering\arraybackslash}p{\dimexpr0.125983492\linewidth-1.385818410\tabcolsep\relax}>{\centering\arraybackslash}p{\dimexpr0.125983492\linewidth-1.385818410\tabcolsep\relax}>{\centering\arraybackslash}p{\dimexpr0.200725515\linewidth-2.207980668\tabcolsep\relax}>{\centering\arraybackslash}p{\dimexpr0.174639982\linewidth-1.921039807\tabcolsep\relax}@{}}
\caption{\rev{Entity discovery over all 574 FFG questions and 431 reference entities. TP counts matched entities, FP extra reports and FN missed references. Micro-precision is $100\,\mathrm{TP}/(\mathrm{TP}+\mathrm{FP})$ and recall is $100\,\mathrm{TP}/431$. Ranges reflect feasible identity matches. All 14 configurations use the same reference set.}}\label{tab:selection_precision_recall}\\

\arrayrulecolor{black!70}
\toprule
{\fontsize{8.5}{10.03}\selectfont\bfseries Model} & {\fontsize{8.5}{10.03}\selectfont\bfseries TP} & {\fontsize{8.5}{10.03}\selectfont\bfseries FP} & {\fontsize{8.5}{10.03}\selectfont\bfseries FN} & {\fontsize{8.5}{10.03}\selectfont\bfseries Precision (\%)} & {\fontsize{8.5}{10.03}\selectfont\bfseries Recall (\%)} \\
\midrule
\endfirsthead
\caption[]{(continued)}\\

\arrayrulecolor{black!70}
\toprule
{\fontsize{8.5}{10.03}\selectfont\bfseries Model} & {\fontsize{8.5}{10.03}\selectfont\bfseries TP} & {\fontsize{8.5}{10.03}\selectfont\bfseries FP} & {\fontsize{8.5}{10.03}\selectfont\bfseries FN} & {\fontsize{8.5}{10.03}\selectfont\bfseries Precision (\%)} & {\fontsize{8.5}{10.03}\selectfont\bfseries Recall (\%)} \\
\midrule
\endhead
\midrule
\endfoot
\endlastfoot

\rowcolor{EgoTableFrames}\multicolumn{6}{l}{\textbf{Chronological frames}} \\
GPT-6 Astra & 278--279 & 224--225 & 152--153 & 55.27--55.47 & 64.50--64.73 \\
GPT-5.6 Sol & 278--279 & 332--333 & 152--153 & 45.50--45.66 & 64.50--64.73 \\
Gemini 3.8 Flash & 184--190 & 132--138 & 241--247 & 57.14--59.01 & 42.69--44.08 \\
Gemini 3.1 Pro & 224--239 & 251--266 & 192--207 & 45.71--48.78 & 51.97--55.45 \\
Llama 4 Maverick & 130--157 & 305--332 & 274--301 & 28.14--33.98 & 30.16--36.43 \\
Llama 4 Scout & 123--143 & 198--218 & 288--308 & 36.07--41.94 & 28.54--33.18 \\
Qwen3.8 Max & 258--263 & 319--324 & 168--173 & 44.33--45.19 & 59.86--61.02 \\
Qwen3.8 Flash & 235--247 & 336--348 & 184--196 & 40.31--42.37 & 54.52--57.31 \\
\midrule
\rowcolor{EgoTableVideo}\multicolumn{6}{l}{\textbf{Native video}} \\
Qwen3.8 Max & 193--201 & 256--264 & 230--238 & 42.23--43.98 & 44.78--46.64 \\
GLM-5.3 Flash & 139--149 & 185--195 & 282--292 & 41.62--44.61 & 32.25--34.57 \\
MiniMax M3 & 101--114 & 226--239 & 317--330 & 29.71--33.53 & 23.43--26.45 \\
Gemma 4 31B & 95--106 & 136--147 & 325--336 & 39.26--43.80 & 22.04--24.59 \\
Qwen3.5-397B & 61--66 & 253--258 & 365--370 & 19.12--20.69 & 14.15--15.31 \\
Qwen3.8 Flash & 76--88 & 132--144 & 343--355 & 34.55--40.00 & 17.63--20.42 \\
\bottomrule
\end{longtable}
\endgroup

\endgroup\par\addvspace{8pt}

\rev{Figure~\ref{fig:outcomes} separates the 431 reference entities into correctly labeled matches, incorrectly labeled matches and omissions, retaining an unresolved component where identity affects the allocation.}
\input{figure_outcomes}

\FloatBarrier
\subsection{Recognizing Insufficient Evidence}
\label{app:uncertainty}
\rev{Table~\ref{tab:uncertainty} reports the full-benchmark diagnostic discussed in Section~\ref{sec:uncertainty}. All 72 questions have an insufficient-evidence reference: 27 concern wearer translation and 45 concern passage relations. The same planned denominators apply to every configuration.}
\par\addvspace{8pt}\begingroup

\noindent\begin{minipage}{\linewidth}\normalfont\normalsize\raggedright
\textbf{Insufficient-evidence diagnostic.} Table~\ref{tab:uncertainty} separates the wearer-translation and passage-relation questions whose references explicitly require insufficient evidence. Every configuration is evaluated against the same planned denominators, with unavailable responses retained. This diagnostic tests whether a model withholds an unsupported definite judgment, rather than whether it identifies an irrelevant entity.
\end{minipage}\par\smallskip

\color{RevisionBlue}
\centering\color{RevisionBlue}

\begingroup
\EgoTableSetup{9}{2.4}
\begin{longtable}{@{\hspace{\tabcolsep}}>{\raggedright\arraybackslash}p{\dimexpr0.392497131\linewidth-2.747479916\tabcolsep\relax}>{\centering\arraybackslash}p{\dimexpr0.206969159\linewidth-1.448784115\tabcolsep\relax}>{\centering\arraybackslash}p{\dimexpr0.209747568\linewidth-1.468232975\tabcolsep\relax}>{\centering\arraybackslash}p{\dimexpr0.190786142\linewidth-1.335502993\tabcolsep\relax}@{}}
\caption{\rev{Recognizing insufficient evidence on all 72 such questions in the 2,201-question benchmark: 27 wearer-translation and 45 passage-relation questions. Entries are correct/planned counts; unavailable responses remain in the denominator.}}\label{tab:uncertainty}\\

\arrayrulecolor{black!70}
\toprule
{\fontsize{8.5}{10.03}\selectfont\bfseries Model} & {\fontsize{8.5}{10.03}\selectfont\bfseries \shortstack{Wearer\\translation}} & {\fontsize{8.5}{10.03}\selectfont\bfseries \shortstack{Passage\\relation}} & {\fontsize{8.5}{10.03}\selectfont\bfseries Overall} \\
\midrule
\endfirsthead
\caption[]{(continued)}\\

\arrayrulecolor{black!70}
\toprule
{\fontsize{8.5}{10.03}\selectfont\bfseries Model} & {\fontsize{8.5}{10.03}\selectfont\bfseries \shortstack{Wearer\\translation}} & {\fontsize{8.5}{10.03}\selectfont\bfseries \shortstack{Passage\\relation}} & {\fontsize{8.5}{10.03}\selectfont\bfseries Overall} \\
\midrule
\endhead
\midrule
\endfoot
\endlastfoot

\rowcolor{EgoTableFrames}\multicolumn{4}{l}{\textbf{Chronological frames}} \\
GPT-6 Astra & 1/27 & 2/45 & 3/72 \\
GPT-5.6 Sol & 0/27 & 0/45 & 0/72 \\
Gemini 3.8 Flash & 0/27 & 0/45 & 0/72 \\
Gemini 3.1 Pro & 0/27 & 0/45 & 0/72 \\
Llama 4 Maverick & 0/27 & 1/45 & 1/72 \\
Llama 4 Scout & 0/27 & 0/45 & 0/72 \\
Qwen3.8 Max & 0/27 & 0/45 & 0/72 \\
Qwen3.8 Flash & 0/27 & 8/45 & 8/72 \\
\midrule
\rowcolor{EgoTableVideo}\multicolumn{4}{l}{\textbf{Native video}} \\
Qwen3.8 Max & 1/27 & 0/45 & 1/72 \\
GLM-5.3 Flash & 0/27 & 0/45 & 0/72 \\
MiniMax M3 & 0/27 & 1/45 & 1/72 \\
Gemma 4 31B & 0/27 & 0/45 & 0/72 \\
Qwen3.5-397B & 0/27 & 0/45 & 0/72 \\
Qwen3.8 Flash & 0/27 & 0/45 & 0/72 \\
\bottomrule
\end{longtable}
\endgroup

\endgroup\par\addvspace{8pt}

\FloatBarrier
\subsection{False-positive Penalty Sensitivity}
\label{app:penalty_sensitivity}
We varied the penalty for extra entities on nonempty questions by replacing $\mathrm{FP}$ with $\lambda\,\mathrm{FP}$ in Eq.~\ref{eq:object_score}. Averaging over all 574 questions, Astra's Discovery scores are 60.02--60.10\%, 55.69--55.81\%, and 55.14--55.31\% for $\lambda=0.5,1,2$. The empty-set rule is held fixed. At every weight the full-set score is above the 36.76\% always-empty baseline. Precision and recall in Table~\ref{tab:prfull} do not use this penalty.

\FloatBarrier
\subsection{Reading the benchmark comparison}\label{app:nearby}
Table~\ref{tab:benchmark_comparison} codes an explicit test when its task definition requires the indicated output, and a related test when a component is addressed. RiskBench uses simulated traffic scenarios with specified risk actors and evaluates localization, anticipation and driving decisions~\revcite{10610270}. DRAMA contains 17,785 driving clips with risk boxes, object-level questions and captions about a driver's situation~\revcite{Malla_2023_WACV}. SAW-Bench uses 786 real videos and 2,071 human-annotated questions, including spatial affordance and action feasibility~\revcite{li2026sawbench}. These tasks ask where a hazardous actor is, why it warrants attention, or which action is feasible. In the yielding example (Figure~\ref{fig:approaching}), EgoPass measures a specific quantity, the change in the pedestrian's physical constraint on the wearer's path over two times. The independent FFG response must discover the relevant set before labeling each change. The 21 misjudged passage changes in Table~\ref{tab:targets} show why recognizing a stationary object is insufficient to judge whether it increasingly obstructs the walker's path.

}

\clearpage
\section{Evaluation Prompts}
\label{app:evaluation_prompts}
This appendix provides the English prompts used for the five evaluation tasks. \rev{The templates below show the frame-based condition. Each question is asked independently with timestamped frame grids; the native-video configurations receive the corresponding clip.} Exactly one context condition from Section~\ref{app:prompt_context} is inserted at \texttt{\{history\_instruction\}} in the corresponding task template. Other braced fields are replaced with the video and clip identifiers, frame timestamps, queried times, question text, and answer options. Object descriptions are supplied only for tasks with a specified target; entity discovery instead uses a neutral identifier for the queried time pair. \rev{The tasks and scoring definitions are shared across input formats. Appendix~\ref{app:scoring_methods} describes the handling of format corrections.} Responses are required in JavaScript Object Notation (JSON).

\newenvironment{EvaluationPrompt}{%
  \par\Needspace{4\baselineskip}\smallskip\begingroup
  \begin{list}{}{\setlength{\leftmargin}{7pt}\setlength{\rightmargin}{7pt}%
    \setlength{\topsep}{3pt}\setlength{\parsep}{1pt}\setlength{\itemsep}{0pt}}
  \item[]\fontsize{9}{10.6}\selectfont\sffamily\raggedright
  \setlength{\parindent}{0pt}\setlength{\parskip}{1pt}%
  \noindent\rule{\linewidth}{0.4pt}\par\nobreak\smallskip
}{%
  \par\nobreak\smallskip\noindent\rule{\linewidth}{0.4pt}%
  \end{list}\endgroup\smallskip
}
\newcommand{\PromptHeading}[1]{\par\addvspace{3pt}\noindent\textbf{#1}\par\nobreak}

\subsection{Context Conditions}
\label{app:prompt_context}
The condition instructions are listed once here and inserted into every independent request. No History uses only the current clip. History also includes the immediately preceding clip from the same video. Both conditions require an answer about the current question.

\begin{EvaluationPrompt}
\textbf{No History}\par\nobreak\smallskip
\PromptHeading{[NO-HISTORY CONDITION]}
\noindent Use only the current slice images and the current question. Do not assume access to earlier slices, earlier answers, or events outside the supplied evidence.\par
\end{EvaluationPrompt}

\begin{EvaluationPrompt}
\textbf{History}\par\nobreak\smallskip
\PromptHeading{[PREVIOUS-SLICE VISUAL-HISTORY CONDITION]}
\noindent For current slice S\_\allowbreak{}k, the input contains only the frames of the immediately preceding slice S\_\allowbreak{}(k-1) of the same video and the current slice S\_\allowbreak{}k. No frames from S\_\allowbreak{}(k-2) or earlier slices are provided. For the first slice, use only its current-slice frames.\par
\noindent No previous questions, model answers, human answers, or textual summaries of earlier events are provided. Treat this question as an independent request, not a continuing conversation.\par
\noindent Use the frames' absolute timestamps to determine event order and associate the same object across these two slices. Visual evidence from the immediately preceding slice may help infer current identity, motion state, and trajectory, and distinguish object motion from wearer motion.\par
\noindent Do not assume an earlier state persists automatically. Integrate the previous and current frames and update the inference when current evidence supports a state change. If occlusion, a temporal gap, or uncertain identity prevents a reliable inference, retain uncertainty rather than inventing continuity.\par
\noindent Answer only the current question for its specified interval or timestamp pair. Relevance in the previous slice alone does not establish relevance at a queried endpoint. Use only the supplied frames; do not assume access to older slices or unprovided later images.\par
\end{EvaluationPrompt}

\subsection{Wearer Translation}

\label{app:prompt_wearer_translation}

\begin{EvaluationPrompt}
\PromptHeading{[LABEL DEFINITIONS]}
\noindent \textbf{STATIONARY:} There is sufficient visual evidence that the wearer undergoes no discernible positional translation in the physical world during the specified interval.\par
\noindent \textbf{MOVING:} The wearer undergoes confirmable positional translation at any time in the specified interval, including starting, stopping, or changing speed.\par
\noindent \textbf{UNCERTAIN:} The available frames are insufficient to reliably distinguish STATIONARY from MOVING, or the visual evidence is contradictory.\par
\par\addvspace{2pt}
\PromptHeading{[CURRENT INPUT]}
\noindent Video: \{video\_\allowbreak{}id\}\par
\noindent Current slice: \{slice\_\allowbreak{}id\}\par
\noindent You will see contact sheets of first-person video frames. Each frame carries its absolute timestamp in the original video.\par
\noindent Current-slice frame timestamps: [\{allowed\_\allowbreak{}times\_\allowbreak{}s\}].\par
\noindent Available current-slice visual evidence spans \{first\_\allowbreak{}frame\_\allowbreak{}time\_\allowbreak{}s\}s to \{last\_\allowbreak{}frame\_\allowbreak{}time\_\allowbreak{}s\}s in the original video.\par
\noindent Within each contact sheet, frames are ordered left to right, then top to bottom. Contact sheets are ordered from earlier slices to later slices; the final sheet is the current slice.\par
\par\addvspace{2pt}
\noindent \{history\_\allowbreak{}instruction\}\par
\par\addvspace{2pt}
\PromptHeading{[TASK]}
\noindent Determine whether the wearer undergoes real positional translation from \{t1\_\allowbreak{}s\}s to \{t2\_\allowbreak{}s\}s.\par
\noindent 1. Global image translation, rotation, or blur may be caused by head or camera rotation and does not by itself prove wearer translation.\par
\noindent 2. Use ground texture, nearby fixed structures, perspective-scale change, parallax, and cross-frame relative positions to determine real translation.\par
\noindent 3. Do not mistake the motion of other people or vehicles for wearer motion.\par
\noindent 4. Judge only whether real translation occurs. Do not judge direction, speed trend, turn type, or risk.\par
\noindent 5. If evidence is insufficient, select UNCERTAIN. Do not guess that the wearer starts moving based on scene conventions.\par
\par\addvspace{2pt}
\PromptHeading{[QUESTION]}
\noindent \{question\_\allowbreak{}text\}\par
\par\addvspace{2pt}
\noindent \{options\_\allowbreak{}block\}\par
\par\addvspace{2pt}
\noindent Return only JSON: \texttt{\{"answer":"one of \{option\_\allowbreak{}letters\}"\}}\par
\par\addvspace{2pt}
\PromptHeading{[EVALUATION BOUNDARY]}
\noindent Answer only from the attached images, permitted history, and this prompt. Do not use tools, access local files, browse the web, or request external information. Do not infer reference answers from question wording, option order, or question identifiers. Return only the specified JSON, without explanations, Markdown, or additional fields.\par
\end{EvaluationPrompt}

\Needspace{7\baselineskip}\subsection{Viewpoint and Path Turning}

\label{app:prompt_viewpoint_path_turning}

\begin{EvaluationPrompt}
\PromptHeading{[EVENT COMPONENT DEFINITIONS]}
\noindent \textbf{NONE:} Sufficient visual evidence supports no discernible gaze-direction change and no actual travel-path turn during the specified interval.\par
\noindent \textbf{HEAD\_\allowbreak{}TURN\_\allowbreak{}LEFT / HEAD\_\allowbreak{}TURN\_\allowbreak{}RIGHT:} Sufficient visual evidence supports a leftward or rightward head/camera orientation change while the wearer remains stationary or continues along a translational path without a corresponding turn.\par
\noindent \textbf{PATH\_\allowbreak{}TURN\_\allowbreak{}LEFT / PATH\_\allowbreak{}TURN\_\allowbreak{}RIGHT:} Sufficient visual evidence supports real wearer translation and a travel path that curves left or right in the physical world.\par
\noindent \textbf{UNCERTAIN:} The turning source, direction, or temporal order cannot be reliably determined, or visual evidence is contradictory.\par
\par\addvspace{2pt}
\PromptHeading{[CURRENT INPUT]}
\noindent Video: \{video\_\allowbreak{}id\}\par
\noindent Current slice: \{slice\_\allowbreak{}id\}\par
\noindent You will see contact sheets of first-person video frames. Each frame carries its absolute timestamp in the original video.\par
\noindent Current-slice frame timestamps: [\{allowed\_\allowbreak{}times\_\allowbreak{}s\}].\par
\noindent Available current-slice visual evidence spans \{first\_\allowbreak{}frame\_\allowbreak{}time\_\allowbreak{}s\}s to \{last\_\allowbreak{}frame\_\allowbreak{}time\_\allowbreak{}s\}s in the original video.\par
\noindent Within each contact sheet, frames are ordered left to right, then top to bottom. Contact sheets are ordered from earlier slices to later slices; the final sheet is the current slice.\par
\par\addvspace{2pt}
\noindent \{history\_\allowbreak{}instruction\}\par
\par\addvspace{2pt}
\PromptHeading{[OPTION INTERPRETATION]}
\noindent Each event option lists all discernible turning events within the queried interval. Unlisted periods in that option indicate no discernible head/camera rotation or travel-path turn, not missing information.\par
\noindent This is the hypothesis asserted by the option, not visual evidence: verify it against the permitted images. Do not infer an absence of turning merely because an option omits it. Select UNCERTAIN when the visual evidence cannot reliably determine the turning source, direction, or temporal order.\par
\noindent No discernible turning does not mean the wearer is stationary; the wearer may continue moving along an unchanged path.\par
\par\addvspace{2pt}
\PromptHeading{[TASK]}
\noindent Select the event hypothesis that most accurately describes the source, direction, and temporal order of turning from \{t1\_\allowbreak{}s\}s to \{t2\_\allowbreak{}s\}s.\par
\noindent 1. Interpret left and right from the wearer's perspective.\par
\noindent 2. Image rotation alone does not establish PATH\_\allowbreak{}TURN. A path turn requires evidence of translation and changing travel direction, such as ground trajectory, nearby-structure parallax, and sustained motion.\par
\noindent 3. HEAD\_\allowbreak{}TURN also requires positive evidence. Failure to establish PATH\_\allowbreak{}TURN does not automatically establish HEAD\_\allowbreak{}TURN.\par
\noindent 4. Select UNCERTAIN when head/camera rotation cannot be reliably distinguished from path turning.\par
\noindent 5. Use the supplied timestamps to judge direction and temporal order. Do not invent unprovided timestamps. If several events are identifiable, choose the option matching their composition and order.\par
\noindent 6. NONE means evidence supports no turn; UNCERTAIN means evidence is insufficient for a reliable determination. Do not interchange them.\par
\noindent 7. Judge only turning, not speed trend, external-object motion, or risk.\par
\par\addvspace{2pt}
\PromptHeading{[QUESTION]}
\noindent \{question\_\allowbreak{}text\}\par
\par\addvspace{2pt}
\noindent \{options\_\allowbreak{}block\}\par
\par\addvspace{2pt}
\noindent Return only JSON: \texttt{\{"answer":"one of \{option\_\allowbreak{}letters\}"\}}\par
\par\addvspace{2pt}
\PromptHeading{[EVALUATION BOUNDARY]}
\noindent Answer only from the attached images, permitted history, and this prompt. Do not use tools, access local files, browse the web, or request external information. Do not infer reference answers from question wording, option order, or question identifiers. Return only the specified JSON, without explanations, Markdown, or additional fields.\par
\end{EvaluationPrompt}

\subsection{External Object Motion}

\label{app:prompt_external_object_motion}

\begin{EvaluationPrompt}
\PromptHeading{[LABEL DEFINITIONS]}
\noindent \textbf{STATIONARY:} There is sufficient visual evidence that the specified object undergoes no discernible positional translation in the physical world during the interval. A stationary person, parked vehicle, or fixed structure may be STATIONARY.\par
\noindent \textbf{MOVING:} Sufficient evidence supports that the object remains in translational motion over the interval, without a confirmable transition between stationary and moving. Turning, acceleration, or deceleration without such a transition remains MOVING.\par
\noindent \textbf{MOTION\_\allowbreak{}STATE\_\allowbreak{}CHANGED:} The same object shows a confirmable transition from stationary to moving or from moving to stationary within the interval, including one or more starts or stops. A change of direction or speed alone does not qualify.\par
\noindent \textbf{UNCERTAIN:} Evidence is insufficient to reliably determine the object's real motion state or maintain its identity.\par
\par\addvspace{2pt}
\PromptHeading{[CURRENT INPUT]}
\noindent Video: \{video\_\allowbreak{}id\}\par
\noindent Current slice: \{slice\_\allowbreak{}id\}\par
\noindent You will see contact sheets of first-person video frames. Each frame carries its absolute timestamp in the original video.\par
\noindent Current-slice frame timestamps: [\{allowed\_\allowbreak{}times\_\allowbreak{}s\}].\par
\noindent Available current-slice visual evidence spans \{first\_\allowbreak{}frame\_\allowbreak{}time\_\allowbreak{}s\}s to \{last\_\allowbreak{}frame\_\allowbreak{}time\_\allowbreak{}s\}s in the original video.\par
\noindent Within each contact sheet, frames are ordered left to right, then top to bottom. Contact sheets are ordered from earlier slices to later slices; the final sheet is the current slice.\par
\par\addvspace{2pt}
\noindent \{history\_\allowbreak{}instruction\}\par
\par\addvspace{2pt}
\noindent Specified object: \{target\_\allowbreak{}object\}\par
\noindent The description identifies the target only; it is not a motion-state or risk conclusion.\par
\par\addvspace{2pt}
\PromptHeading{[TASK]}
\noindent Determine the observed motion state of this object itself from \{t1\_\allowbreak{}s\}s to \{t2\_\allowbreak{}s\}s.\par
\noindent 1. Distinguish real object motion from wearer translation and head or camera rotation. Use fixed backgrounds, ground references, and cross-frame parallax.\par
\noindent 2. Image-plane motion alone does not establish physical motion. A lack of visible motion alone is not sufficient evidence of STATIONARY.\par
\noindent 3. Treat a rider and the ridden vehicle as one motion unit while they move together.\par
\noindent 4. Judge one specified entity or rider-vehicle unit, not an aggregate of independent entities. Inspect the full queried interval: a confirmed start or stop takes precedence over a single endpoint state. If the interval contains a confirmed stationary-to-moving or moving-to-stationary transition, select MOTION\_\allowbreak{}STATE\_\allowbreak{}CHANGED even when both endpoints have the same state.\par
\noindent 5. A fixed structure being STATIONARY does not imply it is safe. Do not judge attention priority or future risk in this task.\par
\noindent 6. If identity or motion cannot be reliably determined, select UNCERTAIN instead of guessing.\par
\par\addvspace{2pt}
\PromptHeading{[QUESTION]}
\noindent \{question\_\allowbreak{}text\}\par
\par\addvspace{2pt}
\noindent \{options\_\allowbreak{}block\}\par
\par\addvspace{2pt}
\noindent Return only JSON: \texttt{\{"answer":"one of \{option\_\allowbreak{}letters\}"\}}\par
\par\addvspace{2pt}
\PromptHeading{[EVALUATION BOUNDARY]}
\noindent Answer only from the attached images, permitted history, and this prompt. Do not use tools, access local files, browse the web, or request external information. Do not infer reference answers from question wording, option order, or question identifiers. Return only the specified JSON, without explanations, Markdown, or additional fields.\par
\end{EvaluationPrompt}

\subsection{Passage Relations for Specified Objects}

\label{app:prompt_passage_relations}

\begin{EvaluationPrompt}
\PromptHeading{[TASK AND INPUT]}
\noindent Compare the attention required by external objects at \{t2\_\allowbreak{}s\}s with \{t1\_\allowbreak{}s\}s. Attention means monitoring, preparation, or action adjustment warranted by a supported physical conflict or safe-passage constraint, not visibility or general interest.\par
\noindent Video: \{video\_\allowbreak{}id\}. Current slice: \{slice\_\allowbreak{}id\}.\par
\noindent Current-slice timestamps: [\{allowed\_\allowbreak{}times\_\allowbreak{}s\}]. Visual evidence spans \{first\_\allowbreak{}frame\_\allowbreak{}time\_\allowbreak{}s\}s to \{last\_\allowbreak{}frame\_\allowbreak{}time\_\allowbreak{}s\}s.\par
\noindent Frames within each contact sheet run left to right, then top to bottom; sheets run from earlier to later slices, with the current slice last.\par
\noindent \{history\_\allowbreak{}instruction\}\par
\par\addvspace{2pt}
\PromptHeading{[EVIDENCE AND MOTION]}
\noindent Use all condition-permitted frames to associate identities and estimate each endpoint state. This is a retrospective endpoint comparison, not future-hidden online forecasting.\par
\noindent At each endpoint, separately project the next 2 seconds using the wearer’s and object's motion trends supported at that endpoint. Later stopping, turning, successful avoidance, or absence of collision must not replace the earlier trend assessment.\par
\noindent Infer wearer translation, direction, speed trend, and actual path turning before assessing relative motion. Head/camera rotation alone does not establish a path turn. When evidence supports no discernible wearer translation, keep the occupied position fixed; do not assume a later start. Do not carry earlier speed forward after stopping.\par
\noindent Do not invent sudden movement, cooperative avoidance, unseen trajectories, or precise distances, speeds, or probabilities. Occlusion or leaving view alone proves neither safety nor reduced attention. If unresolved motion or identity affects the comparison, retain uncertainty.\par
\par\addvspace{2pt}
\PromptHeading{[STEP 1: DECIDE WHETHER AN OBJECT QUALIFIES]}
\noindent Include an object only when concrete visual evidence at at least one queried endpoint supports a reasonable possibility of physical conflict or insufficient safe-passage space under the wearer’s and object's motion trends at that endpoint. The conflict or constraint must be relevant within the next 2 seconds, or an adjustment to avoid it must need to begin within those 2 seconds even if contact would occur later.\par
\noindent Assess relative position, trajectory, occupied width, timing, and available passage space. A collision need not be certain, and avoidance need not already have begun.\par
\noindent Determine qualification before assigning a change label. Do not select objects merely because they are visible, moving, approaching, nearby, or visually prominent. If the specified object clearly fails to qualify at both endpoints, select NOT\_\allowbreak{}RELEVANT. Do not report a change label to justify including an object that fails the gate.\par
\noindent Normal passage along already separated paths with adequate space is not an object-induced need to adjust. Paths crossing at clearly different times do not alone establish conflict. Parallel motion alone proves neither conflict nor safety. Inability to rule out hypothetical danger is not positive inclusion evidence.\par
\noindent Apply this gate equally to moving objects and static structures. Treat each rider and their ridden vehicle as one occupied unit with nonzero width. Do not infer safety or relevance from traffic rules, right of way, or signal colors; do not list traffic signals.\par
\par\addvspace{2pt}
\PromptHeading{[STEP 2: COMPARE QUALIFYING OBJECTS]}
\noindent Compare the overall evidence for conflict likelihood, urgency, passage constraints, and necessary action adjustment. Do not compare distance or image position alone, and do not use the slice maximum instead of either endpoint.\par
\noindent \textbf{HIGHER:} Supported attention requirement is greater at the second endpoint.\par
\noindent \textbf{LOWER:} Supported attention requirement is smaller at the second endpoint.\par
\noindent \textbf{INSUFFICIENT\_\allowbreak{}EVIDENCE:} Relevant endpoint states or their comparison cannot be resolved reliably, including conflicting cues without a supported overall direction.\par
\noindent When both endpoints are assessable, clearly nonqualifying to qualifying means HIGHER; the reverse means LOWER. Uncertainty is not proof of nonqualification.\par
\par\addvspace{2pt}
\PromptHeading{[JUDGE THE SPECIFIED OBJECT]}
\noindent Object: \{target\_\allowbreak{}object\}\par
\noindent This name identifies the object; it does not imply relevance.\par
\noindent 1. If evidence supports nonqualification at both endpoints, select NOT\_\allowbreak{}RELEVANT.\par
\noindent 2. If relevance cannot be determined reliably, or the required comparison is unresolved, select INSUFFICIENT\_\allowbreak{}EVIDENCE.\par
\noindent 3. Otherwise, at least one endpoint qualifies; select HIGHER or LOWER using Step 2.\par
\par\addvspace{2pt}
\noindent Question: \{question\_\allowbreak{}text\}\par
\noindent \{options\_\allowbreak{}block\}\par
\par\addvspace{2pt}
\noindent Return only JSON: \texttt{\{"answer":"one of \{option\_\allowbreak{}letters\}"\}}\par
\par\addvspace{2pt}
\PromptHeading{[RESPONSE BOUNDARY]}
\noindent Use only the attached images, permitted history, and this prompt. Do not use tools or external information, or infer reference answers from wording, identifiers, or option order. Return only the specified JSON, using English for any free-text field, without explanations, absolute attention levels, rankings, timelines, or separate predictions.\par
\end{EvaluationPrompt}

\subsection{Autonomous Relevant Entity Discovery}

\label{app:prompt_entity_discovery}

\begin{EvaluationPrompt}
\PromptHeading{[TASK AND INPUT]}
\noindent Compare the attention required by external objects at \{t2\_\allowbreak{}s\}s with \{t1\_\allowbreak{}s\}s. Attention means monitoring, preparation, or action adjustment warranted by a supported physical conflict or safe-passage constraint, not visibility or general interest.\par
\noindent Video: \{video\_\allowbreak{}id\}. Current slice: \{slice\_\allowbreak{}id\}.\par
\noindent Current-slice timestamps: [\{allowed\_\allowbreak{}times\_\allowbreak{}s\}]. Visual evidence spans \{first\_\allowbreak{}frame\_\allowbreak{}time\_\allowbreak{}s\}s to \{last\_\allowbreak{}frame\_\allowbreak{}time\_\allowbreak{}s\}s.\par
\noindent Frames within each contact sheet run left to right, then top to bottom; sheets run from earlier to later slices, with the current slice last.\par
\noindent \{history\_\allowbreak{}instruction\}\par
\par\addvspace{2pt}
\PromptHeading{[EVIDENCE AND MOTION]}
\noindent Use all condition-permitted frames to associate identities and estimate each endpoint state. This is a retrospective endpoint comparison, not future-hidden online forecasting.\par
\noindent At each endpoint, separately project the next 2 seconds using the wearer’s and object's motion trends supported at that endpoint. Later stopping, turning, successful avoidance, or absence of collision must not replace the earlier trend assessment.\par
\noindent Infer wearer translation, direction, speed trend, and actual path turning before assessing relative motion. Head/camera rotation alone does not establish a path turn. When evidence supports no discernible wearer translation, keep the occupied position fixed; do not assume a later start. Do not carry earlier speed forward after stopping.\par
\noindent Do not invent sudden movement, cooperative avoidance, unseen trajectories, or precise distances, speeds, or probabilities. Occlusion or leaving view alone proves neither safety nor reduced attention. If unresolved motion or identity affects the comparison, retain uncertainty.\par
\par\addvspace{2pt}
\PromptHeading{[STEP 1: DECIDE WHETHER AN OBJECT QUALIFIES]}
\noindent Include an object only when concrete visual evidence at at least one queried endpoint supports a reasonable possibility of physical conflict or insufficient safe-passage space under the wearer’s and object's motion trends at that endpoint. The conflict or constraint must be relevant within the next 2 seconds, or an adjustment to avoid it must need to begin within those 2 seconds even if contact would occur later.\par
\noindent Assess relative position, trajectory, occupied width, timing, and available passage space. A collision need not be certain, and avoidance need not already have begun.\par
\noindent Determine qualification before assigning a change label. Do not select objects merely because they are visible, moving, approaching, nearby, or visually prominent. Do not include objects that fail to qualify. If no object qualifies, return an empty objects array. Do not report a change label to justify including an object that fails the gate.\par
\noindent Normal passage along already separated paths with adequate space is not an object-induced need to adjust. Paths crossing at clearly different times do not alone establish conflict. Parallel motion alone proves neither conflict nor safety. Inability to rule out hypothetical danger is not positive inclusion evidence.\par
\noindent Apply this gate equally to moving objects and static structures. Treat each rider and their ridden vehicle as one occupied unit with nonzero width. Do not infer safety or relevance from traffic rules, right of way, or signal colors; do not list traffic signals.\par
\par\addvspace{2pt}
\PromptHeading{[STEP 2: COMPARE QUALIFYING OBJECTS]}
\noindent Compare the overall evidence for conflict likelihood, urgency, passage constraints, and necessary action adjustment. Do not compare distance or image position alone, and do not use the slice maximum instead of either endpoint.\par
\noindent \textbf{HIGHER:} Supported attention requirement is greater at the second endpoint.\par
\noindent \textbf{LOWER:} Supported attention requirement is smaller at the second endpoint.\par
\noindent \textbf{INSUFFICIENT\_\allowbreak{}EVIDENCE:} Relevant endpoint states or their comparison cannot be resolved reliably, including conflicting cues without a supported overall direction.\par
\noindent When both endpoints are assessable, clearly nonqualifying to qualifying means HIGHER; the reverse means LOWER. Uncertainty is not proof of nonqualification.\par
\par\addvspace{2pt}
\PromptHeading{[DISCOVER QUALIFYING OBJECTS WITHOUT TARGET HINTS]}
\noindent Independently identify every object with positive qualification evidence at at least one endpoint. Do not report an object solely because relevance is uncertain at both endpoints.\par
\noindent For each included object, apply Step 2. If qualification is established at one endpoint but the other endpoint or cross-time identity is unresolved, use INSUFFICIENT\_\allowbreak{}EVIDENCE. Omit objects nonqualifying at both endpoints; do not emit NOT\_\allowbreak{}RELEVANT.\par
\noindent Use neutral, localizable descriptions. Report each physical object once; do not merge unrelated people or vehicles. There is no fixed or hidden target count.\par
\noindent If no object has positive qualification evidence, return objects: []. This means no qualifying object was established, not proof of absolute safety.\par
\par\addvspace{2pt}
\PromptHeading{[OUTPUT]}
\noindent Return one comparison with these exact fields. Replace illustrative values; select one attention\_\allowbreak{}change label per object:\par
\begingroup\ttfamily\fontsize{9}{10.5}\selectfont
\noindent\{"comparisons": [\{"pair\_\allowbreak{}id": "\{pair\_\allowbreak{}id\}",\par
\noindent\hspace*{0.9em}"t1": "\{t1\_\allowbreak{}s\}s", "t2": "\{t2\_\allowbreak{}s\}s", "objects": [\{\par
\noindent\hspace*{0.9em}"object": "neutral, localizable description",\par
\noindent\hspace*{0.9em}"attention\_\allowbreak{}change": "HIGHER | LOWER | INSUFFICIENT\_\allowbreak{}EVIDENCE"\}]\}]\}\par
\endgroup
\noindent For an empty result, retain the comparison fields and set only objects to [].\par
\par\addvspace{2pt}
\PromptHeading{[RESPONSE BOUNDARY]}
\noindent Use only the attached images, permitted history, and this prompt. Do not use tools or external information, or infer reference answers from wording, identifiers, or option order. Return only the specified JSON, using English for any free-text field, without explanations, absolute attention levels, rankings, timelines, or separate predictions.\par
\end{EvaluationPrompt}

\clearpage
\begingroup
\setlength{\parindent}{0pt}
\setlength{\parskip}{5pt}
\setlength{\tabcolsep}{4.5pt}
\renewcommand{\arraystretch}{1.17}
\newcommand{\para}[1]{\par\noindent\textbf{#1}\enspace\ignorespaces}
\newcommand{\tabtitle}[1]{\par\noindent\textbf{#1}\par}
\section{Matched Questions and Object Discovery}
\label{app:matched_questions}
\subsection{The motion--passage gap on matched questions}\label{app:matched_input}

For GPT-5.6 Sol, the accuracy gap between object motion and passage judgments is 40.87 percentage points (pp) with the questions and options alone and 39.77 pp with frames (Table~\ref{tab:nohistory}, computed from the underlying counts). The question-only run thus reproduces the magnitude of the full-set gap. To examine how the two tasks change on common observations, we use the 394 motion--passage comparisons in Table~\ref{tab:targets}. Of these, 348 ask about the same target in the same clip with identical start and end times.

On those \textbf{348 matched comparisons}, the number of correct passage answers rises from 92 to 167 between the question-only and frame conditions, a gain of 21.55 pp. Correct motion answers rise from 273 to 302, a gain of 8.33 pp. The passage gain exceeds the motion gain by \textbf{13.22 pp} (paired 95\% CI 4.85 to 21.58). Thus the motion--passage gap on these same targets narrows from 52.01 to 38.79 pp. All 394 comparisons show a similar pattern (Table F1).

\tabtitle{Table F1. Sol's answers under the two input protocols used in Table~\ref{tab:nohistory}.}
The middle columns count correct answers. The last column subtracts the motion gain from the passage gain. Each successive row restricts the preceding question set.

{\small\centering
\begin{tabularx}{\linewidth}{@{}Xrrrr@{}}
\toprule
 & Questions & Object motion & Passage & Extra passage gain\\
Question set & motion/passage & Text $\to$ frames & Text $\to$ frames & pp [95\% interval]\\
\midrule
All questions & 425/675 & $333\to368$ & $253\to316$ & $+1.10$ [$-5.64$, $+7.84$]\\
Same clip and target & 394/394 & $314\to346$ & $105\to185$ & $+12.18$ [$+4.60$, $+19.76$]\\
Also same endpoints & 348/348 & $273\to302$ & $92\to167$ & $+13.22$ [$+4.85$, $+21.58$]\\
\bottomrule
\end{tabularx}\par}

Question order and the direction of change help explain why aggregate scores can move little. Across the 494 chronologically ordered passage questions, Sol goes from 139 to 232 correct with frames. On the 181 derived questions with the comparison order reversed, it goes from 114 to 84. Among the original chronological questions with a directional reference, correct decreases rise from 0/39 to 17/39 while correct increases fall from 117/142 to 105/142. Together these changes yield five additional correct answers across the 181 original directional questions.

Class frequencies also matter. Of the 394 linked passage questions, 217 have the reference \emph{irrelevant at both times}. Always choosing it scores 55.08\%, compared with Sol's 46.95\% with frames. Giving each of the four reference classes equal weight, mean recall rises from 22.80\% with text to 40.55\% with frames.

The question-only prompt asks Sol for its best answer from the question and options. The frame prompt directs it to judge from the supplied observations. The gains therefore describe a change in both the available evidence and the instructions. Paired intervals use 10,000 shared resamples of source recordings, stratified by collection as in Appendix~\ref{app:scoring_methods}.

\newpage
\subsection{Sensitivity to the choice of recording pairs}

For two targets with the same motion state, can a respondent identify an increasing passage constraint in one clip and a decreasing constraint in another? We retrospectively pair the original answers. Both directions must be correct for a pair to earn credit. We report the result for one selected set alongside the range over \emph{every} largest admissible set under each of three rules.

Of the 348 comparisons in Section~\ref{app:matched_input}, 119 have a moving or stationary target and a chronological increase (96) or decrease (23) in constraint. Each pair combines an increase and a decrease from different source recordings, matched on object category and reference motion state. Each question retains its original target description, wording and option order.

\tabtitle{Table F2. Candidate pairs and variation across eligible selections.}
\textit{(a) Pairing rules.} Rule 1 matches interval duration and uses each source recording once. Rules 2 and 3 each relax one condition. All three use overlapping data.

{\small\centering
\begin{tabularx}{\linewidth}{@{}lXlrrrr@{}}
\toprule
 & Unit used & Duration & Candidate & \multicolumn{2}{c}{Eligible pool} & Maximum\\
\cmidrule(lr){5-6}
Rule & once per set & must match & pairs & Questions & Recordings & pairs\\
\midrule
1 & Recording & Yes & 56 & 60 & 27 & 12\\
2 & Clip      & Yes & 56 & 60 & 27 & 18\\
3 & Recording & No  & 229 & 80 & 36 & 13\\
\bottomrule
\end{tabularx}\par}

\textit{(b) Correct answers.} The single-question columns count increasing and decreasing constraints in the rule 1 pool. The last three columns give the minimum and maximum counts of fully correct pairs over maximum-size sets.

{\small\centering
\begin{tabularx}{\linewidth}{@{}Xrrrrrr@{}}
\toprule
 & \multicolumn{2}{c}{Single questions} & \multicolumn{1}{c}{Selected set} & \multicolumn{3}{c}{Range across maximum-size sets}\\
\cmidrule(lr){2-3}\cmidrule(lr){4-4}\cmidrule(l){5-7}
Respondent and input & Up / 38 & Down / 22 & Rule 1 / 12 & Rule 1 / 12 & Rule 2 / 18 & Rule 3 / 13\\
\midrule
GPT-5.6 Sol, text   & 30 & 0  & 0  & 0 & 0 & 0\\
GPT-5.6 Sol, frames & 31 & 11 & 5  & 3--6 & 8--10 & 1--6\\
GPT-6 Astra, frames & 29 & 12 & 6  & 4--7 & 7--11 & 2--7\\
Human participant A & 29 & 19 & 10 & 6--12 & 9--14 & 6--12\\
Human participant B & 34 & 19 & 11 & 7--11 & 12--15 & 8--13\\
\bottomrule
\end{tabularx}\par}

Under rule 1, 56 candidates give 2,688 sets of 12 pairs, each using 24 different recordings. To select the displayed set using only reference labels and metadata, we shuffled the candidates 2,500 times (seed 1807), skipping used recordings in each pass. We kept the first set maximizing pair count, then category count, then stationary-pair count. It covers five categories and three stationary pairs. Sol with frames completes 5/12 pairs in this set and 3--6/12 across all 2,688 sets.

Rule 2 uses 36 distinct clips for 18 pairs and permits a recording to contribute to several pairs. Rule 3 uses 26 distinct recordings for 13 pairs and allows unequal durations. Sol with frames completes 8--10/18 and 1--6/13 pairs, respectively. Its text-only score of zero under all three rules follows from the same observation in Section~\ref{app:matched_input}, where it misses every original decrease. The ranges show how much the frame result depends on the choice of cases.

\newpage
\subsection{Tracing a target from motion to passage and discovery}

EgoPass-Bench asks how a target moves, how its effect on the wearer's passage changes, and which objects a respondent finds without a target cue. For each target, we link the answers given independently across tasks by Astra with frames and by each of the two participants in Table~\ref{tab:nohistory}.

\para{Motion correct, passage misjudged.} On the 394 shared targets in Table~\ref{tab:targets}, GPT-6 Astra answers 360 motion questions (91.37\%) and 262 passage questions (66.50\%) correctly with frames. It answers only the motion question correctly in 107 comparisons and only the passage question correctly in nine. Of the 388 comparisons with a linked wearer-translation question, Astra answers both the wearer and object motion questions correctly in 313. It still misses \textbf{96/313 passage judgments} (30.67\%, recording-level 95\% CI 25.04--36.30\%). Both human participants answer 74 of those 96 correctly, across 41 source recordings.

\tabtitle{Table F3. Passage errors and discovery omissions in Astra's linked answers.}
\textit{(a) Passage judgments across the 313 comparisons with both motion answers correct.} Each cell counts comparisons, grouped by how many participants answered the same passage question correctly.

{\small\centering
\begin{tabularx}{\linewidth}{@{}Xrrrr@{}}
\toprule
Astra passage answer & Both participants & One & Neither & Total\\
\midrule
Correct   & 155 & 32 & 30 & 217\\
Incorrect & \textbf{74} & 14 & 8 & \textbf{96}\\
\midrule
Total & 229 & 46 & 38 & 313\\
\bottomrule
\end{tabularx}\par}

Restricting the motion and passage questions to identical endpoints retains 277 of these comparisons. Astra misses 84 passage judgments, including 66 both participants answer correctly. Passage errors after two correct motion answers also occur in all six native-video configurations, including 126/276 for Qwen3.8 Max and 100/180 for Gemma 4 31B. Each denominator counts that configuration's own correct motion answers on both tasks.

\para{Named target judged correctly, then omitted without a cue.} In a separate analysis of 152 linked comparisons (Section~\ref{sec:outcomes}), the named target is the sole reference entity in a discovery question with identical ordered times. Astra answers 56 named-target passage questions correctly. Table F3(b) follows those same targets into their independently answered discovery questions.

\textit{(b) Uncued discovery after a correct named-target passage answer.}

{\small\centering
\begin{tabularx}{\linewidth}{@{}Xr@{}}
\toprule
Outcome on the corresponding discovery question & Questions\\
\midrule
Omitted the named target & \textbf{8}\\
Reported the target with another entity or change-label error & 14\\
Returned the exact entity set and correct change labels & 34\\
\midrule
Total & 56\\
\bottomrule
\end{tabularx}\par}

Across all 574 uncued questions, the two participants average 79.46\% on Discovery. Astra scores 55.69--55.81\% across feasible identity matches, and always returning an empty set scores 36.76\% (Table~\ref{tab:emptycontrols}). The human mean exceeds Astra's upper bound by $23.65 \pm 7.45$ pp (paired 95\% CI). The linked answers distinguish misjudging the effect of a named object from omitting an object whose effect the model judged correctly when prompted.
\endgroup

\end{document}